\documentclass[runningheads]{llncs}
\usepackage{graphicx}
\usepackage{tikz}
\usepackage{comment}
\usepackage{amsmath,amssymb} % define this before the line numbering.
\usepackage{color}

\usepackage{tabularx}
\usepackage{graphicx}
\usepackage{booktabs}
\usepackage{varwidth}
\usetikzlibrary{positioning}
\usetikzlibrary{arrows, decorations.markings,shapes,arrows,fit}
\usepgflibrary{shapes.arrows}
\usepackage{mathtools}
\usepackage{multirow}
\usepackage{multicol}
\usepackage{makecell}
\usepackage{fancyhdr}
\usepackage{url}

\usepackage[accsupp]{axessibility}  % Improves PDF readability for those with disabilities.

\begin{document}
% \linenumbers
\pagestyle{headings}
\mainmatter
\def\ECCVSubNumber{7528}  % Insert your submission number here

\title{Autoencoder-based background reconstruction and foreground segmentation with background noise estimation} % Replace with your title

\begin{comment}
\titlerunning{ECCV-22 submission ID \ECCVSubNumber} 
\authorrunning{ECCV-22 submission ID \ECCVSubNumber} 
\author{Anonymous ECCV submission}
\institute{Paper ID \ECCVSubNumber}
\end{comment}
%******************

%\begin{comment}
\titlerunning{Autoencoder-based background reconstruction}
\author{Bruno Sauvalle   \and
Arnaud de La Fortelle }%

\authorrunning{B. Sauvalle et al.}
\institute{Mines ParisTech PSL University 75006 Paris France 
\email{bruno.sauvalle@mines-paristech.fr}}

%\end{comment}
%******************
\maketitle

\begin{abstract}
Even after decades of research, dynamic scene background reconstruction and foreground object segmentation are still considered as open problems due various challenges such as illumination changes, camera movements, or  background noise caused by air turbulence or moving trees. We propose in this paper  to model the background of a frame sequence as a low dimensional manifold using an autoencoder and  compare the reconstructed background provided by this autoencoder with the original image to  compute the foreground/background segmentation masks. The main novelty of the proposed model is that  the autoencoder is also trained to predict the background noise, which allows to compute for each frame a pixel-dependent threshold to perform the foreground segmentation. Although the proposed model does not use any temporal or motion information, it exceeds the state of the art for unsupervised background subtraction on the CDnet 2014 and LASIESTA datasets, with a significant improvement on videos where the camera is moving. It is also able to perform background reconstruction on some non-video image datasets.
\end{abstract}

 \newlength{\largeur}
 
\section{Introduction}

We consider in this paper the tasks of dynamic background reconstruction and foreground/background segmentation. The dynamic background reconstruction task can be described in the following way: The input is a sequence $\mathcal X$ of consecutive frames $X_1, ..X_N$ showing a scene cluttered by various moving objects, such as cars or pedestrians, and the expected output is a sequence $\mathcal {\hat X} = \hat X_1, .., \hat X_N$ of frames showing the backgrounds of each scene without those objects. If the camera is fixed and the illumination conditions do not change, the various frames $\hat X_1, .., \hat X_N$ will be nearly identical. However if this is not the case, then these frames can be very different from each other.  The foreground/background segmentation task similarly  takes as input the same kind of frames sequence 
 $X_1, ..X_N$, but the  expected output is a sequence $\mathcal M$ of foreground masks $M_1, .., M_N$ whose values at the pixel $p$ are equal to zero if this pixel shows the background in the considered frame, and equal to 1 if the background is masked by a foreground moving object at this pixel (Fig. \ref{fig:example1}).

\setlength{\largeur}{20mm}
\begin{figure}
\centering
\begin{tabular}{*{3}{m{22 mm}}}

   \includegraphics[width=\largeur]{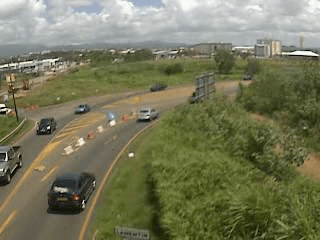} &
  \includegraphics[width=\largeur]{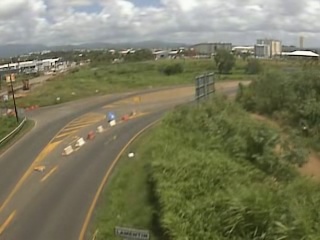} &
   \includegraphics[width=\largeur]{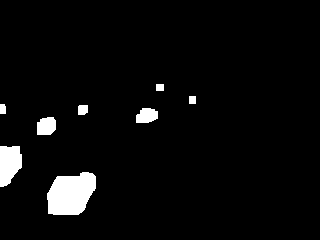} \\
      \includegraphics[width=\largeur]{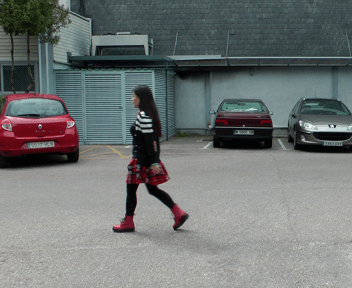} &
  \includegraphics[width=\largeur]{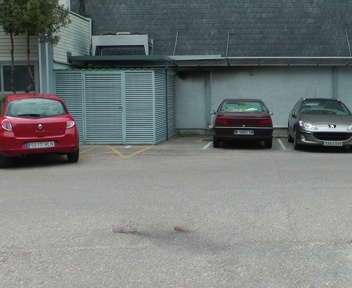} &
    \includegraphics[width=\largeur]{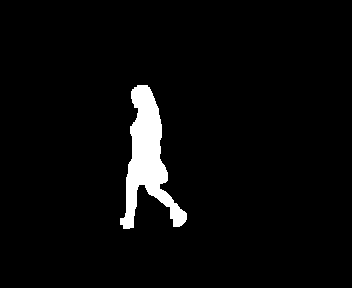} \\
       \includegraphics[width=\largeur]{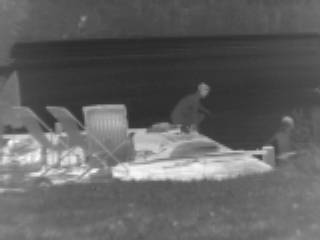} &
  \includegraphics[width=\largeur]{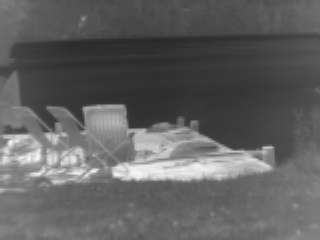} &
    \includegraphics[width=\largeur]{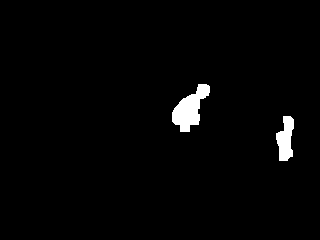} 
  \\

 % \makecell{input \\ frame}  &   \makecell{predicted \\ background}  &   \makecell{foreground \\ mask}  
\end{tabular}
\caption{The proposed model takes as input a frame from the associated video (left column) and provides a reconstruction of the background (middle column) and a foreground mask (right column).  }
\label{fig:example1}
\end{figure}

 This task is often called background subtraction because the pointwise multiplication of the mask $M_k$ and the input image $X_k$ gives an image showing only the foreground moving objects present in $X_k$, the input image background being replaced by a black background. The applications of background subtraction are very diverse \cite{Garcia-Garcia2020a}: road, airport, store, maritime or military surveillance, observation of animals and insects, motion capture, human-computer interface, video matting, fire detection... Although the task of automatic background subtraction has been studied for more than 30 years \cite{Wren1997}, it is still considered as an open problem due to the various challenges appearing in real applications: illumination changes, high level of occlusion of the background, background motions caused by moving trees or water, challenging weather conditions, presence of shadows... Dynamic background reconstruction models  have been used for background subtraction, but are now also implemented as components of unsupervised object detection and tracking models   \cite{Jiang,Henderson2020,Wu2021}.

The model presented in this paper starts from the classical assumption that the dynamic background of a scene can be modeled as a low dimensional manifold and uses an autoencoder to learn this manifold and perform dynamic background reconstruction. It then compares the input frame with the  associated background predicted by the autoencoder to build the foreground/segmentation mask. The main contributions of this paper are the following : \begin{itemize}
\item We implement a new loss function to train the autoencoder,  which gives a high weight to reconstruction errors associated to background pixels and a low weight to reconstruction errors associated to foreground pixels, and shows better performance than the $L_1$ loss usually considered for this task.
\item We train the autoencoder to provide a background reconstruction, but also a background noise estimation, which gives a pixelwise estimate of the uncertainty of the background prediction. This noise estimation map is used to adjust the threshold necessary to compute the background/foreground segmentation mask.
\item We reduce the risk of overfitting by implementing an early stopping criterion and adapting the autoencoder parameter count to the complexity of the background sequence. 
\end{itemize}

The paper is structured as follows : We first review related work in section 2, then describe the proposed model in section 3. Experimental results on various datasets are then provided in section 4.

\section{Related work}

Background subtraction methods can be split between supervised methods, which require labeled data, and unsupervised methods. Fully unsupervised methods are methods which do not require training data and can be applied to any video sequence without any update of the model parameters. 

One can  classify unsupervised methods as statistical methods or reconstruction methods. Statistical methods rely on a statistical modeling of the distribution of background pixel color values or other local features to predict whether a particular pixel is foreground or background. These statistical models can be parametric (univariate gaussian \cite{Wren1997}, mixture of gaussians \cite{Stauffer1999},  clusters \cite{Link2018}, Student's t-distributions \cite{Mukherjee2012}, Dirichlet process mixture models \cite{Borse2015}, Poisson mixture models \cite{Faro2011}, asymmetric generalized gaussian mixture models \cite{Elguebaly2013}, etc.) or non parametric (pixel value histograms \cite{Zhang2008}, kernel density estimation \cite{Elgammal2000},  codebooks \cite{Kim2004}, history of recently observed pixels \cite{Barnich2009,Hofmann2012}, etc.). The efficiency of these methods can be increased by using as input not only the pixel color values, but also  features attached to superpixels \cite{Chen2019c} or local descriptors which are robust to illumination changes, such as SIFT \cite{Smeulders}, LBP or LBSP descriptors. \cite{St-Charles2015c,St-Charles2016b}. 
If the camera is static, the segmentation of moving objects on a scene can also be performed by evaluating  the motion associated to each pixel, using optical flow or flux tensor models. The blobs produced by these models are generally very fuzzy, but can be used as input to more complex models \cite{Bunyak2007,Wang2014}.  Several unsupervised models can be also combined to form a more accurate ensemble model \cite{Bianco2017}.

Reconstruction methods use a background reconstruction model to predict the color (or other features) of the background at a particular pixel. The difference between the current image and the predicted background is then computed and followed by a thresholding  to decide whether the a pixel is background or foreground.  Pixelwise reconstruction models try to predict the value of a  background pixel at a particular frame using the sequence of values of the pixel of the last frames using a filter, which can be a Wiener filter \cite{Toyama1999}, a Kalman filter \cite{Ridder1995} or a Chebychev filter \cite{Chang2004}. A  global prediction of the background can also be performed using the assumption that the background frames form a low dimensional manifold, which motivates the use of dimensionality reduction techniques such as principal component analysis (PCA) \cite{Oliver2000}. One can add to this approach a prior on the sparcity of the foreground objects by using a $L_1$ loss term applied to the foreground residuals, which leads to the development of models based on robust principal component analysis (RPCA) \cite{Wright2009,Candes2011}. More complex norms and additional regularizers have been proposed to improve the performance of this approach \cite{Mairal2010,Liu2015,Xin2015,Javed2017a,Javed2019}. Non-linear dimensionality reduction using an autoencoder for background reconstruction has already been implemented \cite{Farnoosh2019a,Rezaei2020} and is further developed in the proposed model.

Supervised methods require labeled data as input. The labeled data are sets of pairs $ (X_k, M_k)$, where  the image $X_k$ is an image extracted from the sequence $X_1, ..X_N$ and the foreground mask $M_k$ has to be provided by a human intervention. Supervised  algorithms using linear methods such as as  maximum margin criterion \cite{Li2004,Diana2010} or graph signal reconstruction methods \cite{Giraldo2020} have  been proposed, but the current best performing supervised models use deep learning techniques with convolutional encoder-decoder structures  \cite{Lim2018a,Lim2020,Mandal2020}, U-net structures \cite{Rahmon2020,Mondejar-Guerra2020} or GANs  \cite{Sultana2019,Zheng2020}.

Although supervised models can reach very high accuracy results on a given video after training, their ability to generalize to new videos remain a major issue, and evaluation on unseen scenes may lead to unfavorable results compared  to unsupervised algorithms \cite{Mandal2020}.  A spatio-temporal data augmentation strategy has been proposed  \cite{Tezcan2021} to improve generalization.  One can also use  as additional input to the deep learning model the output of an unsupervised background subtraction model  \cite{Rahmon2020,Pardas2021}. A  background  subtraction model can also be substantially improved by combining its results with the output of a supervised semantic segmentation model  \cite{M.BrahamSebastien.Pierard2017,Zeng2019}.

Several surveys  \cite{Bouwmans2014d,Mondejar-Guerra2020,Kalsotra2021,Mandal2021} discuss background reconstruction and background subtraction models.

\section{Model description}

The proposed model is a reconstruction model and has a general structure similar to the DeepPBM model \cite{Farnoosh2019a}: We assume that the background frames form a low dimensional manifold and train an autoencoder to learn this manifold from the complete video using a reconstruction loss. 
We however observe that the DeepPBM model shows the following shortcomings: \begin{itemize}
\item  The structure of the autoencoder and the number of latent variables have to be defined on a scene by scene basis, which requires a human supervision. If the number of latent variables is too high, the autoencoder quiclky  learns to reproduce the foreground objects, a phenomenon we call ovefitting, and fails to generate a proper background.
\item This model is able to handle changes in lightning conditions, but struggles to handle camera movements.
\item The thresholding mechanism is not able to cope with dynamic backgrounds such as  clouds or trees, which leads to false detections.
\end {itemize}

In order to handle these issues, we implement the following features: 
\subsection{Reconstruction loss using background bootstrapping}
\label {sec:loss}

We implement a reconstruction loss using background bootstrapping \cite{Sauvalle2022}, which we found to be more efficient than the $L_1$ loss for dynamic background reconstruction. In the case of dynamic background reconstruction, this loss function allows to reduce the risk of overfitting  to the foreground objects by giving a higher weight to background pixels than to foreground pixels during the optimization process.

The proposed reconstruction loss can be described by the following formulae: 

We note $x_{n,c,i,j}$ the pixel color value of the image $X_n$ for the channel $c$ at the position $(i,j)$ with $1 \le c \le 3$,$1 \le i \le h$ and $1 \le j \le w$, and $\hat x_{n,c,i,j}$ the pixel value of the reconstructed background $\hat X_n$ for the same color and position.
The local $l_1$ error associated to the pixel $(i,j)$ is  
\begin{equation} l_{n,i,j} = \sum_{c=1}^{3} \lvert \hat x_{n,c,i,j} -x_{n,c,i,j} \rvert \label{equ_1}. \end{equation}
If at least one of the color channels give a high error, then $ l_{n,i,j} $ is large and  the pixel $(i,j)$ of the image $X_n$ is considered to be a foreground pixel.
A soft foreground mask $m_n \in [0,1]^{h\times w}$ for the image $X_n$ is then computed using the formula
\begin{equation} m_{n,i,j} = \tanh \left( \frac  {l_{n,i,j}} {\tau_1} \right) \label{equ_2},\end{equation}
where $\tau_1$ is some positive hyperparameter, which can be considered as a soft threshold. 
A spatially smoothed version $ \tilde m_{n,i,j}$ of this mask is then computed by averaging using a square kernel of size $(2k+1)\times (2k+1)$, with $k = \left \lfloor{w/r}\right \rfloor $  (where $w$ is the image width and $r$ is some integer hyperparameter):
\begin{equation}  \tilde m_{n,i,j}(\hat X_n,X_n) = \frac 1{(2k+1)^2} \sum_{l=-k,p=-k}^{l = k, p = k}m_{n,i+l,j+p} \end{equation}
The associated pixel-wise weight $w_{n,i,j}^{\text{bootstrap}}$ is then defined as 
\begin{equation} w_{n,i,j}^{\text{bootstrap}} = e^{- \beta  \tilde m_{n,i,j}}, \end{equation}
where $\beta$ is some positive hyperparameter.
The reconstruction loss of the auto-encoder is then computed by weighting the pixelwise  $L_1$ losses $l_{n,i,j}$ using these bootstrap weights: 
\begin {equation}
\mathcal L_{\text{rec}}(\mathcal {\hat X}, \mathcal {X}) = 
\frac {1}{Nhw}\sum_{n= 1, i=1,j=1}^{N,h,w}  w_{n,i,j}^{\text{bootstrap}}l_{n,i,j}
\end{equation}

We do not use any optical flow or motion estimation input in the proposed model because we want it to be able to handle situations where the camera is moving.

\definecolor{mycolor}{RGB}{200,200,200}

\tikzset{trapezium stretches=true}

\setlength{\largeur}{20mm}

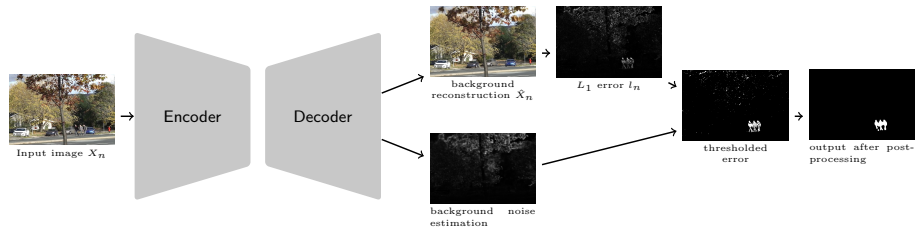
\begin{figure*}{}

\centering

  \scalebox{0.70}{
  
\begin{tikzpicture}[scale=0.8, every node/.append style={thick,rounded corners=2pt,font=\tiny}]		
\node [trapezium, fill=mycolor!100, minimum width=32mm , minimum height=22mm,shape border rotate=270,trapezium right angle=86,trapezium left angle=86, anchor = east,font=\sffamily\small,inner sep=0pt] (E) at (1.5,0)  {Encoder};

\node [trapezium, fill=mycolor!100, minimum width=32mm , minimum height=22mm,shape border rotate=90,trapezium right angle=86,trapezium left angle=86,right=2.5mm of E, anchor = west,font=\sffamily,inner sep=0pt] (D) at (1.5,0) {Decoder};

      \node(input image) at (-3, 0) {\begin{varwidth}{2cm}\includegraphics[width=\largeur]{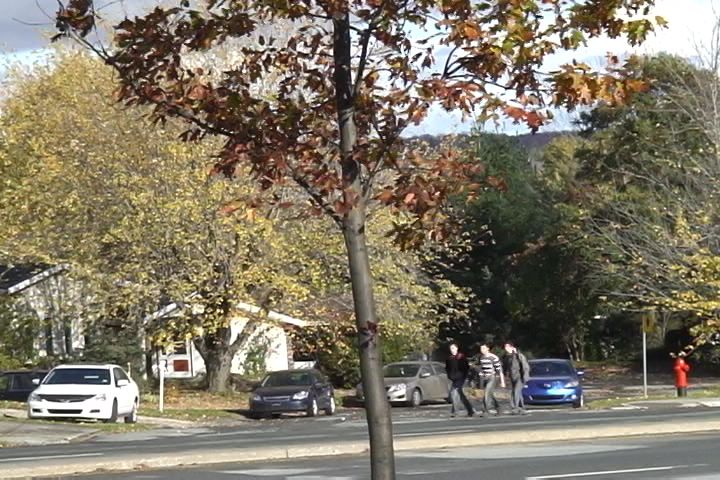}\\ \centering Input image $X_n$  \end{varwidth}};
      \node(background reconstruction) at (7,1.5)  {\begin{varwidth}{2cm} \includegraphics[width=\largeur]{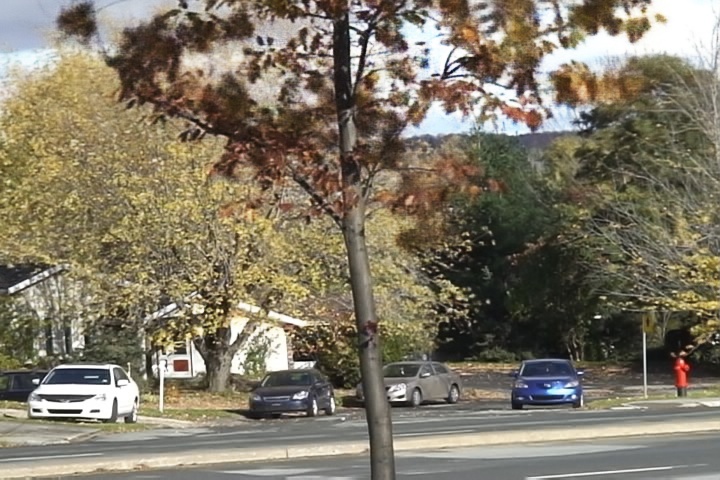} \\ \centering background reconstruction $\hat X_n$ \end{varwidth}};
     \node(noise) at (7,-1.5) {\begin{varwidth}{2cm}\includegraphics[width=\largeur]{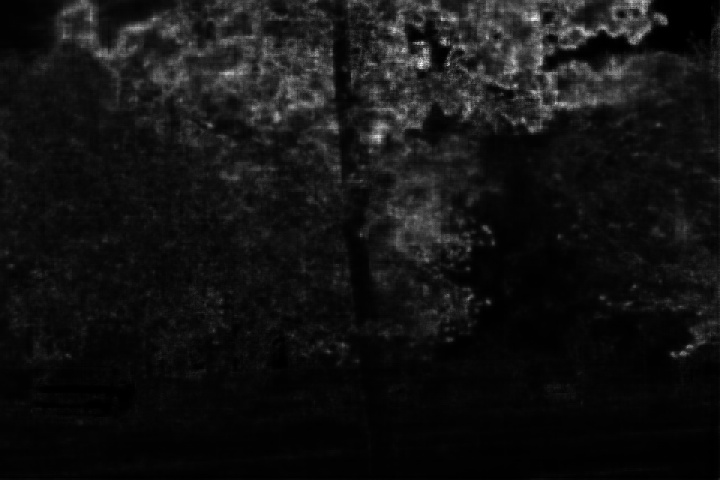} \\ background noise estimation\end{varwidth}};
     \node (diff) at (10 ,1.5)   {\begin{varwidth}{2cm} \includegraphics[width=\largeur]{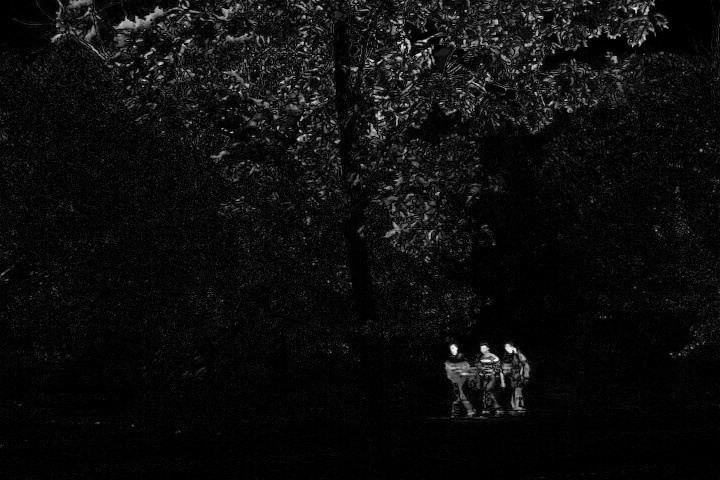}\\  \centering  $L_1$ error  $l_n$ \\ \quad \end{varwidth}};
     \node (thresholded) at (13,0) {\begin{varwidth}{2cm} \includegraphics[width=\largeur]{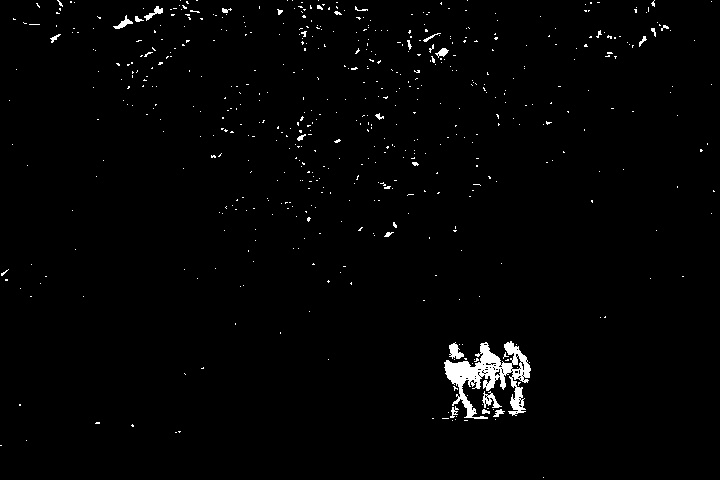}\\ \centering thresholded \\ error \end{varwidth}};
     \node (output) at (16,0) {\begin{varwidth}{2cm} \includegraphics[width=\largeur]{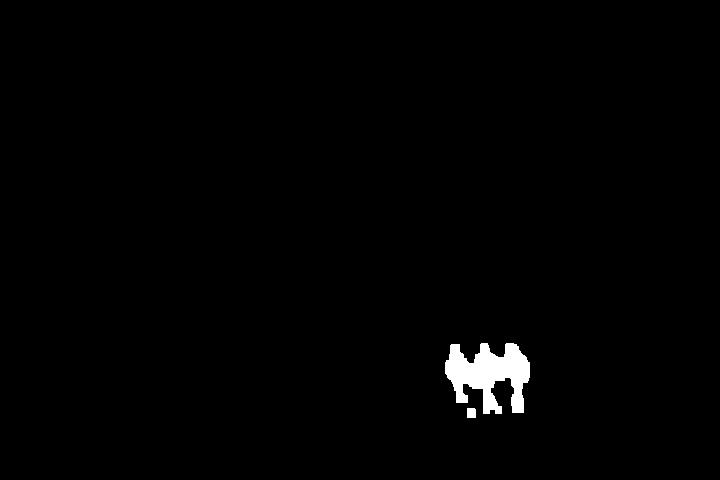}\\ output after post-processing \end{varwidth}};

     \path[->,thick]
         	(input image) edge  (E)    
		(D) edge (background reconstruction)
		(D) edge (noise)
		(background reconstruction) edge (diff)
		(diff) edge (thresholded)
		(noise) edge (thresholded)
		(thresholded) edge (output) ;
		
\end{tikzpicture}
}

\caption{Schematic of the proposed model during inference (Error and noise images are normalized in the range [0,1])}
\label{fig:model}
\end{figure*}

\subsection{Optimized thresholding using background noise estimation}

We remark that the bootstrap pixel weights $w_{n,i,j}^{\text{bootstrap}}$ can be used to get an estimate of the level of background noise of a frame sequence, considering that these weights are close to one when the associated pixel is a background pixel, and close to zero when this is not the case. 

We therefore add a fourth output channel to the auto-encoder, which is dedicated to give an estimate $\hat l_{n,i,j}$ of the value of the $L_1$ error $l_{n,i,j}$ for each pixel $(i,j)$ for the frame $X_n$ (Fig. \ref{fig:model}). 

The associated loss function is weighted using the bootstrap weights in order to limit its scope to background regions:
\begin {equation}
\mathcal L_{\text{noise}} = \frac {1}{3Nhw}\sum_{n=1 i=1,j=1}^{N,h,w}  w_{n,i,j}^{\text{bootstrap}} \lvert \hat l_{n,i,j}-l_{n,i,j}\rvert  \label{equ_noise}
\end{equation}

When the background is very noisy, the autoencoder is not able to predict accurately the value of a background pixel color. As a consequence, the expectation of $l_{n,i,j}$ is large, which leads to a high value of $\hat l_{n,i,j}$.

The autoencoder is trained using the sum of the reconstruction loss and the loss associated to the background noise estimation. The complete loss function is then   
\begin{equation}
\mathcal L = \mathcal L_{\text{rec}}+  \mathcal L_{\text{noise}}.
\end{equation}
The gradients of the weights $w_{n,i,j}^{\text{bootstrap}}$ are not computed during the optimization process \cite{Sauvalle2022}. We also do not use the gradient of $l_{n,i,j}$ in equation \ref{equ_noise} because we do not want the quality of the background reconstruction be impacted by the background noise estimation optimization process.

In order to set the pixelwise  threshold $\tau_{n,i,j}$ associated to the pixel $(i,j)$ of the frame $X_n$ and necessary to compute the background/foreground segmentation mask, we also take into account the average illumination $\hat I_n$ of the reconstructed background $\hat  X_n$, as defined by the formula
\begin {equation}
 \hat I_n= \frac {1}{3hw}\sum_{c=1, i=1,j=1}^{3,h,w}\lvert \hat x_{n,c,i,j} \rvert.
 \label{illumination}
\end {equation}

 The threshold $\tau_{n,i,j}$  is then set according to the formula 
\begin {equation}
 \tau_{n,i,j} =   \alpha_1 \hat I_n + \alpha_2\hat l_{n,i,j}, 
\end {equation}
where $ \alpha_1$ and $ \alpha_2$ are two positive hyperparameters.
The motivation of this  formula is  that if the background noise is very high at some pixel, we have to increase the associated threshold for background/foreground segmentation in order to prevent the misclassification of background pixels as foreground caused by background noise.

For a given frame sequence $X_1,...,X_n$ and a reconstructed background sequence $\hat X_1,...,\hat X_n$, we then compute the foreground mask $M_n$ before post-processing  using the thresholding rule $M_{n,i,j} = 1$ if and only if $l_{n,i,j} > \tau_{n,i,j}$.

A post-processing is then applied in order to remove rain drops, snow flakes, and other spurious detections.
It is composed of two morphological operations: a morphological closing using a $5 \times 5$ square structural element, followed by a morphological opening with a $7 \times7$ square structural element. 
\subsection{Detecting complex background changes}
The improved reconstruction loss function introduced in \ref{sec:loss} reduces the risk of overfitting, but is not able to prevent it completely. We observe that the risk of overfitting increases when the number of optimization iterations and  the number of parameters of the network increase. This is a significant issue because sequences showing background changes require a high number of training iterations and a model with a large number of parameters. In order to prevent overfitting, the number of training iterations and the complexity of the model are therefore adjusted to the complexity of the backgrounds sequence.

The main challenge here is to estimate without any human supervision whether  the video shows substantial background changes or not. We observe however that the proposed model can be used to answer to this question. In order to do this, we first train the model for a fixed small number $N_{\text{eval}}$ of iterations, which is however sufficient to get a rough evaluation of the background changes. 
Using this trained model, we compute $B_{\text{eval}}$ reconstructed backgrounds $\hat X_n$ using frames $X_n$ sampled from the sequence $\mathcal X$.
We then compute the temporal median $\hat X$ of these backgrounds and compare this median background with the reconstructed backgrounds $\hat X_n$,  computing soft masks following the same method and parameters as in formula \ref{equ_1} and \ref{equ_2}: 

\begin{equation} l_{n,i,j} = \sum_{c=1}^{3} \lvert \hat x_{c,i,j} -\hat x_{n,c,i,j} \rvert \end{equation}

\begin{equation} m_{n,i,j} = \tanh \left( \frac  {l_{n,i,j}} {\tau_1} \right).  \end{equation}

We then consider the average soft mask value over the $B_{\text{eval}}$ reconstructed backgrounds  

\begin{equation} \bar m = \frac 1{B_{\text{eval}}hw}\sum_{n,i,j}^{B_{\text{eval}},h,w} m_{n,i,j}. \end{equation}

If $\bar m $  is higher than a threshold $\tau_0$, we consider that  the background is a complex background. The partially trained model is discarded, a new autoencoder is created with more parameters and the number of training iterations is set to $N_{\text{complex}}$ with a minimum of $E_{\text{complex}}$ epochs for very long sequences.

If this ratio is lower than $\tau_0$, we consider that the background is a simple background, keep the partially trained model, and finish the training, with a total number of training iterations set to  $N_{\text{simple}}$.

\section{Experimental results}

\subsection {Evaluation method}

We consider the following benchmark datasets for background subtraction: CDNET 2014, LASIESTA and BMC 2012 and use the same model hyperparameters on these three datasets.

We use the  public implementations of the algorithms PAWCS \cite{St-Charles2016b} and SuBSENSE \cite{St-Charles2015c} provided with the BGS library  \cite {Sobral2013}   to get baseline performance estimates for these methods when they are not available. We rely on published results for the other state of the art methods which do not have public implementations.

We use the  F-measure as main evaluation criteria. To compute the F-measure associated to a sequence of foreground masks predictions $M_1,..,M_n$, we first compute the sums $TP, TN, FP, FN$ of the true positives, true negatives, false positives and false negatives associated to the sequence of masks $M_1,..,M_n$, and then compute the F-measure associated to this sequence as the harmonic mean of precision and recall, which can also be described by the formula

\begin{equation} F = \frac {TP}{TP+\frac 12(FN+FP)} \label{equ_f_measure}. \end{equation}

 Implementation details and autoencoder architecture are described in the appendix.

\subsection{CDnet 2014 dataset}

The CDnet 2014 dataset \footnote{\url{http://changedetection.net/}}  \cite{Wang2014a} is composed of 53 videos, for a total of 153 278 frames, selected to cover the various challenges which have to be addressed for background subtraction: dynamic background (scenes with water or trees), camera jitter, intermittent object motion, presence of shadows, images captured by infrared cameras, challenging weather (snow, fog), images captured with a low frame rate, night images, images filmed by a pan-tilt-zoom camera, air turbulence. Ground truth foreground segmentation masks are provided for all frames of the dataset, with specific labels for shadow pixels which are not considered in the  F-measure computation.
We provide in Table \ref{table:cdnet}  the F-measure results per category of the proposed model for each category of the CDnet 2014 dataset, with a comparison with the results obtained by other unsupervised models. 

The proposed model, despite the various mechanisms implemented to limit overfitting, nevertheless suffers from overfitting  when big foreground objects stay still or moves very slowly  in a frame sequence. This phenomenon is observed  on three sequences of the CDnet dataset, "office", "library" and "canoe" (\ref{fig:overfitting}), although the associated backgrounds are correctly classified as simple by the model.  As a consequence, the average F-score on corresponding categories "baseline", "thermal" and "dynamic backgrounds" are below the scores of other state of the art models. 
\setlength{\largeur}{15mm}

\begin{figure}
\centering
  \scalebox{0.75}{
\begin{tabular}{*{4}{m{17 mm}}}

               \includegraphics[width=\largeur]{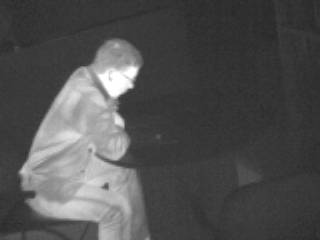} &
                    \includegraphics[width=\largeur]{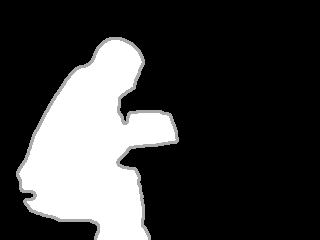} &
  \includegraphics[width=\largeur]{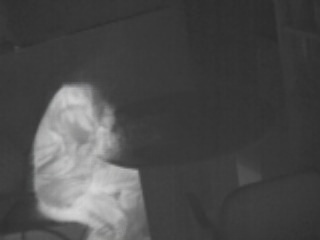} &
    \includegraphics[width=\largeur]{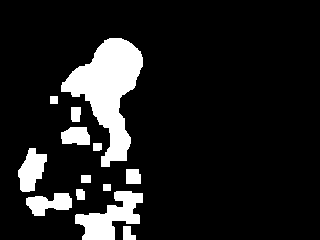} \\

                     \includegraphics[width=\largeur]{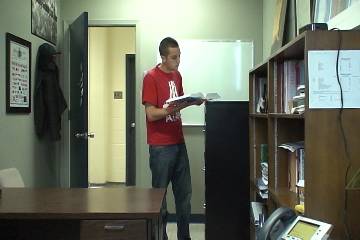} &
                           \includegraphics[width=\largeur]{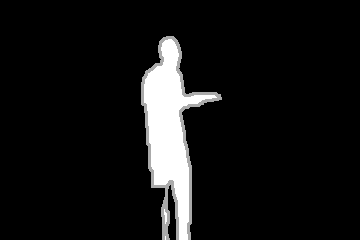} &
  \includegraphics[width=\largeur]{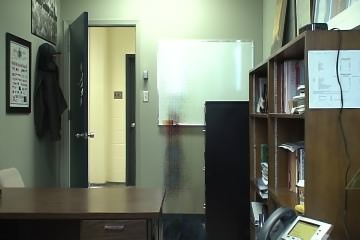} &
    \includegraphics[width=\largeur]{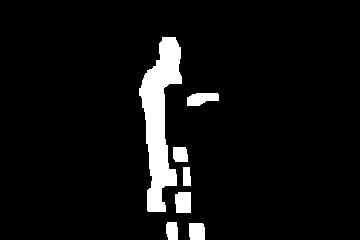} \\

                     \includegraphics[width=\largeur]{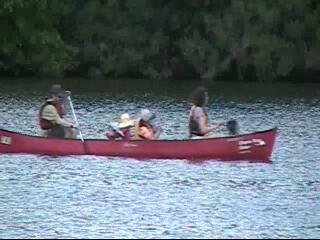} &
                           \includegraphics[width=\largeur]{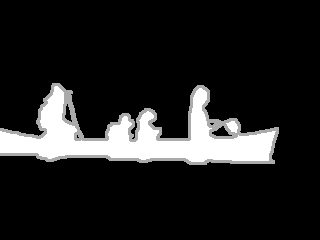} &
  \includegraphics[width=\largeur]{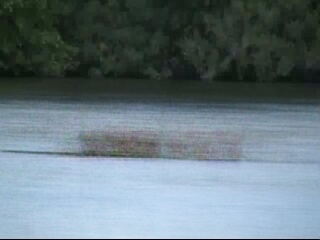} &
    \includegraphics[width=\largeur]{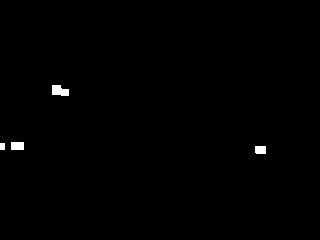} \\

         \includegraphics[width=\largeur]{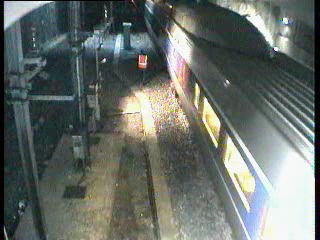} &
               \includegraphics[width=\largeur]{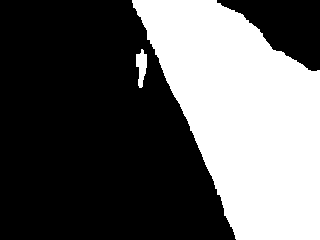} &
  \includegraphics[width=\largeur]{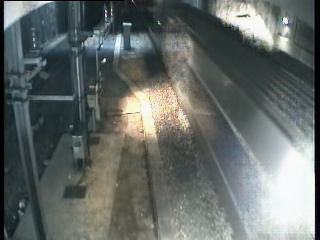} &
    \includegraphics[width=\largeur]{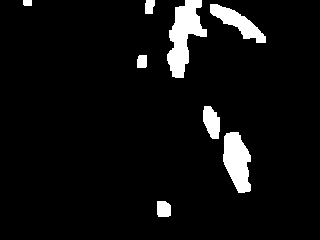} \\

  \makecell{input \\ frame}  &   \makecell{ground \\ truth} &   \makecell{predicted \\ background}  &   \makecell{foreground \\ mask}   
\end{tabular}}

\caption{Examples of overfitting on the datasets CDnet 2014  and BMC 2012 for the sequences "library", "office", "canoe" and "video007"}
\label{fig:overfitting}
\end{figure}

\begin{table*}
  \centering
  \caption{Comparison of top BGS algorithms according to the per-category F-measures on CDNet-2014 \\  (sources :  CDnet website http://jacarini.dinf.usherbrooke.ca/results2014/1158/,\cite{Javed2017a}, \cite{Javed2019}).
  }
  \smallskip
  \scalebox{0.70}{
        \begin{tabular}{l|lllllllllll|l}
            \hline
             Method  & \makecell[l]{Bad\\weather}\! & \!\!\! \makecell[l]{Base-\\line}&\!\!\!\! \makecell[l]{Camera\\jitter} \! & \!\! \makecell[l]{Dynamic\\backgr.}\!\!  &  \!\! \makecell[l]{Int.\ obj.\\motion} & \!\! \makecell[l]{Low\\framerate}\!\!\! & Night & PTZ &Shadow\!\!& Thermal\!\! &    \makecell[l]{Turbu-\\lence} & Overall\\
             \hline
		Unsupervised: &&&&&&&&&&&& \\
	     AE-NE (ours)  & 0.8337& 0.8959& 0.9230 & 0.6225 & 0.8231 & 0.6771  &  0.5172
             & 0.8000  & 0.8947 &   0.7999& 0.8382 & \bf{0.7841}\\
             IUTIS-5 \cite{Bianco2017}& 0.8248&0.9567& 0.8332 & 0.8902 & 0.7296 &  0.7743& 0.5290 & 0.4282 
             & 0.9084  & 0.8303 &   0.7836& 0.7717\\

             WisenetMD \cite{Lee2019b} & 0.8616&0.9487& 0.8228 & 0.8376 & 0.7264 &  0.6404 & 0.5701& 0.3367 
             & 0.8984  & 0.8152 &   0.8304& 0.7535\\
             SuBSENSE \cite{St-Charles2015c} & 0.8619& 0.9503 & 0.8152&0.8177 &0.6569 &   0.6445 & 0.5599 & 0.3476 
             & 0.8986  & 0.8171 & 0.7792& 0.7408 \\
            PAWCS \cite{St-Charles2016b}   & 0.8152& 0.9397 &0.8137& 0.8938 & 0.7764 &  0.6588 & 0.4152 & 0.4615 
             & 0.8913  & 0.8324 &   0.6450& 0.7403\\
             C-EFIC \cite{Allebosch2016} & 0.7867& 0.9309 & 0.8248 & 0.5627 & 0.6229 & 0.6806 & 0.6677 & 0.6207 & 0.8778 & 0.8349 & 0.6275 & 0.7307  \\
             MSCL  \cite{Javed2017a} & 0.83 & 0.87 & 0.83 & 0.85 & 0.80 & n/a & n/a & n/a & 0.82 & 0.80 & 0.80 & n/a \\
             B-SSSR \cite{Javed2019}& 0.92 & 0.97 & 0.93 & 0.95 & 0.74 & n/a & n/a& n/a & 0.93 & 0.86 & 0.87 & n/a \\

             \hline
             Supervised: & & & & & & & & & & &  \\
             FgSegNet v2 \cite{Lim2020}& 0.9904  & 0.9978 & 0.9971 & 0.9951 & 0.9961 & 0.9336 &0.9739  & 0.9862 & 0.9955 &  0.9938 & 0.9727  & 0.9847 \\
         	BSUV-Net 2.0 \cite{Tezcan2021}& 0.8844 & 0.9620 & 0.9004  & 0.9057 & 0.8263 & 0.7902 & 0.5857 & 0.7037 & 0.9562 & 0.8932 & 0.8174 &   0.8387 \\

             \hline 
        \end{tabular}
  }
  \label{table:cdnet}
  %\vglue -2 ex
\end{table*}

Despite this issue, the proposed model gets a higher average F-measure on the CDnet 2014 dataset than all  published unsupervised models, with an average F-measure of  0.784. One can observe  a significant improvement in accuracy with the proposed model in the  "pan-tilt-zoom" (PTZ) category, showing that it is able to better handle situations where the camera is moving.
\subsection {LASIESTA dataset}
\label{implementation}

The LASIESTA dataset\footnote{\url{https://www.gti.ssr.upm.es/data/}} \cite{Cuevas2016}   is composed of 48 videos grouped in 14 categories, for a total of 18 425 video frames. All frames are provided with ground truth pixel labels, with a specific label for pixels associated to stopped moving objects which are excluded from the F-measure computation. We  provide 
 in Table \ref{table:lasiesta}  the average F-measure results of the proposed model for all 14 categories. Out of the 48 videos of the dataset, 4 videos are taken with a moving camera (categories IMC and OMC), and 24 videos include simulated camera motion  (categories ISM and OSM). These 28 videos which include real or simulated camera motion are very difficult for existing background subtraction models and  to our best knowledge, no paper has ever published category-wise evaluation results for these videos. In order to allow a comparison with these published results, we therefore also provide the average F-measure over the 10 categories showing only videos taken from a fixed camera. We observe that the proposed model performs slightly better than available unsupervised algorithms on static scenes, and with a significant improvement on scenes where the camera is moving.

\begin{table*}
  \centering
  \caption{Average per category of video F-measures  on LASIESTA\\ (sources : \cite{Cuevas2016},\cite{Berjon2018a}, authors experiments for PAWCS and SuBSENSE)}
  \smallskip
  \scalebox{0.75}{

          \begin{tabular}{l|llllllllll|llll|ll}
            \hline
            & \multicolumn{10}{c| }{static camera} &  \multicolumn{4}{c|}{moving camera } & \\
            & \multicolumn{10}{c| }{} &  \multicolumn{4}{c|}{ or simulated motion } & \\

              Method &ISI& ICA &IOC &  IIL  &  IMB  &  IBS  & OCL & ORA  &OSN & OSU & IMC  &  ISM  &  OMC &  OSM  & 
               \makecell[l]{Average.\\ 10 categ.} &\makecell[l]{Average.\\ 14 categ.} \\
           
             \hline
             AE-NE (ours) & 0.91 & 0.88 & 0.91 & 0.81 & 0.92 & 0.79 & 0.94 & 0.80 & 0.82 & 0.91 & 0.83 & 0.79 & 0.86 & 0.89 & \bf{0.87} & \bf{0.86} \\
             PAWCS \cite{St-Charles2016b} & 0.90 & 0.88&0.90 &0.79 &0.81 &0.79 & 0.96&0.93 &0.69 &0.82 &0.48&0.77 & 0.43 & 0.75 & 0.85 & 0.78 \\
             SuBSENSE \cite{St-Charles2015c}&0.90 &0.89 &0.95 &0.65 &0.77 &0.73 &0.92 &0.90 &0.81 &0.79 &0.33 & 0.70 & 0.31 & 0.65 & 0.83 & 0.73 \\
             Cuevas \cite{Berjon2018a} & 0.88 &0.84 &0.78 &0.65 &0.93 &0.66 &0.93 &0.87 &0.78 & 0.72& n/a&n/a &n/a &n/a &0.81 & n/a \\
             Haines \cite{Haines2014} &0.89 &0.89 &0.92 &0.85 &0.84 &0.68 &0.83 &0.89 &0.17 &0.86 &n/a&n/a&n/a&n/a&0.78 & n/a\\
             Maddalena \cite{Maddalena2012b} &0.95 &0.86 &0.95 &0.21 &0.91 &0.40 &0.97 &0.90 &0.81 &0.88 &n/a&n/a&n/a&n/a&0.78 & n/a \\
             Maddalena \cite{Maddalena2008a} &0.87 &0.85 &0.91 &0.61 &0.76 &0.42 &0.88 &0.84 &0.58 &0.80 &n/a &n/a&n/a&n/a&0.75 & n/a \\
             \hline
        \end{tabular}
  }
  \label{table:lasiesta}
\end{table*}

\subsection{BMC 2012 dataset}
	The BMC dataset\footnote{\url{http://backgroundmodelschallenge.eu/}}\cite{Vacavant2013} contains 9 videos showing real scenes taken from static cameras  and including the following challenges: shadows, snow, rain, presence of trees or big objects. Three of these sequences are very long (32 965, 117 149 and 107 815 frames).
	 For fair comparison with other published results for this dataset, we  provide the F-measure results for our model obtained using the  the usual F-measure definition described in \ref{implementation}, but also the results obtained using the executable evaluation tool provided with the dataset which does not use the same definition of the F-measure  \cite{Vacavant2013}.	We compute SuBSENSE and PAWCS results on this dataset and provide published evaluation results for other models in \ref{table:BMC2012}.  
	 
	    \begin{table*}
  \centering
  \caption{Comparison of top unsupervised BGS algorithms according to the video F-measure on BMC 2012
  }
  \smallskip
  \scalebox{0.70}{
        \begin{tabular}{l|lllllllll|ll}
            \hline
             Method  & \makecell[l]{Video\\001} &\makecell[l]{Video\\002} &\makecell[l]{Video\\003} &\makecell[l]{Video\\004} &\makecell[l]{Video\\005} &\makecell[l]{Video\\006} &\makecell[l]{Video\\007}  &\makecell[l]{Video\\008} &\makecell[l]{Video\\009} & \makecell[l]{Average\\9 videos}  \\
             \hline 
             F-measure (as defined in equation \ref{equ_f_measure}) &&&&&&&&&&\\
	    AE-NE (ours)  & 0.81&0.72& 0.78 & 0.78 & 0.60 &  0.73 & 0.32 & 0.84 & 0.77 & 0.71  \\
	    
            PAWCS  \cite{St-Charles2016b}  & 0.70 & 0.58 & 0.85 & 0.72 & 0.27 & 0.79 & 0.58 & 0.74  & 0.80 & 0.67\\
            SuBSENSE \cite{St-Charles2015c}& 0.70 & 0.62 & 0.83 & 0.69 & 0.21 & 0.76 & 0.53 & 0.68 & 0.83 & 0.65 \\
             \hline 
             F-measure (using BMC evaluation tool)&&&&&&&&&& \\
             AE-NE (ours) & 0.90 & 0.86 & 0.89 & 0.89 & 0.80 & 0.87 & 0.51 & 0.92 & 0.89 & 0.84  \\
             PAWCS \cite{St-Charles2016b}  & 0.86 & 0.77 & 0.93 & 0.86 & 0.66 & 0.89 & 0.79 & 0.87 & 0.90 & 0.84 \\
             SubSENSE  \cite{St-Charles2015c}& 0.85 & 0.80 & 0.92 & 0.85 & 0.68 & 0.87 & 0.75 & 0.84 & 0.91 & 0.83  \\
             DeepPBM \cite{Farnoosh2019a} & 0.73 & 0.86 & 0.94 & 0.90 & 0.71 & 0.81 & 0.70 & 0.76 & 0.69 & 0.78  \\
             G-LBM  \cite{Rezaei2020} & 0.73 & 0.85 & 0.93 & 0.91 & 0.71 & 0.85 & 0.70 & 0.76 & 0.63 & 0.79  \\
             MSCL-FL \cite{Javed2017a} & 0.84 & 0.84 & 0.88 & 0.90 & 0.83 & 0.80 & 0.78 & 0.85 & 0.94 & 0.86  \\
             B-SSSR \cite{Javed2019} &n/a &n/a&n/a&n/a&n/a&n/a&n/a&n/a&n/a& \bf{0.88} \\
             \hline
        \end{tabular}
  }
  \label{table:BMC2012}
\end{table*}
	 We remark that the F-measure associated to the proposed model is significantly below the state of the art on video  007. This video includes a sequence showing a  train passing on the right lane and occupying a large part of the image (Table \ref{fig:overfitting}, last row). As noted earlier, the proposed model is prone to overfitting when large foreground objects appear to be static  in a video, which is the case here due to the uniform texture of the train and leads to the train being integrated to the background. Despite this  issue, the proposed model gets a competitive average F-measure on the dataset. It also improves upon the state of the art on videos 001 and 008.
	 
 \subsection{Non-video image datasets : Clevrtex, ObjectsRoom, ShapeStacks }
 
The proposed model, which does not use any temporal information, can be adapted to perform background reconstruction and foreground segmentation on some image datasets which are not extracted from video sequences. We have tested this approach on three synthetic images datasets : Clevrtex\footnote{\url{https://www.robots.ox.ac.uk/~vgg/data/clevrtex/}} \cite{Laina2021}, ShapeStacks\footnote{\url{https://ogroth.github.io/shapestacks/}},  \cite{Groth2018}  and ObjectsRoom\footnote{\url{https://github.com/deepmind/multi_object_datasets}} \cite{multiobjectdatasets19}. We use on ShapeStacks and ObjectsRoom the same preprocessing as in \cite{Engelcke2021}\footnote{\url{https://github.com/applied-ai-lab/genesis}}.  Although each image of these datasets shows a different background, the model is able to recognize that all the backgrounds appearing in a given dataset  lie in a low dimensional manifold, which is the case because they have been generated using the same method. These datasets are provided with segmentation annotations for each object appearing in the scenes, which we converted to binary foreground segmentation masks in order to compute the F-measure of the predicted foreground masks. 

We provide in Table \ref {table:non_video_datasets} the average F-measure obtained on the test sets of these datasets and in Figure \ref {fig:clevrtex} some image samples. Considering that on these datasets the risk of overfitting is very low and the background complexity is very high, we  substantially increased the number of iterations, which is set to 500 000. We do not use morphological post-processing on the ShapeStacks and ObjectsRoom datasets, because these images have a very low resolution ($64 \times 64$).
  \begin{table}
  \caption{ F-Measure on the Clevrtex, ShapeStacks and ObjectsRoom datasets }

  \centering
    \scalebox{0.7}{
        \begin{tabular}{lrrrr}
            \hline
            dataset & image size &  \makecell[r]{number of \\ frames \\training set} &  \makecell[r]{number of \\frames \\test set} &  \makecell[r]{average \\F-measure \\on test set}    \\
             \hline 
              Clevrtex  & $128\times128$ & 40000& 5000& 0.78  \\
             ObjectsRoom & $64\times64$ &980000 & 20000 &  0.84 \\
             ShapeStacks & $64\times64$ & 217888 & 46656  & 0.83  \\ 
             \hline
        \end{tabular}
  }
  \label{table:non_video_datasets}
\end{table}
 
 \setlength{\largeur}{16mm}
 \begin{figure}[h]
\centering
  \scalebox{0.7}{
\begin{tabular}{c*{8}{m{18 mm}}}

     \makecell{input \\ frame}  &             \includegraphics[width=\largeur]{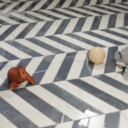} &
                     \includegraphics[width=\largeur]{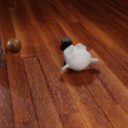} &
   \includegraphics[width=\largeur]{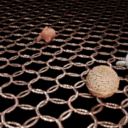} &
    \includegraphics[width=\largeur]{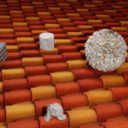} &  \includegraphics[width=\largeur]{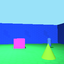} & \includegraphics[width=\largeur]{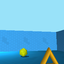} &
     \includegraphics[width=\largeur]{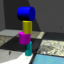} &
      \includegraphics[width=\largeur]{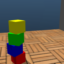} 
    \\             \makecell{ground \\ truth \\ object \\ segmentation} &           \includegraphics[width=\largeur]{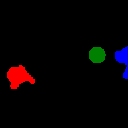} &
                         \includegraphics[width=\largeur]{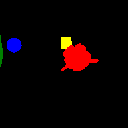}&
  \includegraphics[width=\largeur]{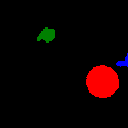} &
   \includegraphics[width=\largeur]{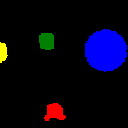} & 
      \includegraphics[width=\largeur]{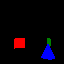} & 
        \includegraphics[width=\largeur]{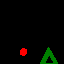}  &
         \includegraphics[width=\largeur]{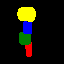} & 
        \includegraphics[width=\largeur]{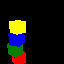}  
   \\

        \makecell{predicted \\ background}  &                 \includegraphics[width=\largeur]{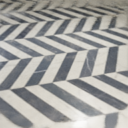}  &
                         \includegraphics[width=\largeur]{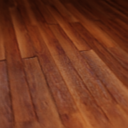} &
 \includegraphics[width=\largeur]{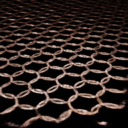}  &
 \includegraphics[width=\largeur]{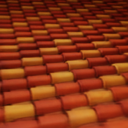} & 
  \includegraphics[width=\largeur]{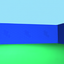}  &
 \includegraphics[width=\largeur]{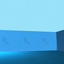} &
   \includegraphics[width=\largeur]{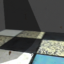}  &
 \includegraphics[width=\largeur]{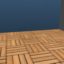}
  \\

   \makecell{predicted \\ foreground \\ mask}   &         \includegraphics[width=\largeur]{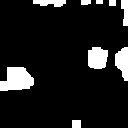} &
             \includegraphics[width=\largeur]{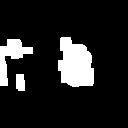}  &
  \includegraphics[width=\largeur]{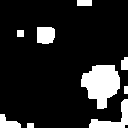}  &
 \includegraphics[width=\largeur]{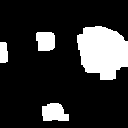}  &
  \includegraphics[width=\largeur]{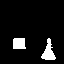}  &
   \includegraphics[width=\largeur]{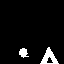} &
     \includegraphics[width=\largeur]{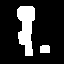}  &
   \includegraphics[width=\largeur]{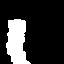} 
  \\
\end{tabular}}
\caption{Examples of background reconstruction and foreground segmentations on the datasets Clevrtex (columns 1-4),  ObjectsRoom (columns 5-6)  and ShapeStacks (columns 7-8)}
\label{fig:clevrtex}
\end{figure}

\subsection {Computation time}
 We provide in  Table \ref{table:computation_time} some computation time measurements, obtained using a desktop computer with an Intel Core i7 7700K@4,2GHz CPU  and a Nvidia RTX 2080 TI GPU.

 \begin{table}
  \caption{Computation time of the proposed model, PAWCS and SubSENSE for some sequences of the CDnet and BMC datasets }
  \smallskip
  \centering
    \scalebox{0.75}{
        \begin{tabular}{lrrrrr}
            \hline
            sequence  name& highway & Video&blizzard  & zoomin& continuous  \\
            & & 009&  & zooomout & pan  \\

           image size &  240x320  &  288x352 &480x720  & 240x320 &  480x704 \\
           number of frames &1700 & 107817 & 7000 & 1130 & 1700 
            \\
           background complexity & simple &simple & simple&  complex &complex \\
             \hline 
              \multicolumn{2}{l}{ computation times (seconds) } \\
             AE-NE (proposed model) \\
             \quad   - training  & 173 &  197&783 & 2716 & 14221  \\
             \quad  - backgrounds \\
             \quad and masks generation & 7 & 578 & 125 & 5 & 37 \\ 
            \quad  - total  & 180 & 775  & 908 & 2721 & 14258   \\
             SuBSENSE   & 98 & 7093 & 1333 & 68 & 393   \\
             PAWCS  & 182 & 13021& 2418 &  150 & 1013  \\
             \hline
        \end{tabular}
  }
  \label{table:computation_time}

\end{table}
  
The training time of the autoencoder mainly depends on the size of the input images and the complexity of the background and  is not proportional to the number of frames of the dataset, which makes this model  attractive for long frame sequences. The model is indeed faster than PAWCS and SuBSENSE on long videos showing simple backgrounds.  On short sequences with complex backgrounds, PAWCS and SuBSENSE are faster, but fail  to predict accurate foreground masks (cf Table \ref{table:cdnet}, PTZ category and Table   \ref{table:lasiesta}, categories IMC, ISM, OMC, OSM).

\subsection{Ablation study}

In order to assess the impact of the various model features described in this paper, we have implemented several modifications of the proposed model and measured the average F-measure (FM) of these models on the CDnet2014 dataset. The results of these experiments are provided in  Table \ref{table:ablation}.

They show that the design of the loss function and the use of the background noise estimation layer have a substantial impact on the accuracy of the model. More precisely, performing background reconstruction without using the background noise estimation has a very negative impact on the categories dynamic background (FM:  0.078), turbulence (FM: 0.259), pan-tilt-zoom (FM: 0.432) and low frame-rate (FM: 0.474), but the other categories of the dataset are not significantly impacted by this modification. 

The improvement associated to post-processing is also significant, as already observed for other unsupervised background subtraction methods \cite{Shahbaz2015}.

The model remains competitive on CDnet if the background complexity of all frames sequence is set to simple, an option which may  be considered if computation time is an issue and it is known that the camera is fixed.

  \begin{table}
  \centering
  \caption{Evaluation of various ablations of the proposed model}
  \smallskip
  \scalebox{0.70}
  {
        \begin{tabular}{lll}
            \hline
            model description & average  & evolution vs \\
              &  F-measure  on the & reference \\
            &  CDnet dataset &   model \\
             proposed model (reference) & 0.7841  \\
             modified models : \\
	   - no bootstrap weights ($ w_{n,i,j}^{\text{bootstrap}}$  set to 1) & 0.2771 & -64,6 \%  \\
	    - inference without using the background noise \\
	    estimation ($\alpha_2$ set to  0) & 0.6220 & -20.7 \% \\
	     - $ w_{n,i,j}^{\text{bootstrap}}$  set to 1 and $\alpha_2$ set to  0 & 0.4557 & -41,9\%  \\
	     - training with L2 reconstruction loss, $\alpha_2$ set to  0 &0,3384 & -56,8 \% \\
	        - inference without  morphological post-processing & 0.7170 &  -8.5\% \\
	   - all backgrounds are considered as simple ($\tau_0$ set to 1)  & 0,7397 & -5,6 \% \\
             \hline
        \end{tabular}
  }
  \label{table:ablation}
\end{table}
\subsection {Image samples}
We provide  in Figure \ref{figure:example}  some samples of background reconstruction, with the associated predicted foreground mask, and a comparison with foreground masks obtained using PAWCS and SuBSENSE. Other samples are provided in the Appendix.

\setlength{\largeur}{25 mm}
\begin{figure*}
\centering
  \scalebox{0.60}{
\begin{tabular}{*{6}{m{26 mm}}}

       \includegraphics[width=\largeur]{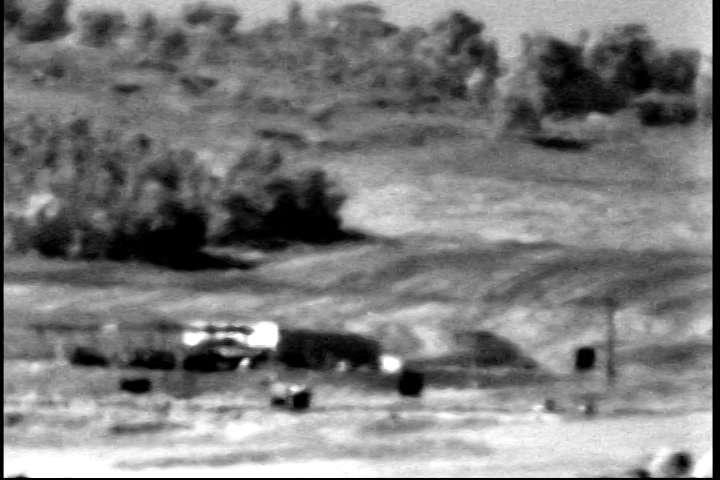} &
 \includegraphics[width=\largeur]{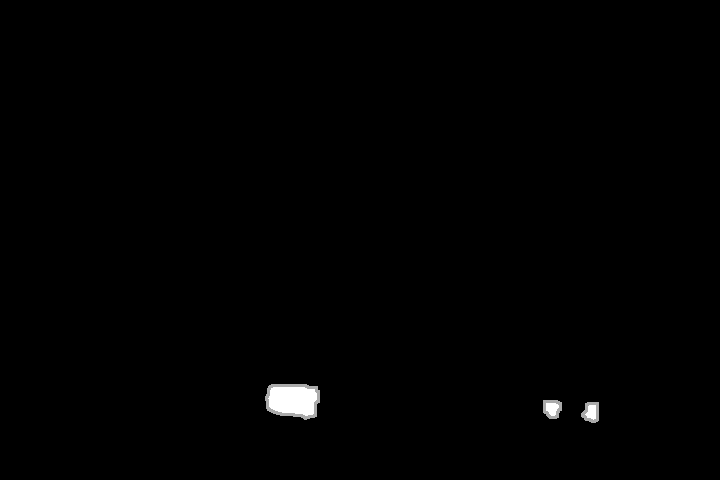} &
  \includegraphics[width=\largeur]{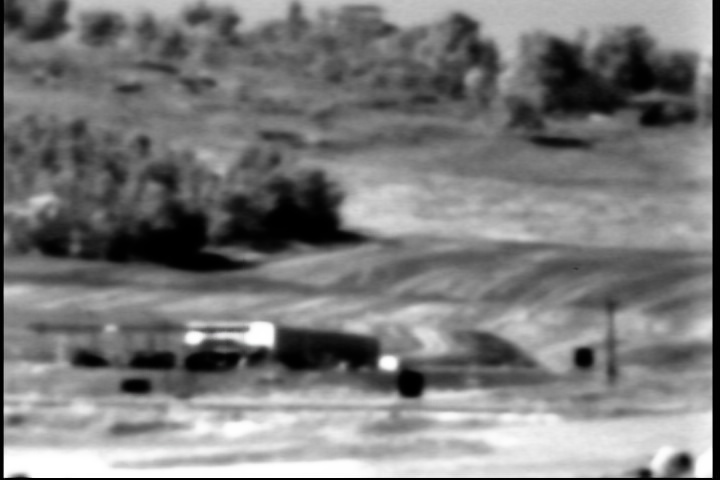} &
    \includegraphics[width=\largeur]{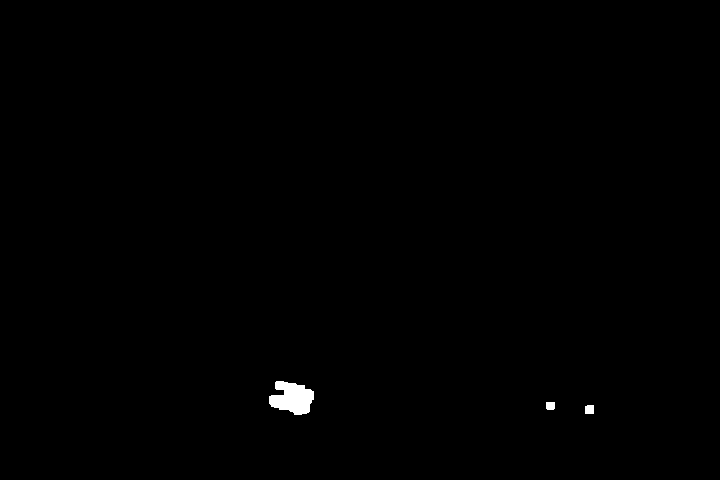} &
 \includegraphics[width=\largeur]{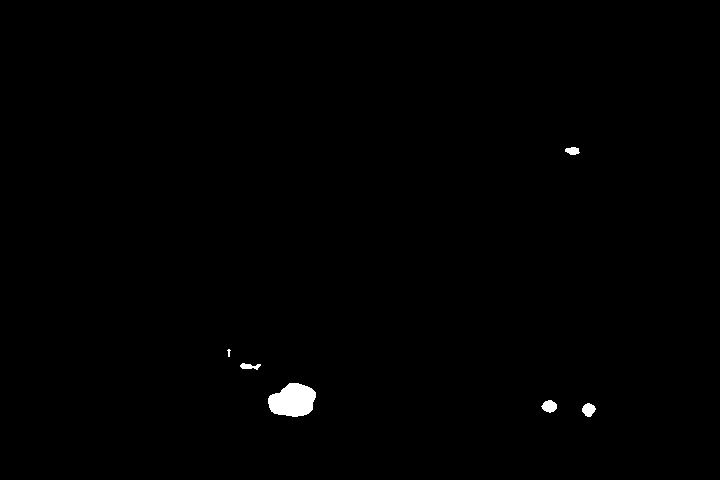} &
  \includegraphics[width=\largeur]{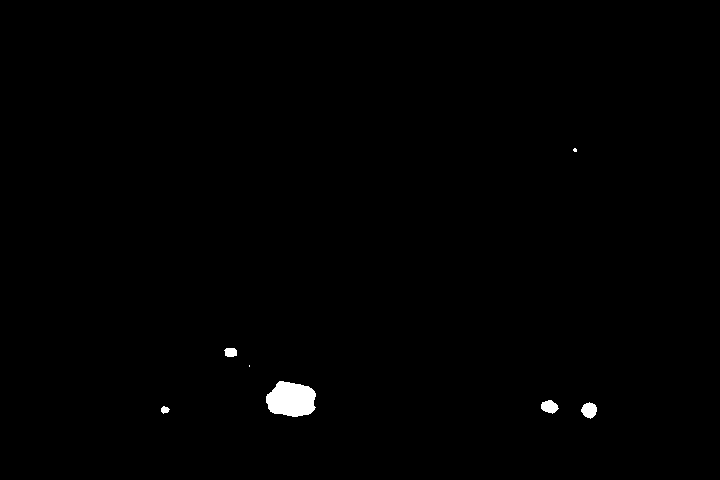} 
  \\

     \includegraphics[width=\largeur]{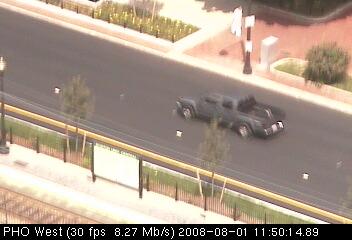} &
 \includegraphics[width=\largeur]{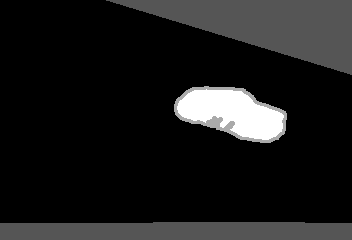} &
  \includegraphics[width=\largeur]{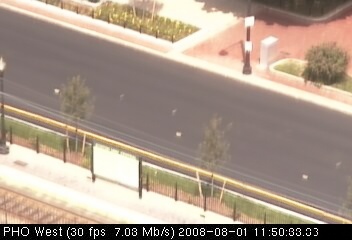} &
    \includegraphics[width=\largeur]{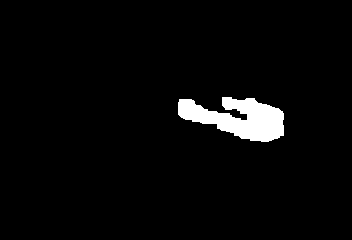} &
 \includegraphics[width=\largeur]{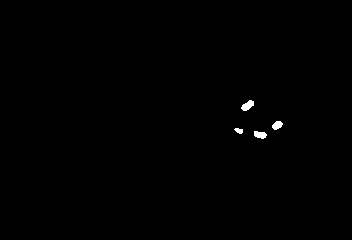} &
  \includegraphics[width=\largeur]{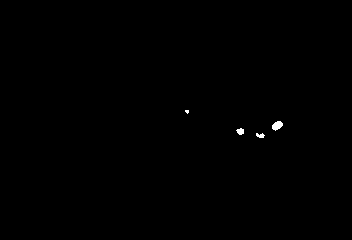} 
  \\

 \includegraphics[width=\largeur]{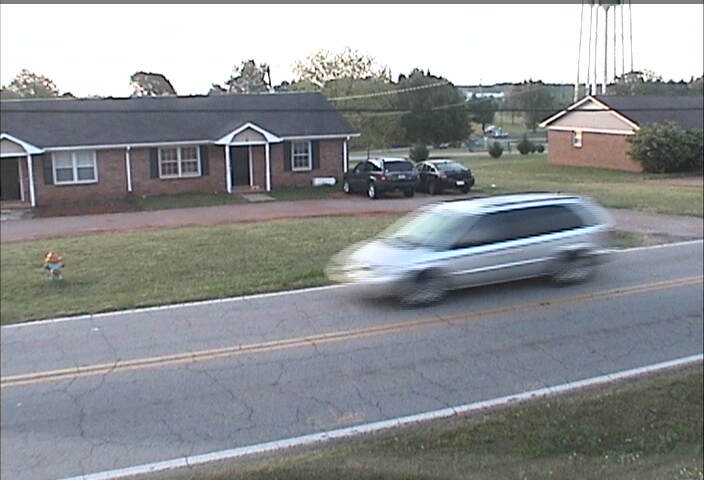} &
 \includegraphics[width=\largeur]{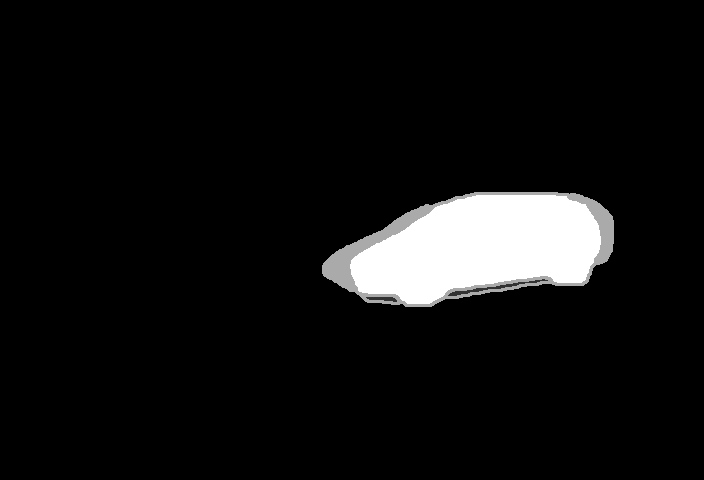} &
  \includegraphics[width=\largeur]{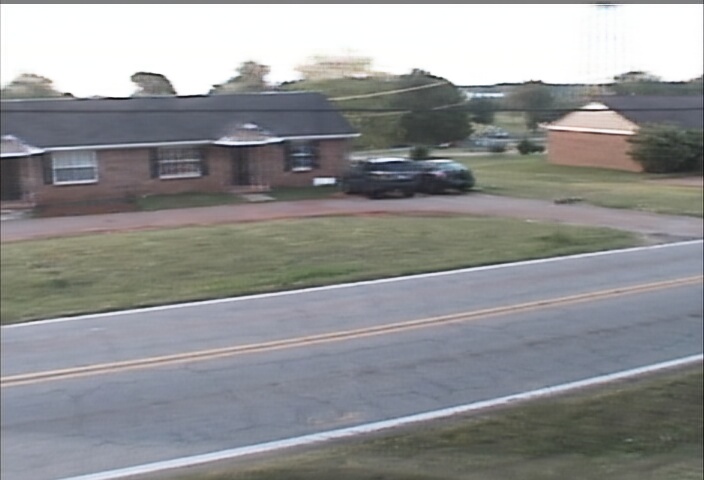} &
    \includegraphics[width=\largeur]{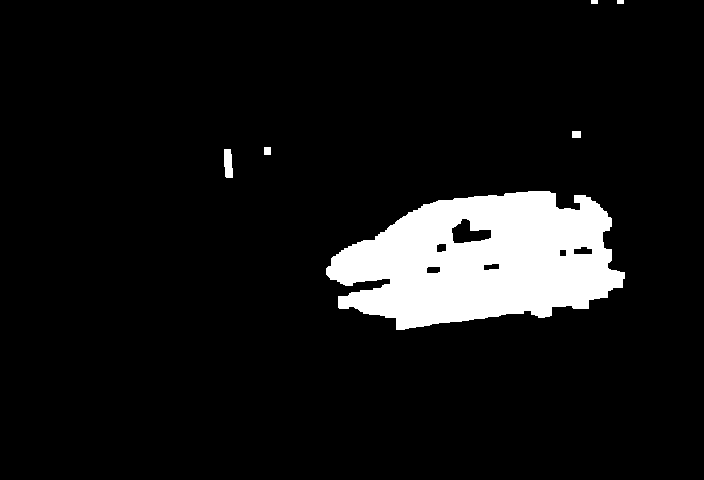} &
 \includegraphics[width=\largeur]{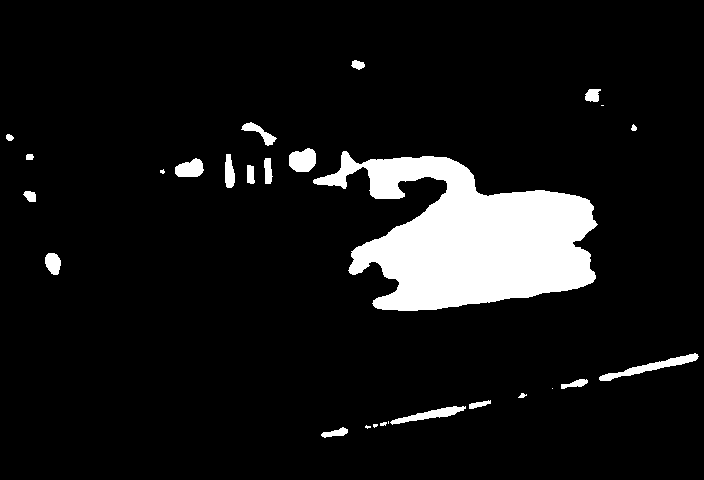} &
  \includegraphics[width=\largeur]{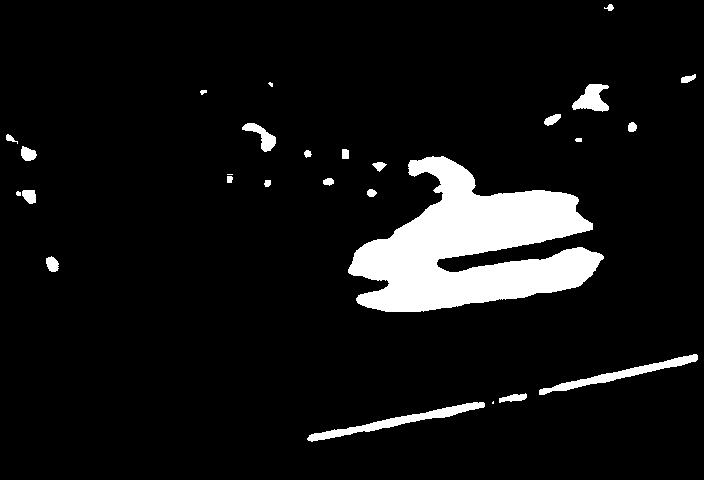} 
  \\

             \includegraphics[width=\largeur]{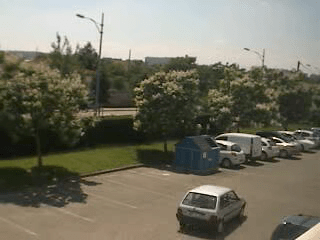} &
    \includegraphics[width=\largeur]{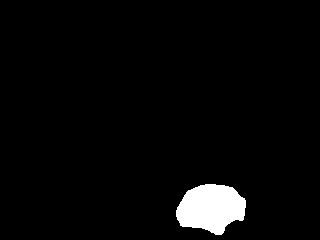} &
  \includegraphics[width=\largeur]{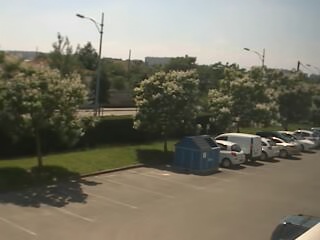} &
    \includegraphics[width=\largeur]{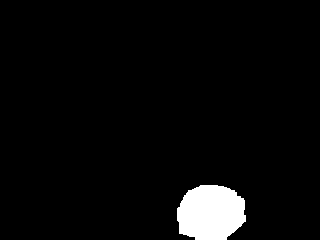} &
 \includegraphics[width=\largeur]{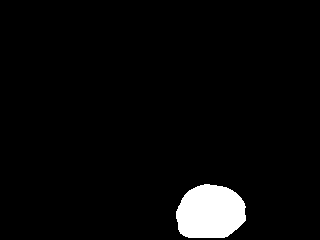} &
  \includegraphics[width=\largeur]{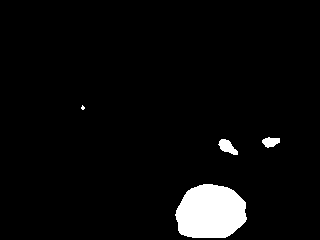}
  \\

    \centering  \makecell{input\\frame}  &   \makecell{foreground \\ mask \\ ground truth}  &   \makecell{predicted \\background \\AE-NE (ours)}  &  \makecell{predicted \\foreground mask \\AE-NE (ours)} &   \makecell{predicted \\ foreground mask  \\PAWCS} & \makecell{predicted \\ foreground mask  \\SuBSENSE}  
\end{tabular}}
\caption{Examples of background reconstruction and foreground segmentation produced using the proposed model and comparison with PAWCS and SuBSENSE}
\label{figure:example}
\end{figure*}
\section {Conclusion}
We have proposed in this paper a new fully unsupervised dynamic background reconstruction and foreground segmentation model which does not use any temporal or motion information. It is on average more accurate than available unsupervised models for background subtraction, significantly improves upon the state of the art on videos taken from a moving camera and is able to perform background reconstruction on some non-video image datasets.

Future works include using the proposed model to perform unsupervised object detection on real world scenes with complex backgrounds.

%\bibliographystyle{splncs04}

\bibliographystyle{ieee_fullname}

%\bibliography{/Users/brunosauvalle/Documents/travailmines/These/bibtex/dynamic_background_reconstruction_paper}
\bibliography{dynamic_background_reconstruction_paper}

\section{Appendix}
\subsection{Autoencoder architecture}
The autoencoder is deterministic and takes as input a  RGB image of size $h\times w$, and produces  a RGB image (3 channels) and an error estimation map of the same size (1 channel).

The encoder and decoder structures in the proposed model are computed dynamically using as input the size (height $h$ and width $w$) of the input frames of the dataset. The number of latent variables produced by the encoder is fixed to 16.

We use a fully convolutional autoencoder architecture, which appears to be more robust to overfitting than architectures including fully connected layers or locally connected layers. We  add two  fixed positional encoding channels as inputs to all layers of the encoder and the decoder, one channel coding for the horizontal coordinates, the other one for the vertical coordinates .

The encoder is a sequence of blocks composed of a convolution layer with kernel size 5, stride 3 and padding equal to 2, followed by a group normalization layer and a CELU nonlinearity layer.  The generator is a symmetric sequence of blocks composed of transpose convolution layers with kernel size 5 and stride 3 and padding equal to 2 followed by group normalization and a CELU nonlinearity, except for the last layer where the transpose convolution layer is followed by a sigmoid to generate the final image. The number of layers of the encoder and the decoder is then equal to 5 or 6 depending on the image size (assuming that the maximum of the image height and image width is in the range $200-1000$).
The number of channels per convolutional layer is fixed according to Table \ref{table:channel}, depending on the image size and the background complexity.
\begin{table}[h]
  \centering
  \caption{Number of channels for each layer of the encoder and decoder  (excluding positional encoding input channels)
  }
  \smallskip
  \scalebox{0.80}{
        \begin{tabular}{l|l|ll}
        
            \hline
            \makecell[l]{background\\complexity} &\makecell[l]{ image \\size \\ max(h,w)}   & Encoder & Decoder \\
             \hline

	     simple & 200-405  & (3,64,160,160,32,16) & (16,32,256,256,144,4) \\
	     simple & 406-1000  & (3,64,160,160,160,32,16) & (16,32,256,512,256,144,4) \\

	     complex & 200-405  & (3,64,160,160,16,16) & (16,16,640,640,144,4) \\
	     complex & 406-1000     &   (3,64,160,160,160,16,16) & (16,16,640,1280,640,144,4) \\
             \hline
        \end{tabular}
  }
  \label{table:channel}

\end{table}

These channel distributions are motivated by the fact  that a larger number of parameters is required in the generator in order to handle complex backgrounds, but that we have experimentally observed that a large number of channels in the last layer of the encoder and the first layer of the decoder increases the risk of overfitting  on foreground objects, so that reducing this number for long training schedule is necessary to  improve the robustness of the auto-encoder with respect to the risk of overfitting.  For example, we have measured that increasing the numbers of channels  in the last hidden layer of the encoder and first hidden layer of the decoder  to 160 and 256  leads to de  2,3 \%  degradation of the average F-Measure on the CDnet dataset. 

For non-video dataset experiments, which handle small images, we use a smaller stride, set to 2 instead of 3. The autoencoder architectures for $64 \times 64$ images (ShapeStacks and ObjectRooms datasets) and $128 \times 128$ images (Clevrtex dataset) are described in Table \ref{table:64images} and \ref{table:128images}: 

\begin{table}
  \caption{autoencoder architecture for $64 \times 64$ images
  }

    \scalebox{0.60}{
        \setlength{\tabcolsep}{2.mm}{
            \begin{tabular}{ccccc}
           \multicolumn{5}{c}{Encoder}  \tabularnewline
            \tabularnewline
            \toprule
            Layer           & Size & Ch & Stride & Norm./Act.   \tabularnewline \hline \hline
            Input           & 64       &    3    &              \tabularnewline
            Conv $5\times5$ & 32   & 64  & 2      & GroupNorm/CELU \tabularnewline
            Conv $5\times5$ & 16  & 160    & 2      & GroupNorm/CELU \tabularnewline
            Conv $5\times5$ & 8   & 320   & 2      & GroupNorm/CELU \tabularnewline
            Conv $5\times5$ & 4   & 160   & 2      & GroupNorm/CELU \tabularnewline
            Conv $4\times4$ & 2   & 16   & 2      & GroupNorm/CELU \tabularnewline
            Conv $2\times2$ & 1    & 16  & 1      & GroupNorm/CELU \tabularnewline
            \bottomrule
            \end{tabular}
            \quad
            \begin{tabular}{ccccc}
               \multicolumn{5}{c}{Decoder}  \tabularnewline
            \tabularnewline
            \toprule
            Layer           & Size & Ch &  Stride & Norm./Act.   \tabularnewline \hline \hline
            
            Input & 1      & 16      & &  \tabularnewline
            Conv Transp $2\times2$ & 2   &  16 & 1      & GroupNorm/CELU \tabularnewline
            Conv Transp $4\times4$ & 4    & 640 & 2      & GroupNorm/CELU \tabularnewline
            Conv Transp $5\times5$ & 8    & 1280 & 2     & GroupNorm/CELU \tabularnewline
           Conv Transp $5\times5$ & 16  &   640 & 2      & GroupNorm/CELU \tabularnewline
            Conv Transp $5\times5$ & 32  &   144 & 2      & GroupNorm/CELU \tabularnewline
            Conv Transp $5\times5$ & 64  &   4 & 2      &  \tabularnewline
            Sigmoid & 64 & 4 &\tabularnewline
            \bottomrule
            \end{tabular}
        }
    }
  \label{table:64images}
\end{table}
\begin{table}
  \caption{autoencoder architecture  for $128 \times 128$ images}
    \scalebox{0.60}{
        \setlength{\tabcolsep}{2.mm}{
            \begin{tabular}{ccccc}
             \multicolumn{5}{c}{Encoder}  \tabularnewline
            \tabularnewline
            \toprule
            Layer           & Size & Ch & Stride & Norm./Act.   \tabularnewline \hline \hline
            Input           & 128       &    3    &              \tabularnewline
            Conv $5\times5$ & 64   & 64  & 2      & GroupNorm/CELU \tabularnewline
            Conv $5\times5$ & 32   & 320  & 2      & GroupNorm/CELU \tabularnewline
            Conv $5\times5$ & 16  & 640    & 2      & GroupNorm/CELU \tabularnewline
            Conv $5\times5$ & 8   & 640  & 2      & GroupNorm/CELU \tabularnewline
            Conv $5\times5$ & 4   & 320   & 2      & GroupNorm/CELU \tabularnewline
            Conv $4\times4$ & 2   & 16   & 2      & GroupNorm/CELU \tabularnewline
            Conv $2\times2$ & 1    & 16  & 1      & GroupNorm/CELU \tabularnewline  
            \bottomrule
            \end{tabular}
\quad
            \begin{tabular}{ccccc}
            \multicolumn{5}{c}{Decoder}  \tabularnewline
            \tabularnewline
            \toprule
            Layer           & Size & Ch &  Stride & Norm./Act.   \tabularnewline \hline \hline
            
            Input & 1      & 16      & & \tabularnewline
            Conv Transp $2\times2$ & 2   &  16 & 1      & GroupNorm/CELU \tabularnewline
            Conv Transp $4\times4$ & 4    & 320 & 2      & GroupNorm/CELU \tabularnewline
            Conv Transp $5\times5$ & 8    & 640 & 2     & GroupNorm/CELU \tabularnewline
           Conv Transp $5\times5$ & 16  &   1280 & 2      & GroupNorm/CELU \tabularnewline
            Conv Transp $5\times5$ & 32  &   640 & 2      & GroupNorm/CELU \tabularnewline
            Conv Transp $5\times5$ & 64  &   144 & 2      & GroupNorm/CELU \tabularnewline
            Conv Transp $5\times5$ & 128  &   4 & 2      &  \tabularnewline
            Sigmoid &128 & 4  &\tabularnewline 
            \bottomrule
            \end{tabular}
        }
    }
\label{table:128images} 
\end{table}
%\vspace{-2em}
\subsection{Implementation details}

The proposed model is implemented using Python and the Pytorch framework. The associated code will be made available on the Github platform. Optimization is performed using the Adam optimizer  with a learning rate of $5. 10^{-4}$ and batch size equal to 32. The learning rate is divided by 10 when the number of optimization iterations reaches 80\% of the total number of iterations.  The same set of hyperparameters is used for  the experiments on CDnet, LASIESTA and BMC datasets, i.e.  . $\beta=6$, $r=75$, $\tau_0 = 0.24$, $\tau_1 = 0.25$, $\alpha_1 = 96/255 $, $\alpha_2 = 7$, $N_{\text{eval}} = 2000$, $B_{\text{eval}} = 480$, $N_{\text{simple}} = 2500$, $N_{\text{complex}} = 24000$, $E_{\text{complex}} = 20$. These hyperparameter values as well as the channel distributions described in Table \ref{table:channel} were found empirically to give good results, although a full hyperparameter and architecture search has not been performed and is beyond the scope of this paper.

For non-video dataset experiments, which take small images ($64 \times 64$ and $128 \times 128$) as inputs, the batch size and learning rate are   increased to 128 and $2. 10^{-3}$, the number of iterations $N_{\text{complex}}$ is set to 500 000 and no morphological post-processing is performed on $64 \times 64$ images. The other hyperparameters remain the same.

\subsection{Additional image samples}

We provide in figures $6-12$
 additional samples of background reconstruction and foreground segmentation obtained using the proposed model.

\setlength{\largeur}{25 mm}

\begin{figure*}
\centering
  \scalebox{0.70}{
\begin{tabular}{*{6}{m{26 mm}}}

         \includegraphics[width=\largeur]{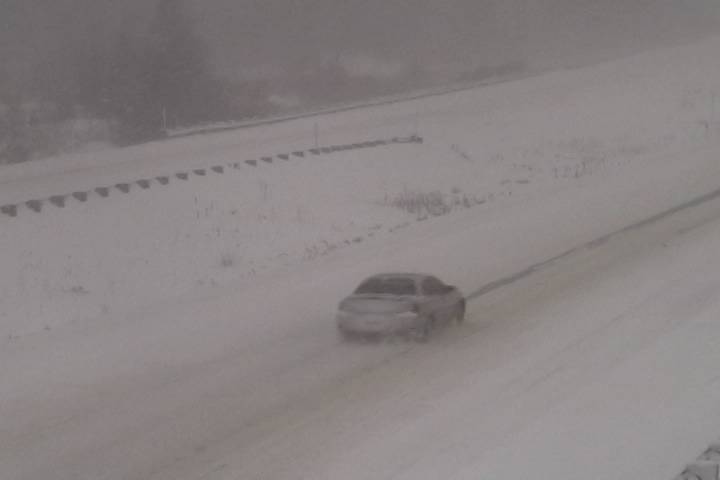} &
 \includegraphics[width=\largeur]{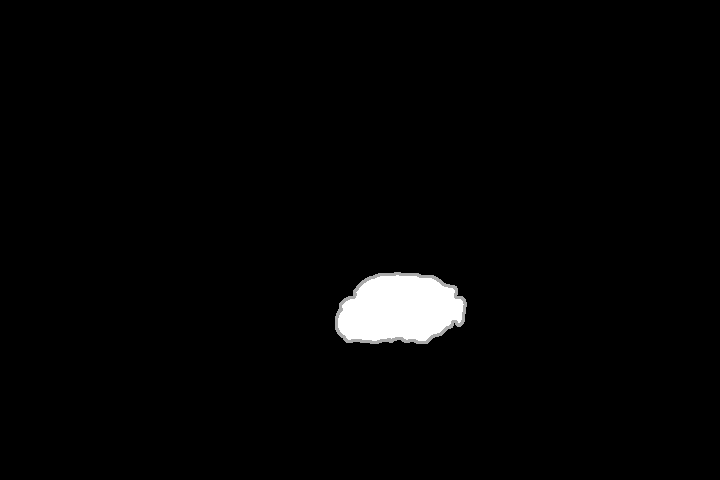} &
  \includegraphics[width=\largeur]{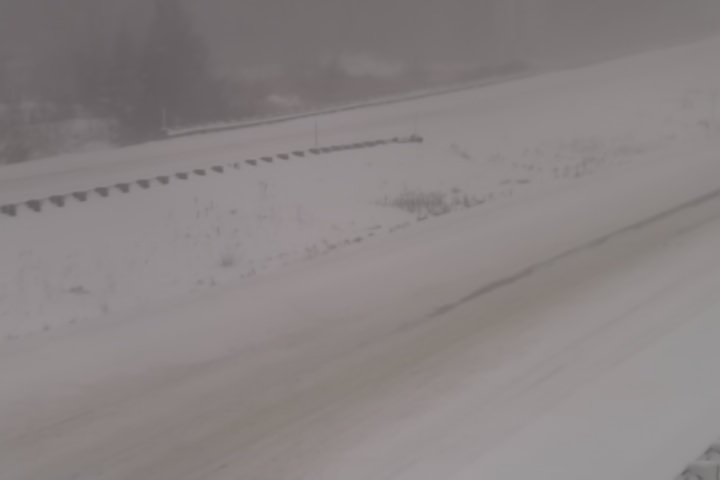} &
    \includegraphics[width=\largeur]{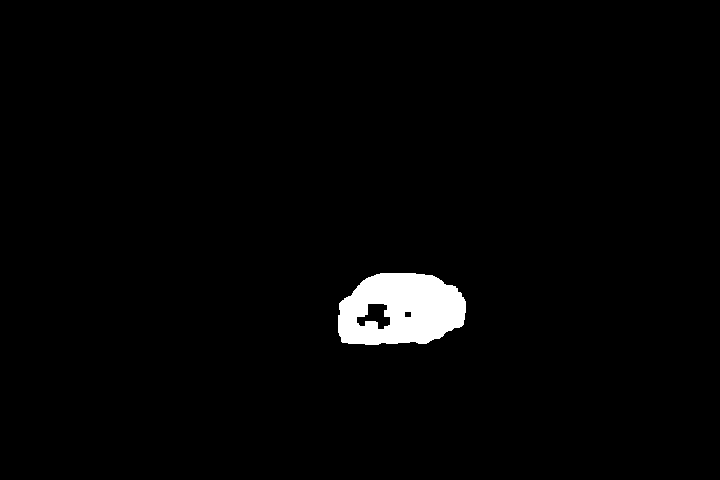} &
 \includegraphics[width=\largeur]{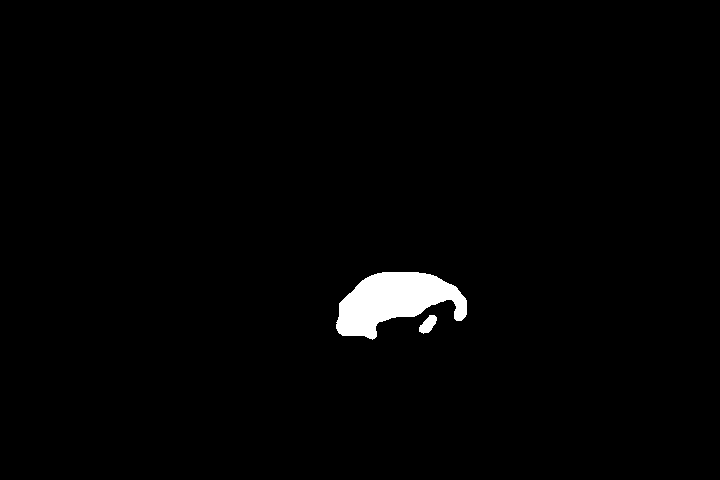} &
 \includegraphics[width=\largeur]{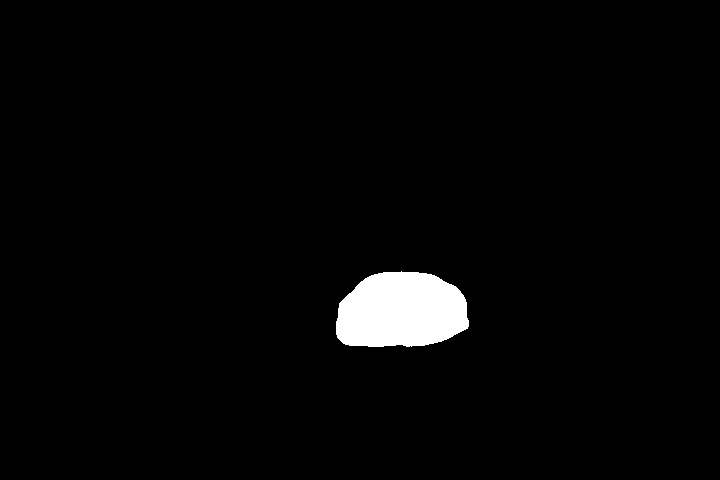} 
  \\

           \includegraphics[width=\largeur]{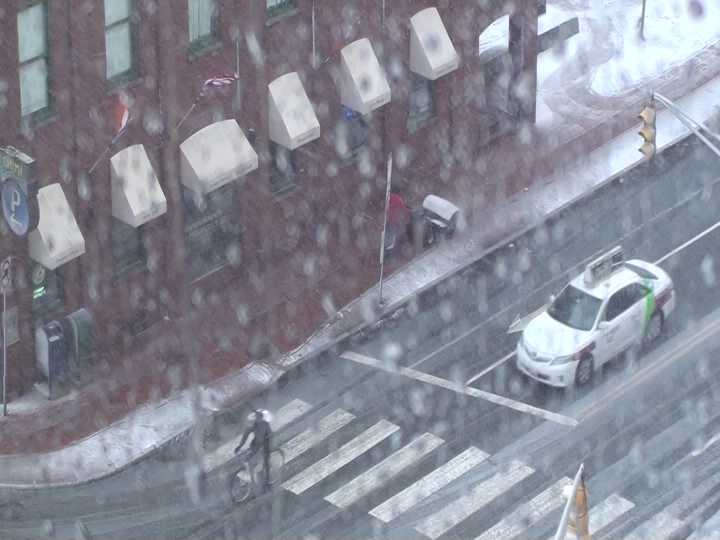} &
 \includegraphics[width=\largeur]{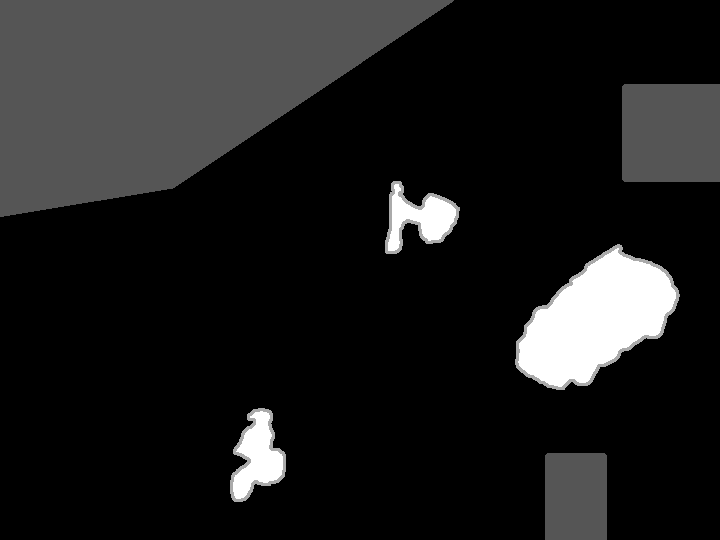} &
 \includegraphics[width=\largeur]{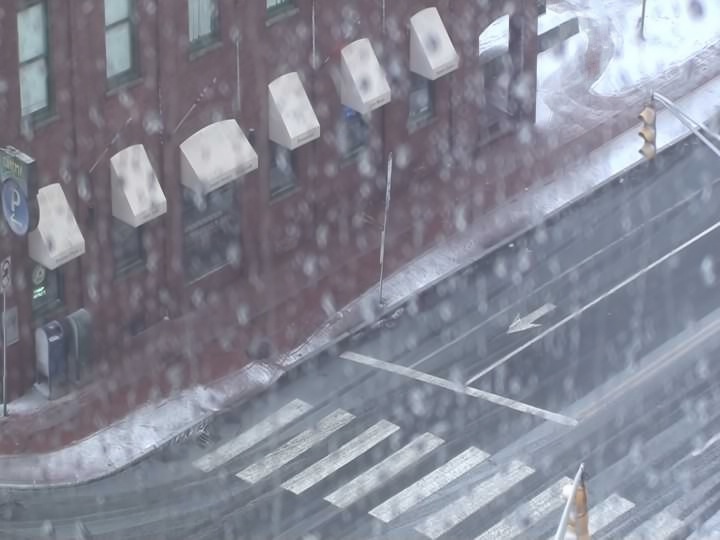} &
    \includegraphics[width=\largeur]{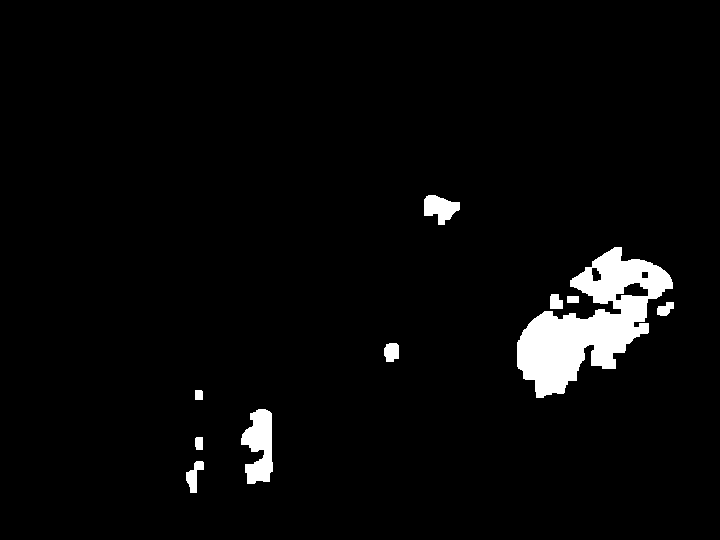} &
 \includegraphics[width=\largeur]{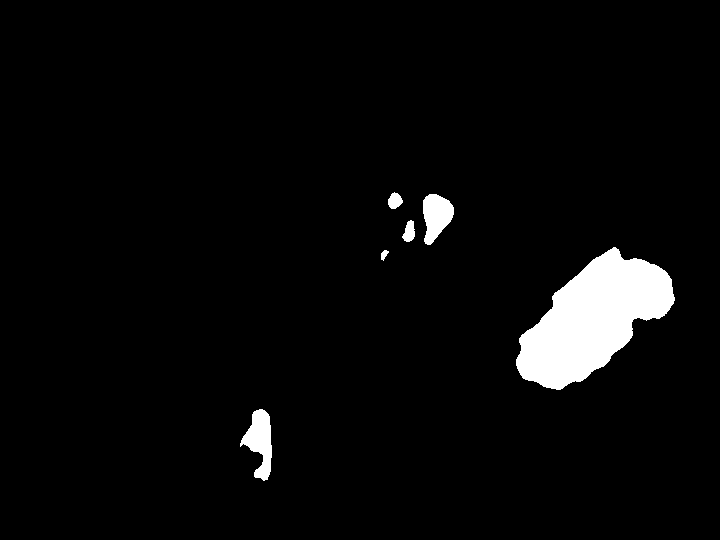} &
 \includegraphics[width=\largeur]{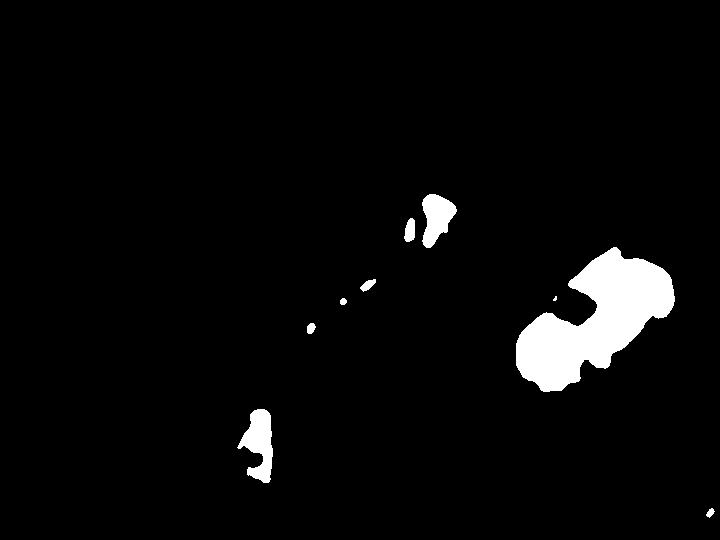} 
  \\
  
           \includegraphics[width=\largeur]{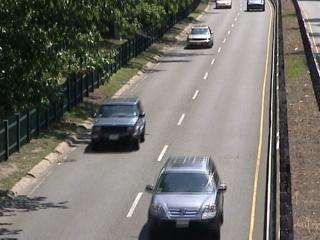} &
 \includegraphics[width=\largeur]{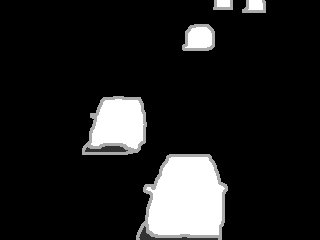} &
 \includegraphics[width=\largeur]{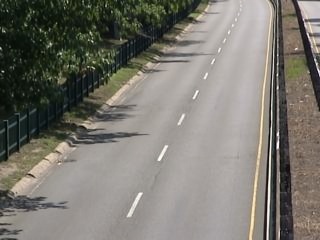} &
    \includegraphics[width=\largeur]{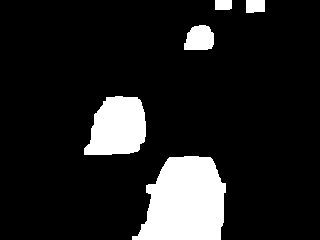} &
 \includegraphics[width=\largeur]{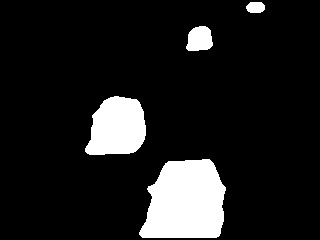} &
 \includegraphics[width=\largeur]{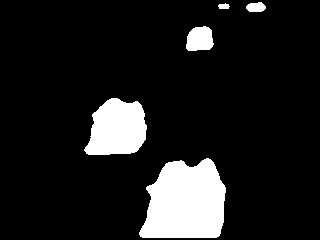} 
  \\				

           \includegraphics[width=\largeur]{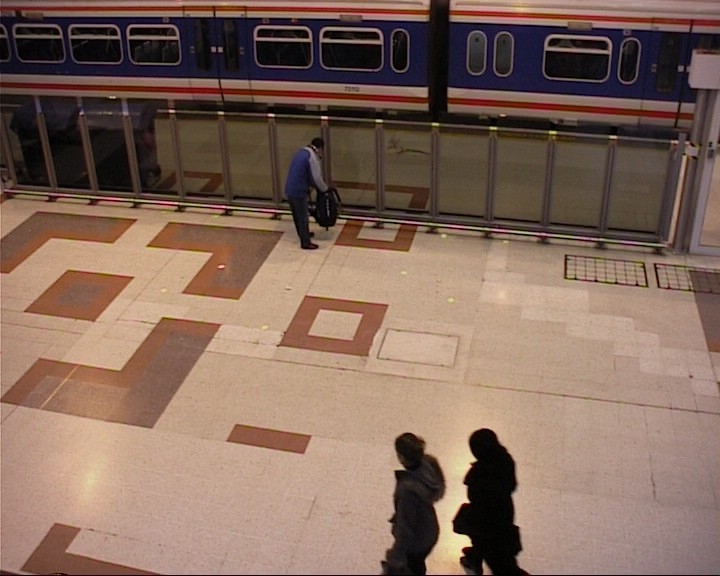} &
 \includegraphics[width=\largeur]{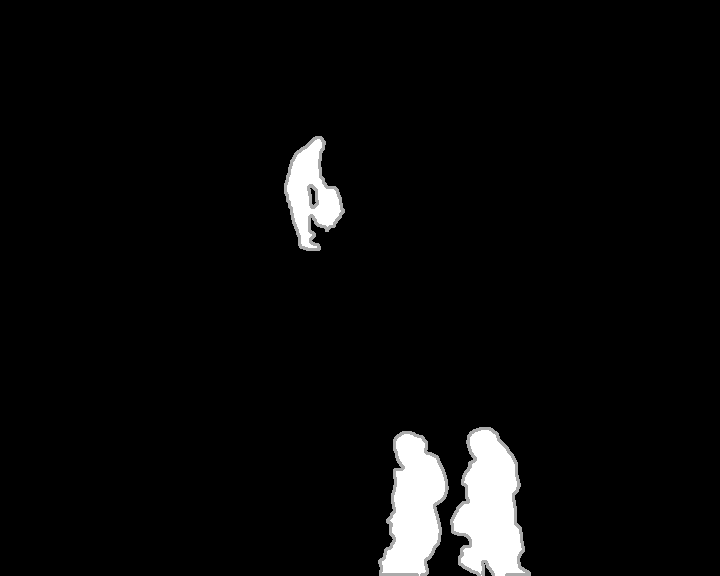} &
 \includegraphics[width=\largeur]{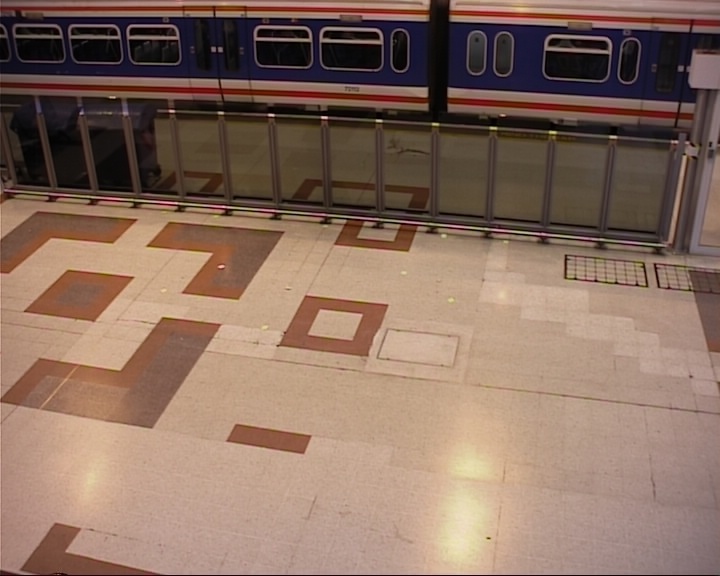} &
    \includegraphics[width=\largeur]{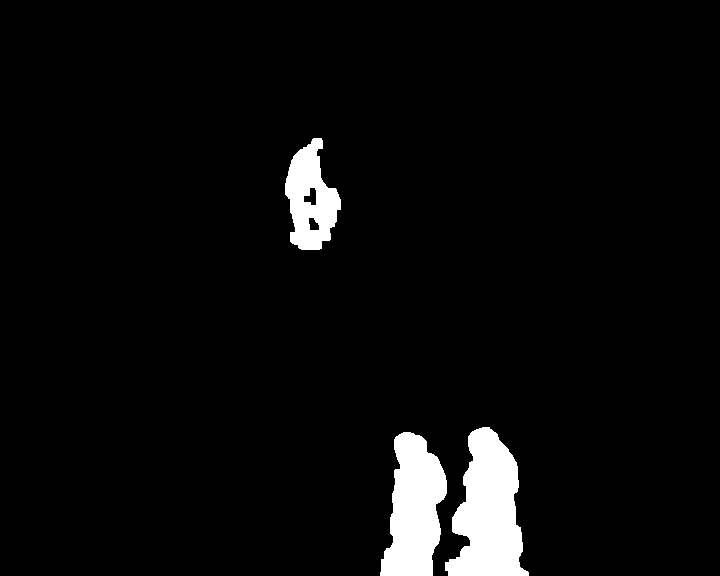} &
 \includegraphics[width=\largeur]{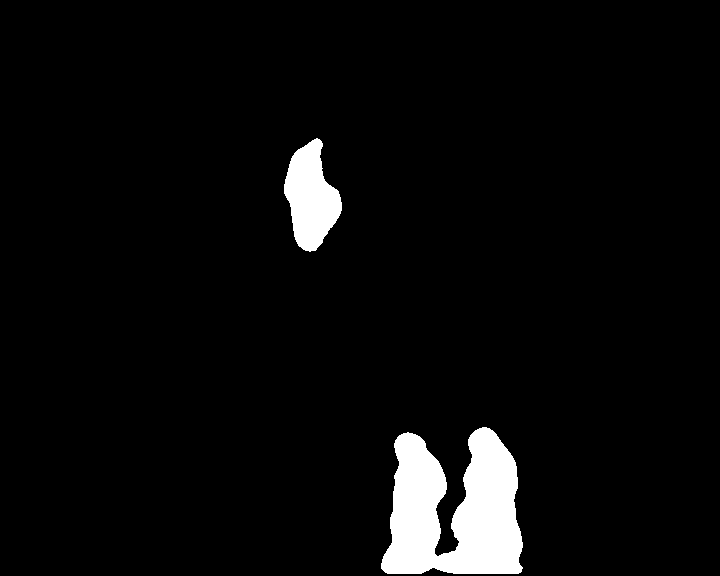} &
 \includegraphics[width=\largeur]{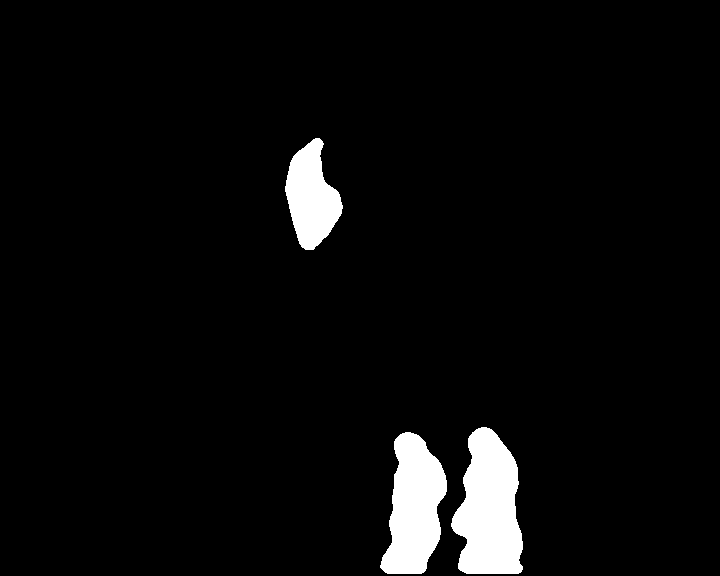} 
  \\

            \includegraphics[width=\largeur]{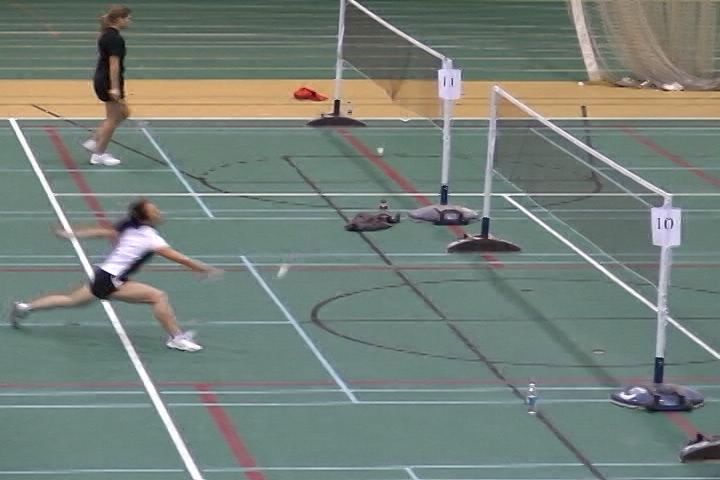} &
 \includegraphics[width=\largeur]{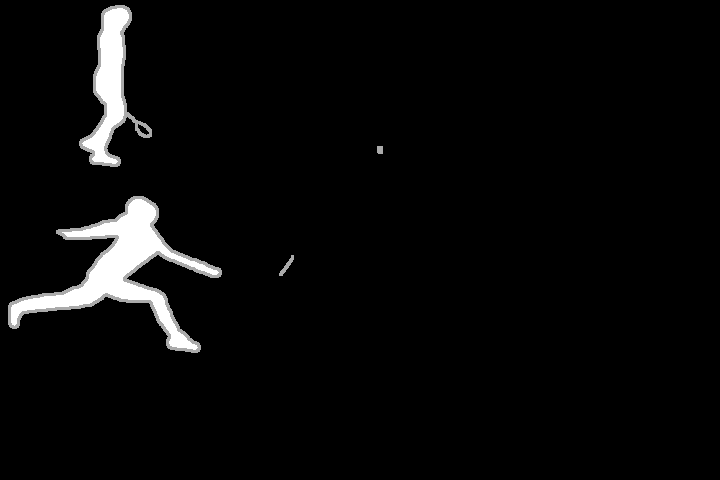} &
 \includegraphics[width=\largeur]{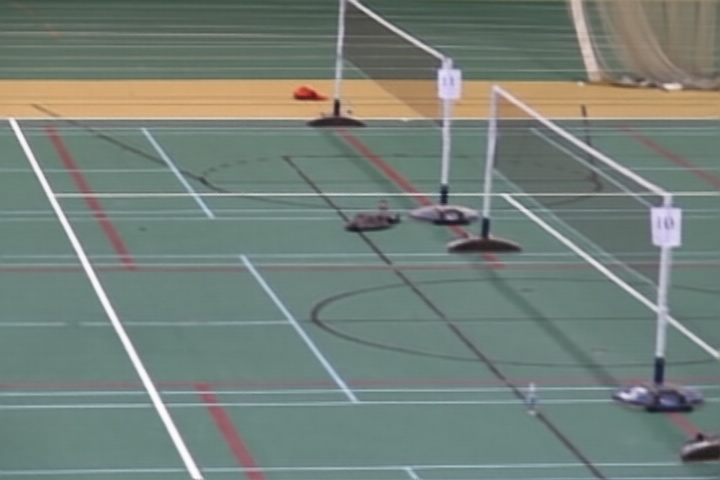} &
    \includegraphics[width=\largeur]{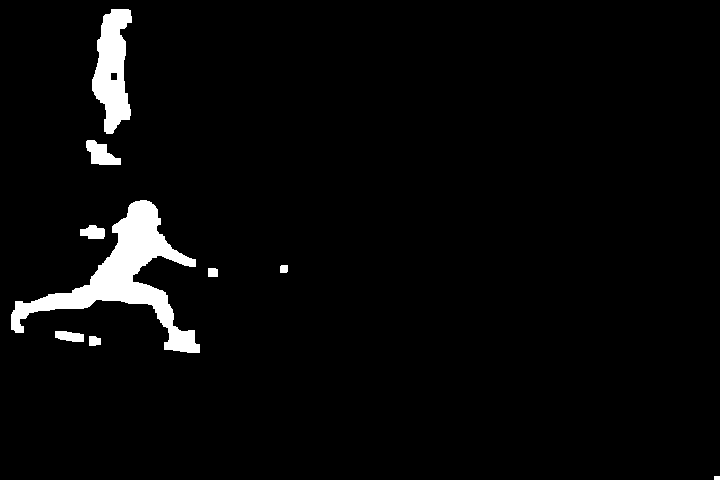} &
 \includegraphics[width=\largeur]{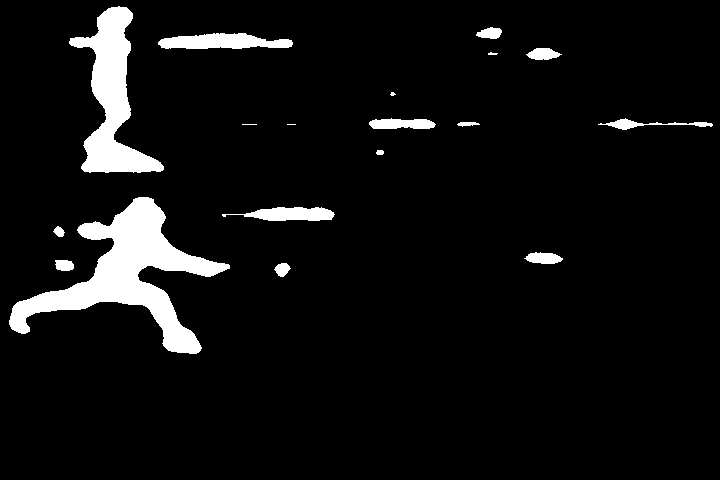} &
 \includegraphics[width=\largeur]{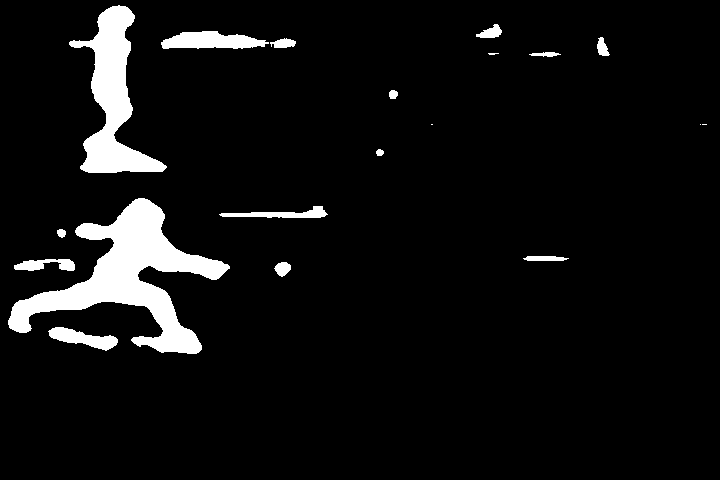} 
  \\
  
    \centering  \makecell{input\\frame}  &   \makecell{foreground \\ mask \\ ground truth}  &   \makecell{predicted \\background \\AE-NE (ours)}  &  \makecell{predicted \\foreground mask \\AE-NE (ours)} &   \makecell{predicted \\ foreground mask  \\PAWCS} & \makecell{predicted \\ foreground mask  \\SuBSENSE}  
\end{tabular}}
\caption{Examples of background reconstruction and foreground segmentation on the CDnet 2014 dataset produced using the proposed model and comparison with PAWCS and SuBSENSE}
\label{figure:cdnet1}
\end{figure*}

\begin{figure*}
\centering
  \scalebox{0.70}{
\begin{tabular}{*{6}{m{26 mm}}}

              \includegraphics[width=\largeur]{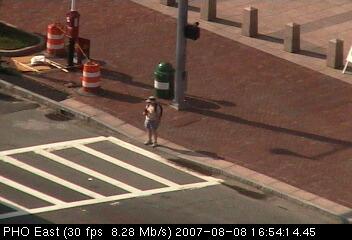} &
 \includegraphics[width=\largeur]{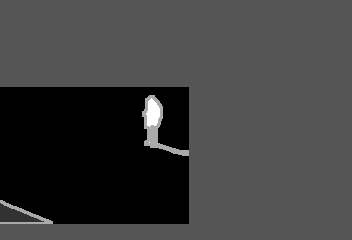} &
 \includegraphics[width=\largeur]{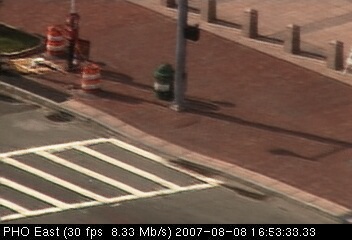} &
    \includegraphics[width=\largeur]{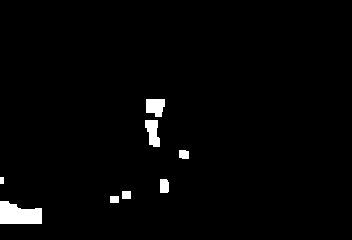} &
 \includegraphics[width=\largeur]{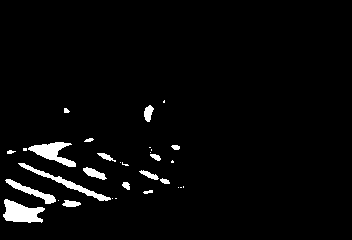} &
 \includegraphics[width=\largeur]{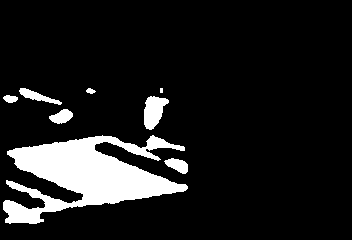} 
  \\
  
              \includegraphics[width=\largeur]{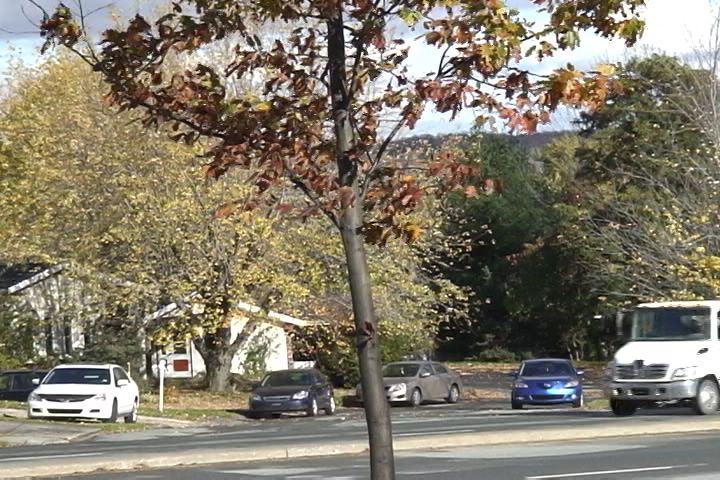} &
 \includegraphics[width=\largeur]{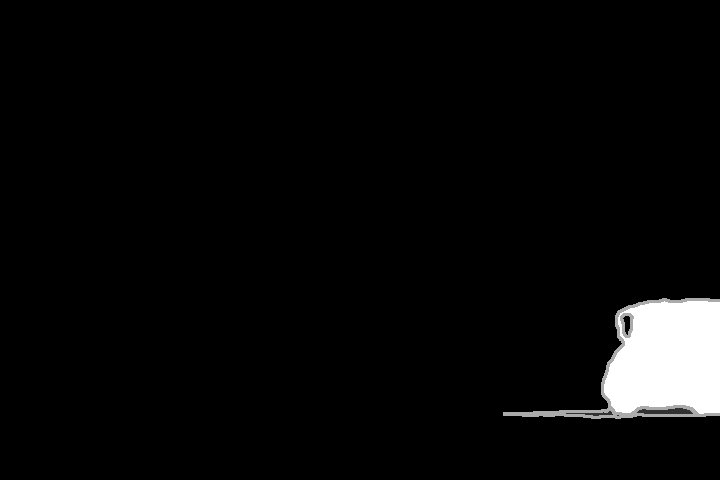} &
 \includegraphics[width=\largeur]{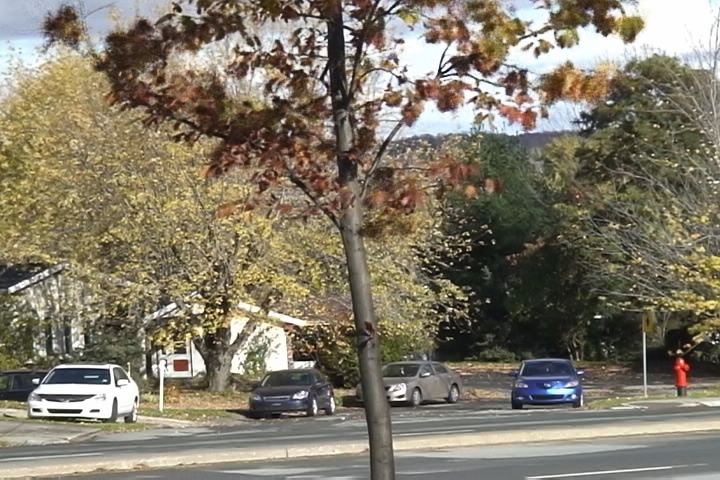} &
    \includegraphics[width=\largeur]{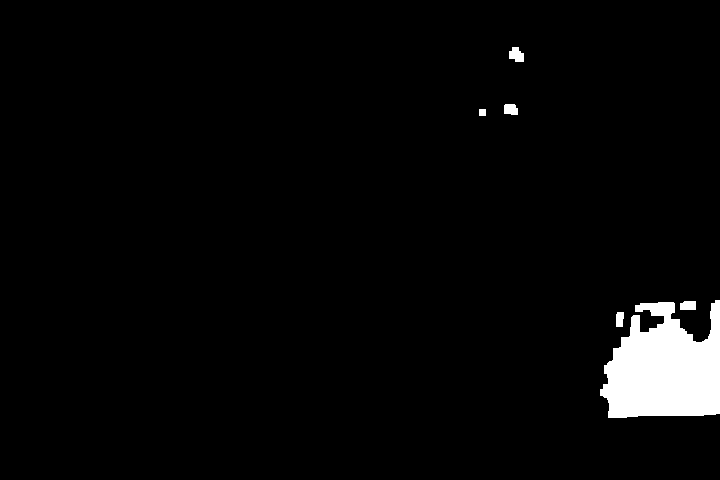} &
 \includegraphics[width=\largeur]{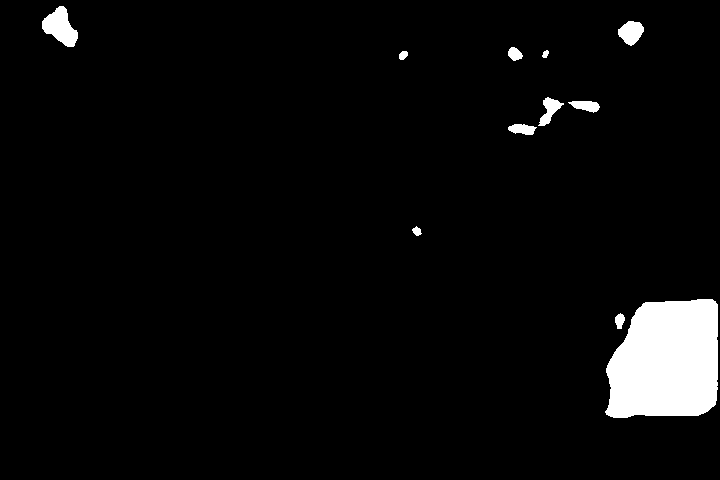} &
 \includegraphics[width=\largeur]{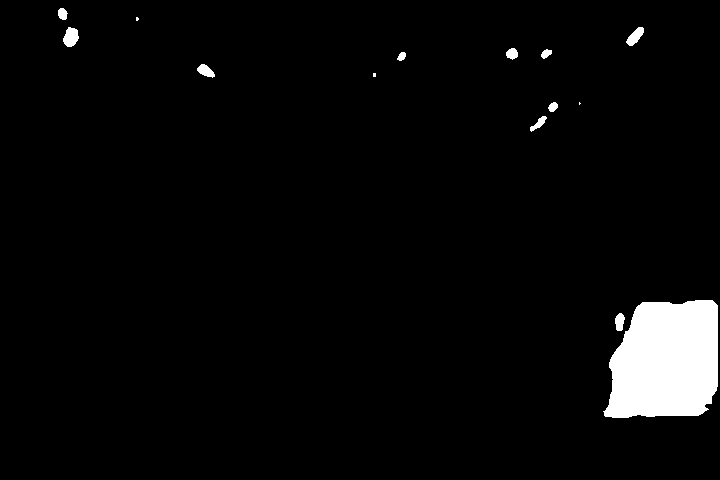} 
  \\
  
                \includegraphics[width=\largeur]{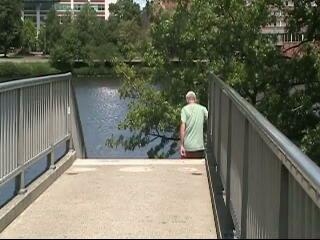} &
 \includegraphics[width=\largeur]{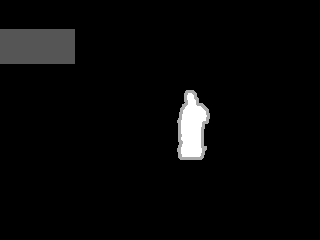} &
  \includegraphics[width=\largeur]{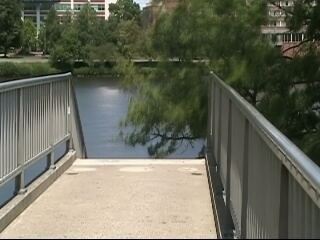} &
    \includegraphics[width=\largeur]{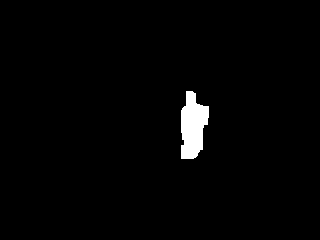} &
 \includegraphics[width=\largeur]{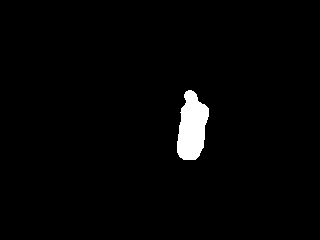} &
 \includegraphics[width=\largeur]{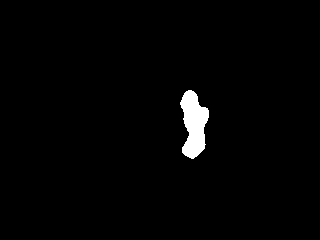} 
  \\
                  \includegraphics[width=\largeur]{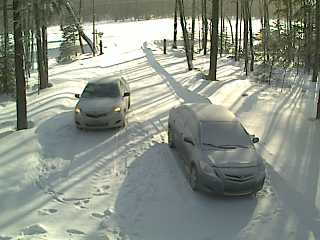} &
 \includegraphics[width=\largeur]{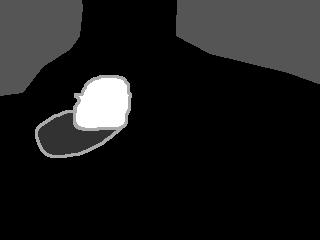} &
  \includegraphics[width=\largeur]{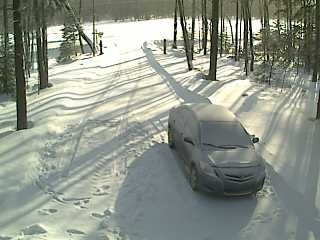} &
    \includegraphics[width=\largeur]{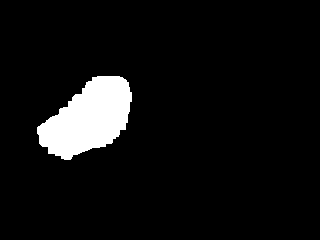} &
 \includegraphics[width=\largeur]{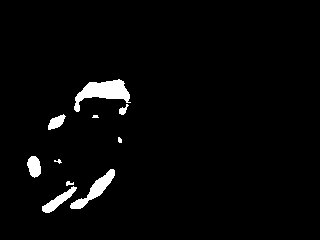} &
 \includegraphics[width=\largeur]{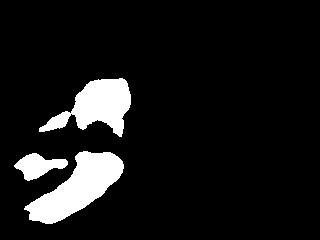} 
  \\
  
                    \includegraphics[width=\largeur]{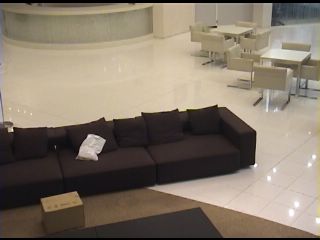} &
 \includegraphics[width=\largeur]{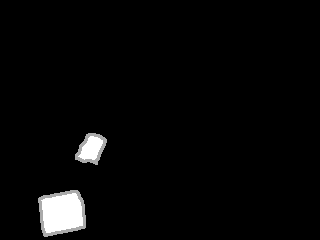} &
  \includegraphics[width=\largeur]{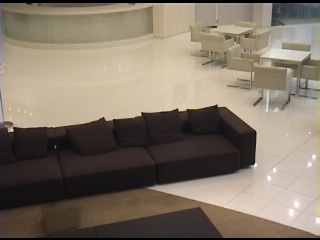} &
    \includegraphics[width=\largeur]{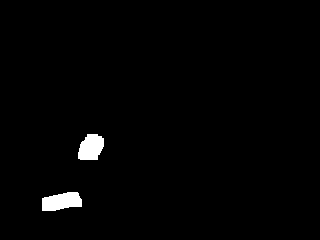} &
 \includegraphics[width=\largeur]{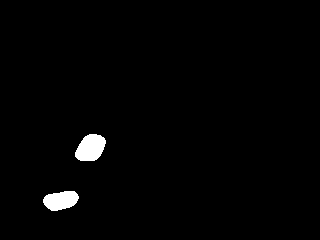} &
 \includegraphics[width=\largeur]{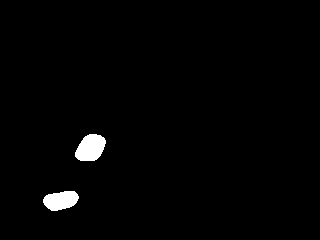} 
  \\

                    \includegraphics[width=\largeur]{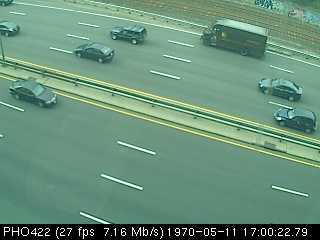} &
 \includegraphics[width=\largeur]{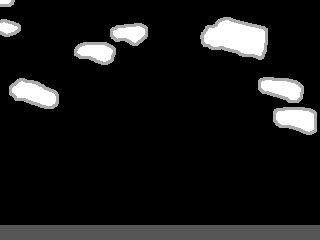} &
  \includegraphics[width=\largeur]{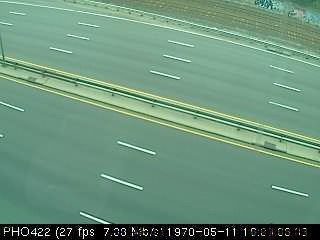} &
    \includegraphics[width=\largeur]{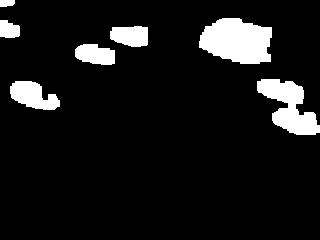} &
 \includegraphics[width=\largeur]{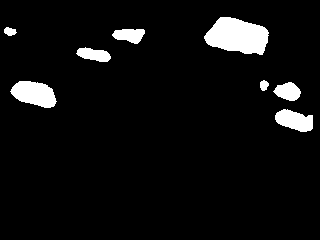} &
 \includegraphics[width=\largeur]{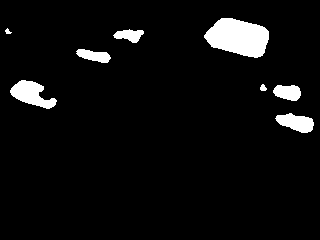} 
  \\
  
                  \includegraphics[width=\largeur]{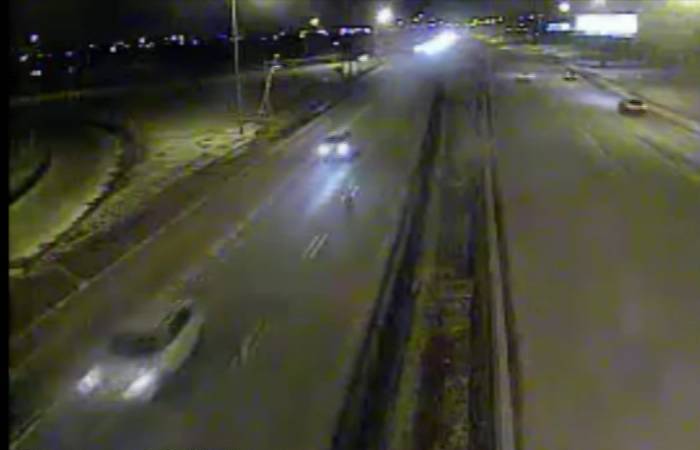} &
 \includegraphics[width=\largeur]{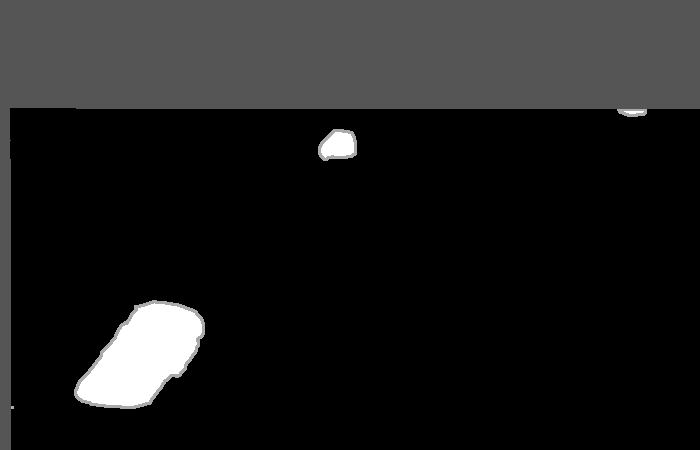} &
  \includegraphics[width=\largeur]{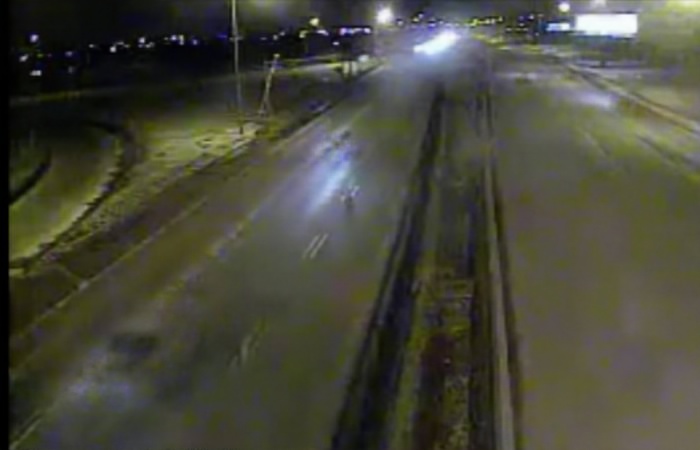} &
    \includegraphics[width=\largeur]{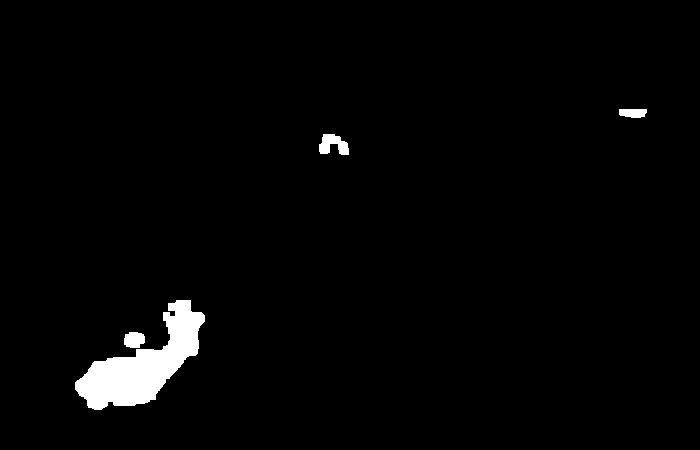} &
 \includegraphics[width=\largeur]{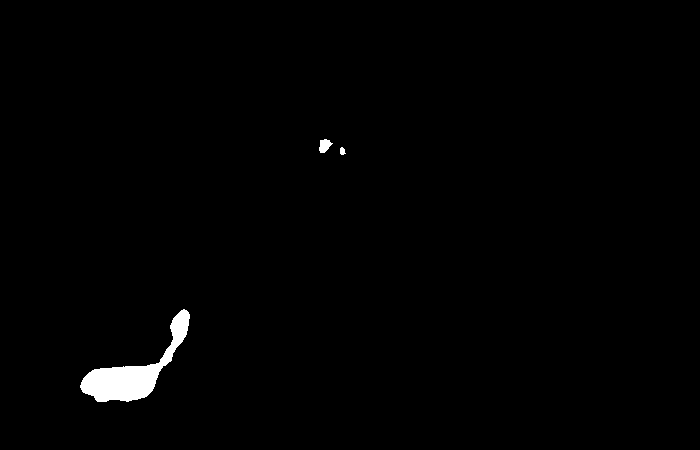} &
 \includegraphics[width=\largeur]{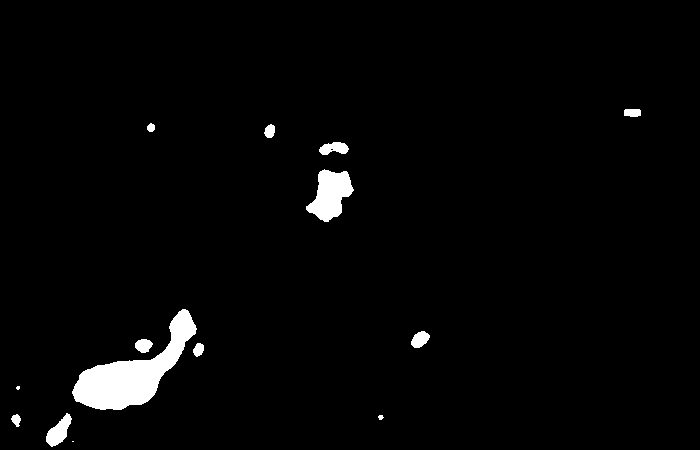} 
  \\
    
                \includegraphics[width=\largeur]{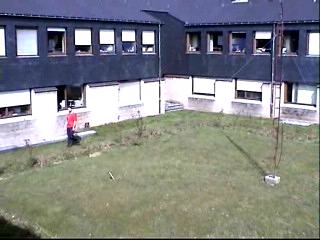} &
 \includegraphics[width=\largeur]{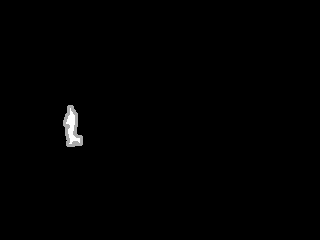} &
  \includegraphics[width=\largeur]{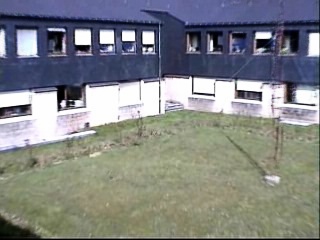} &
    \includegraphics[width=\largeur]{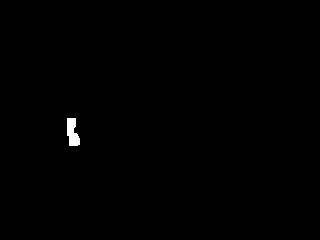} &
 \includegraphics[width=\largeur]{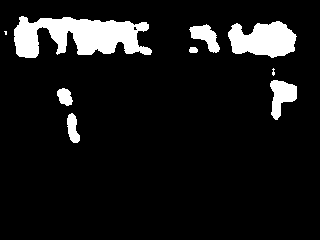} &
 \includegraphics[width=\largeur]{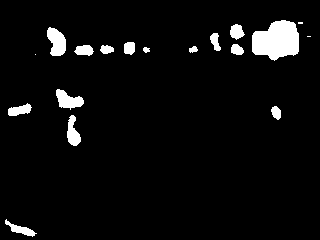} 
  \\
  
                  \includegraphics[width=\largeur]{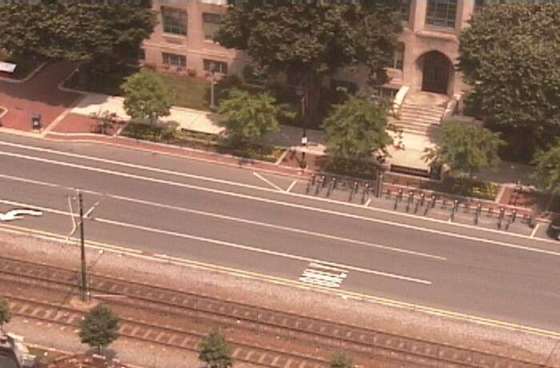} &
 \includegraphics[width=\largeur]{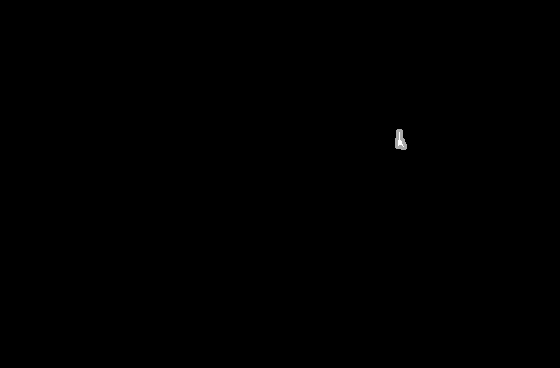} &
  \includegraphics[width=\largeur]{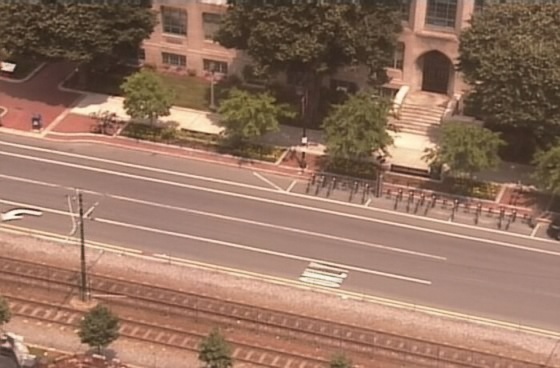} &
    \includegraphics[width=\largeur]{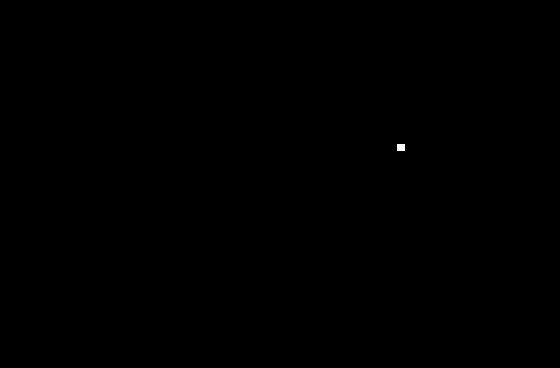} &
 \includegraphics[width=\largeur]{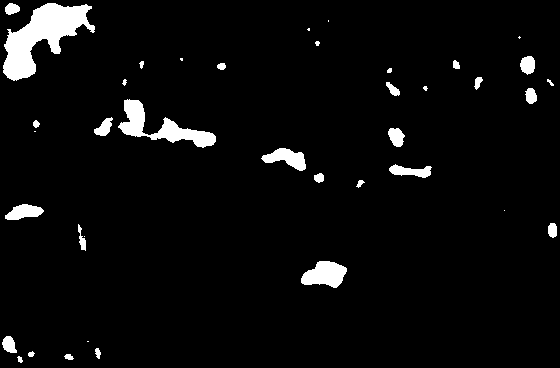} &
 \includegraphics[width=\largeur]{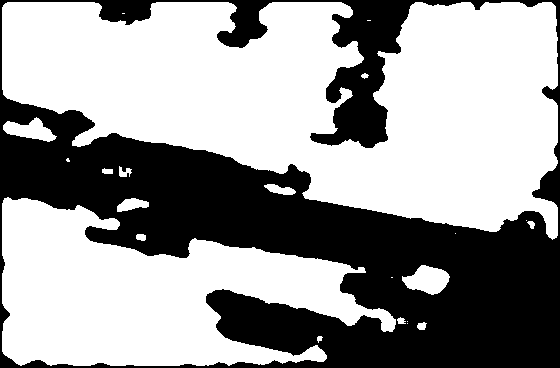} 
  \\
  
                      \includegraphics[width=\largeur]{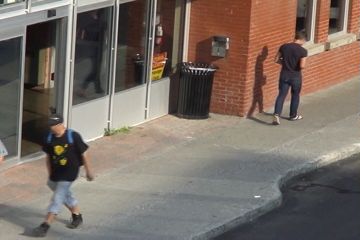} &
 \includegraphics[width=\largeur]{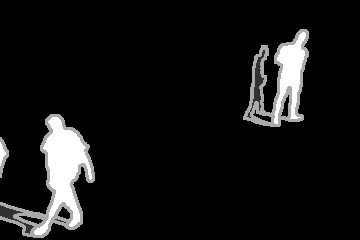} &
  \includegraphics[width=\largeur]{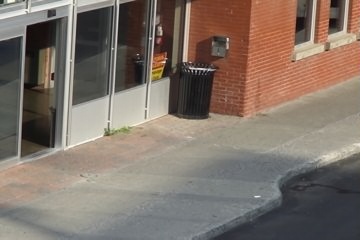} &
    \includegraphics[width=\largeur]{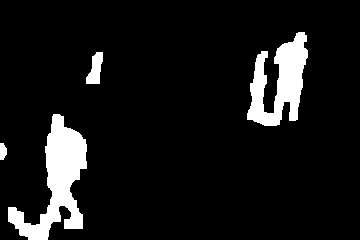} &
 \includegraphics[width=\largeur]{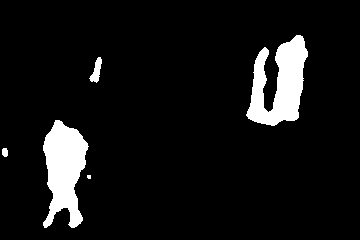} &
 \includegraphics[width=\largeur]{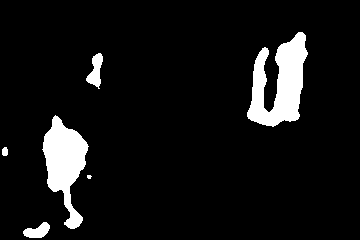} 
  \\
  
                    \includegraphics[width=\largeur]{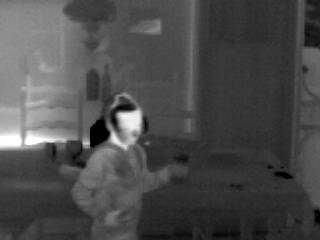} &
 \includegraphics[width=\largeur]{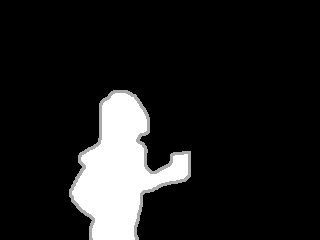} &
  \includegraphics[width=\largeur]{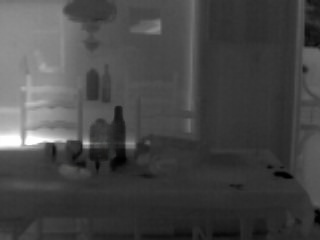} &

    \includegraphics[width=\largeur]{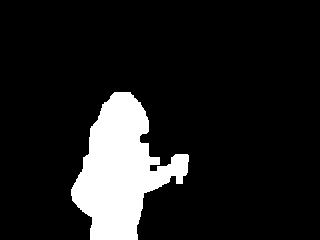} &
 \includegraphics[width=\largeur]{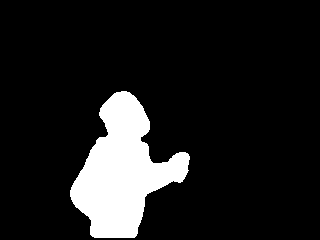} &
 \includegraphics[width=\largeur]{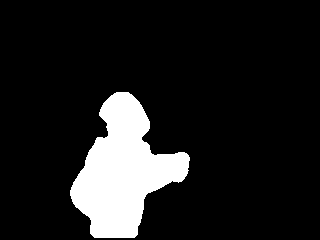} 
  \\

                  \includegraphics[width=\largeur]{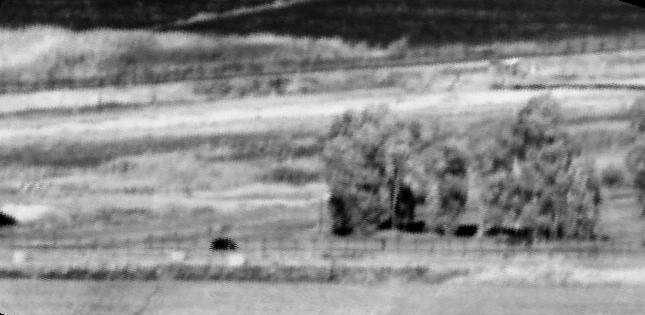} &
 \includegraphics[width=\largeur]{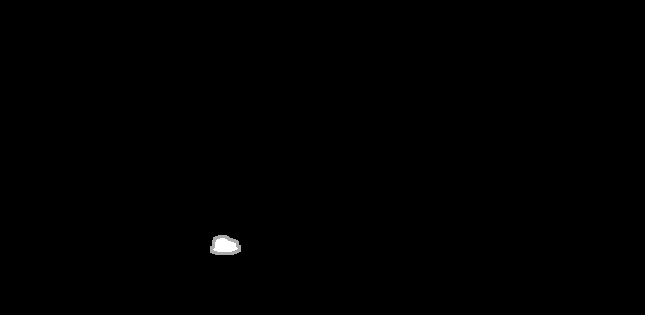} &
  \includegraphics[width=\largeur]{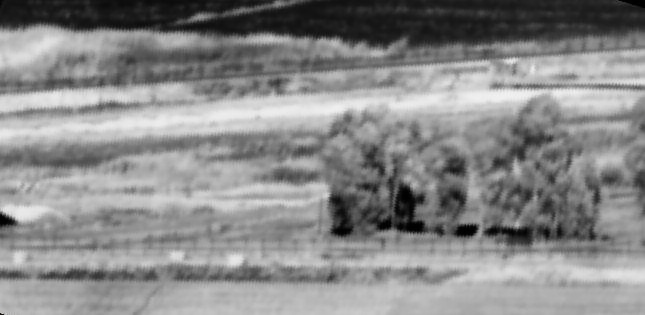} &
    \includegraphics[width=\largeur]{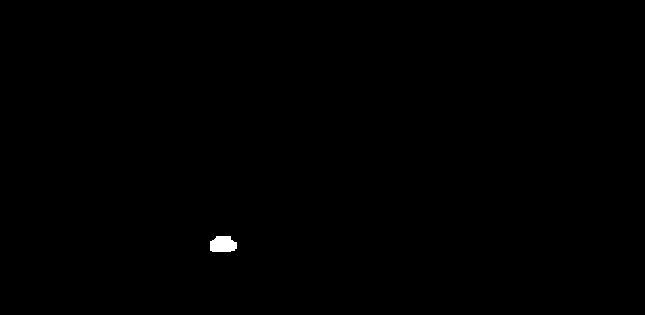} &
 \includegraphics[width=\largeur]{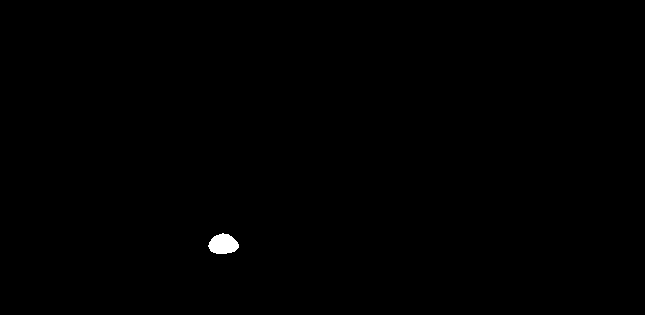} &
 \includegraphics[width=\largeur]{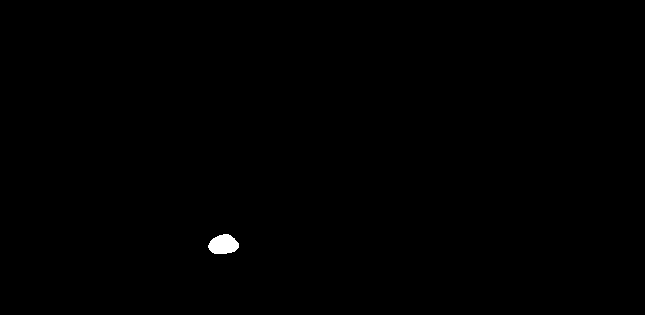} 
  \\

    \centering  \makecell{input\\frame}  &   \makecell{foreground \\ mask \\ ground truth}  &   \makecell{predicted \\background \\AE-NE (ours)}  &  \makecell{predicted \\foreground mask \\AE-NE (ours)} &   \makecell{predicted \\ foreground mask  \\PAWCS} & \makecell{predicted \\ foreground mask  \\SuBSENSE}  
\end{tabular}}
\caption{Examples of background reconstruction and foreground segmentation on the CDnet 2014 dataset produced using the proposed model and comparison with PAWCS and SuBSENSE}
\label{figure:cdnet2}
\end{figure*}

\begin{figure*}
\centering
  \scalebox{0.70}{
\begin{tabular}{*{6}{m{26 mm}}}

      \includegraphics[width=\largeur]{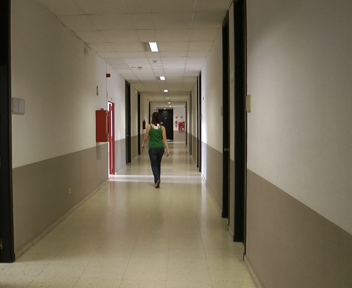} &
    \includegraphics[width=\largeur]{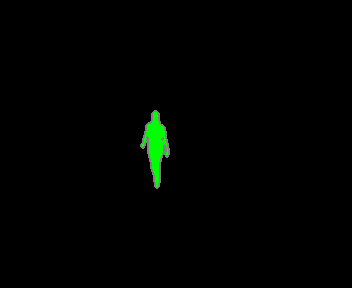} &
     \includegraphics[width=\largeur]{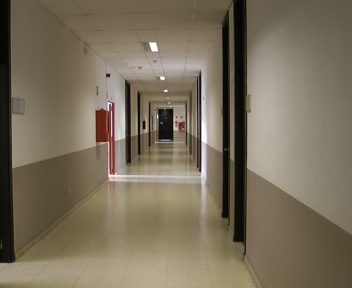} & 
       \includegraphics[width=\largeur]{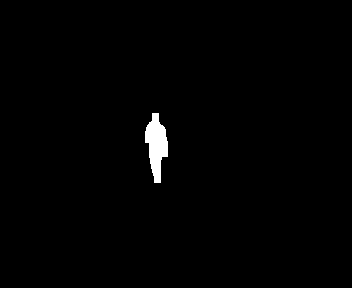} & 
          \includegraphics[width=\largeur]{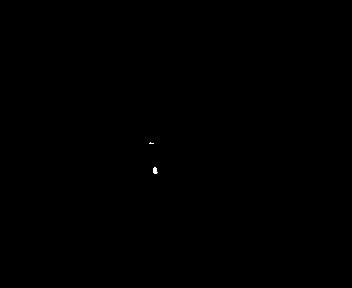} &
       \includegraphics[width=\largeur]{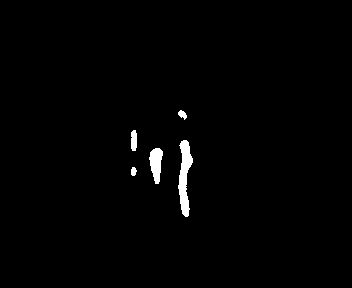} \\
       
         \includegraphics[width=\largeur]{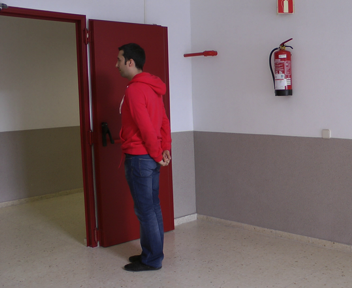} &
    \includegraphics[width=\largeur]{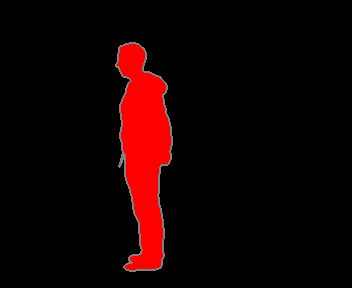} &
     \includegraphics[width=\largeur]{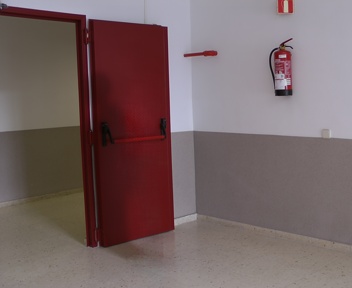} & 
       \includegraphics[width=\largeur]{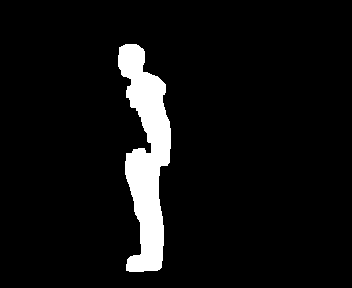} & 
          \includegraphics[width=\largeur]{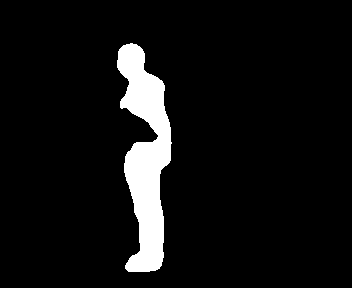} &
       \includegraphics[width=\largeur]{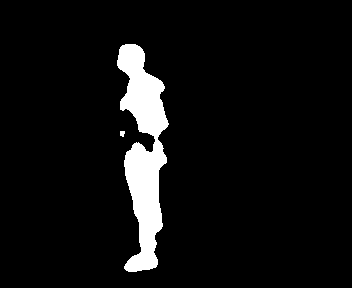} \\

            \includegraphics[width=\largeur]{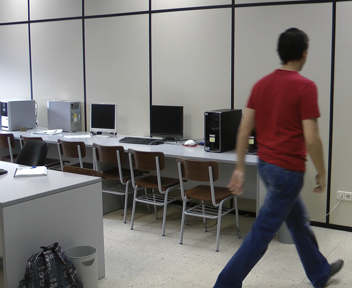} &
    \includegraphics[width=\largeur]{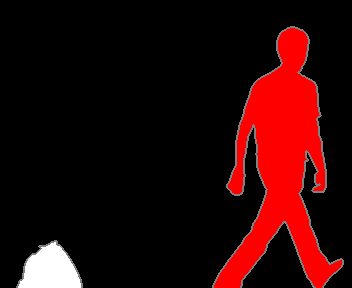} &
     \includegraphics[width=\largeur]{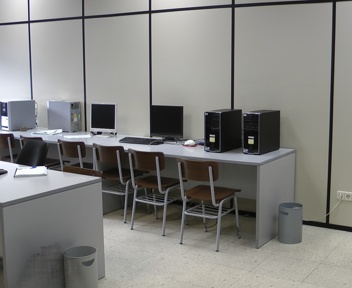} & 
       \includegraphics[width=\largeur]{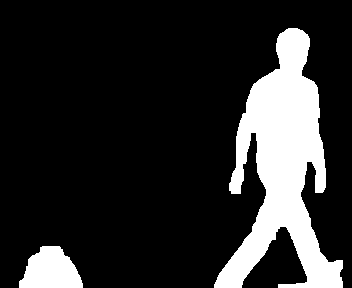} & 
          \includegraphics[width=\largeur]{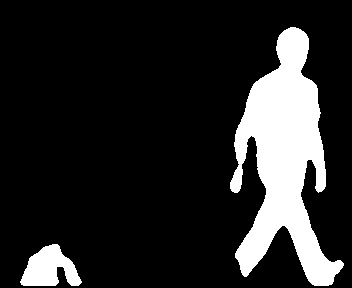} &
       \includegraphics[width=\largeur]{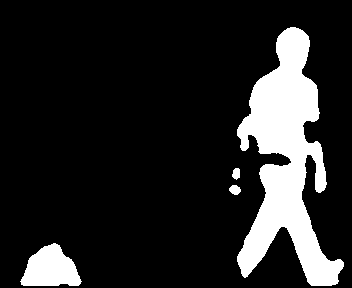} \\

                \includegraphics[width=\largeur]{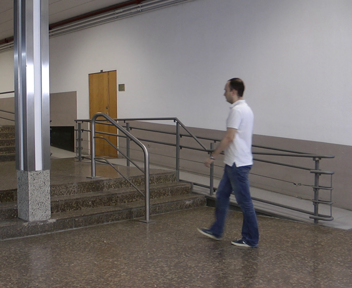} &
    \includegraphics[width=\largeur]{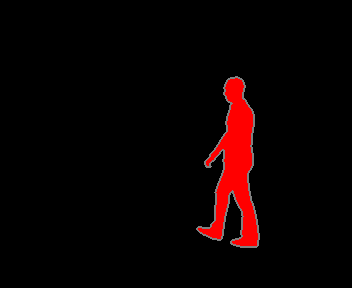} &
     \includegraphics[width=\largeur]{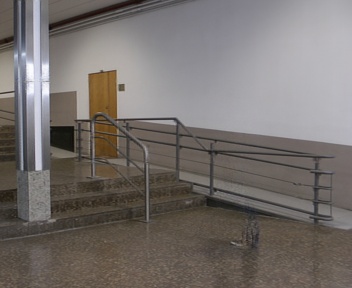} & 
       \includegraphics[width=\largeur]{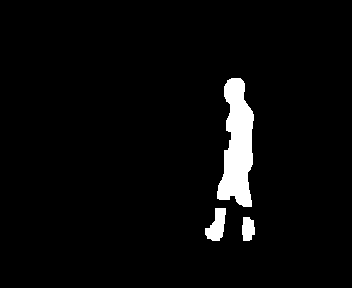} & 
          \includegraphics[width=\largeur]{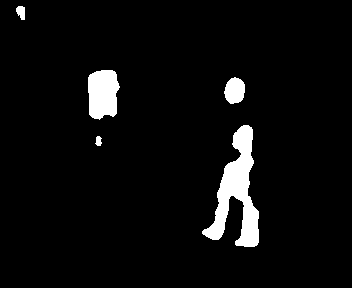} &
       \includegraphics[width=\largeur]{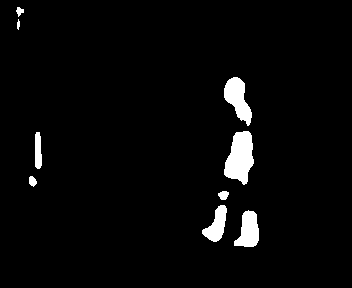} \\
       
                \includegraphics[width=\largeur]{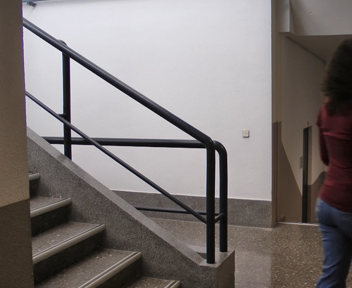} &
    \includegraphics[width=\largeur]{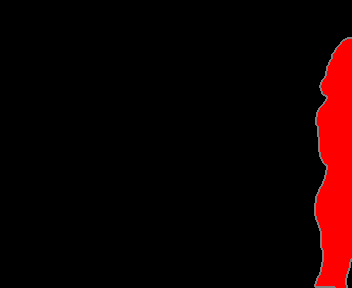} &
     \includegraphics[width=\largeur]{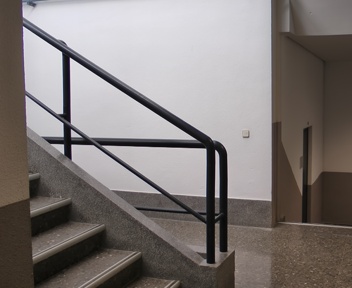} & 
       \includegraphics[width=\largeur]{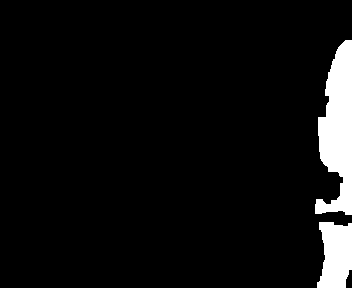} & 
          \includegraphics[width=\largeur]{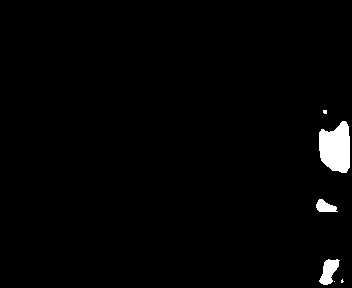} &
       \includegraphics[width=\largeur]{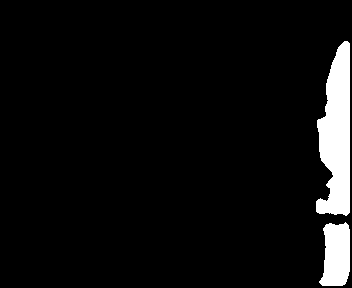} \\
       
                \includegraphics[width=\largeur]{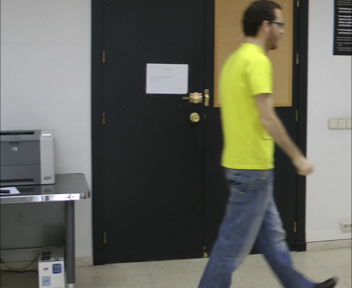} &
    \includegraphics[width=\largeur]{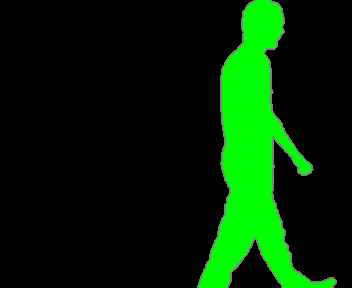} &
     \includegraphics[width=\largeur]{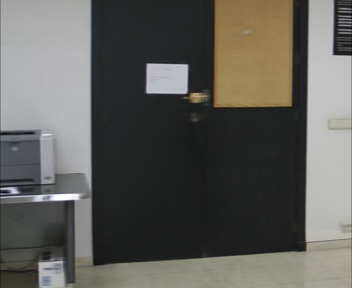} & 
       \includegraphics[width=\largeur]{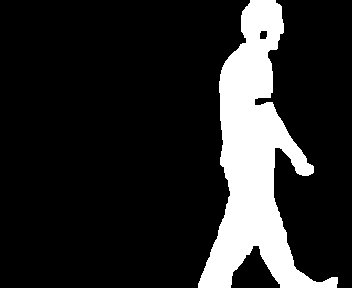} & 
          \includegraphics[width=\largeur]{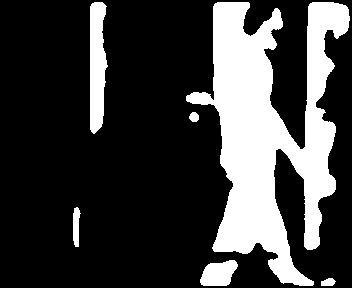} &
       \includegraphics[width=\largeur]{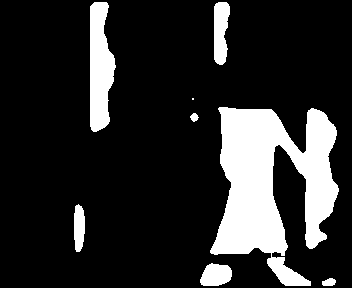} \\
       
                \includegraphics[width=\largeur]{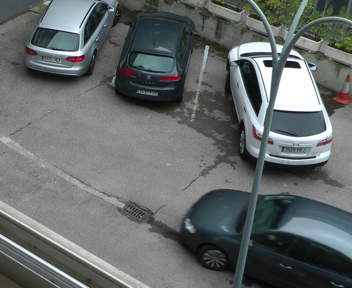} &
    \includegraphics[width=\largeur]{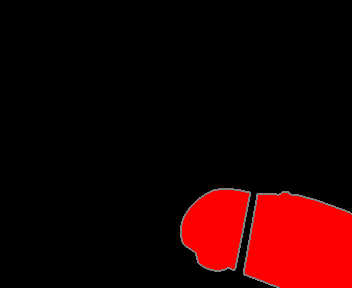} &
     \includegraphics[width=\largeur]{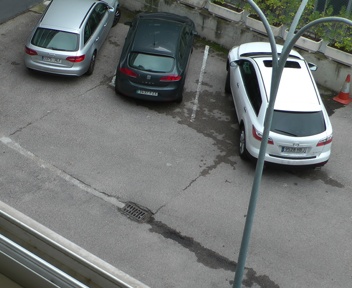} & 
       \includegraphics[width=\largeur]{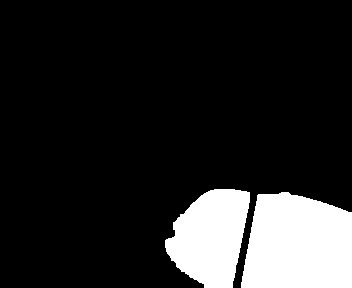} & 
          \includegraphics[width=\largeur]{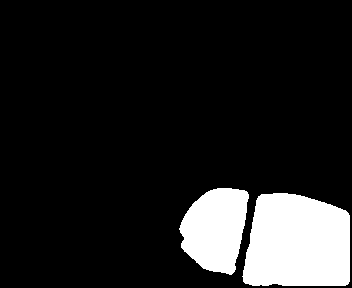} &
       \includegraphics[width=\largeur]{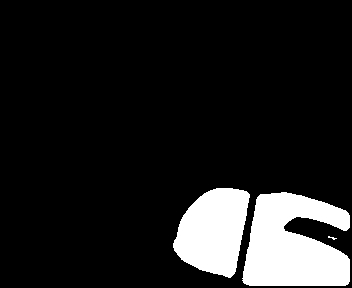} \\
       
                 \includegraphics[width=\largeur]{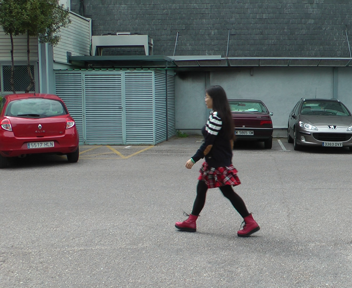} &
    \includegraphics[width=\largeur]{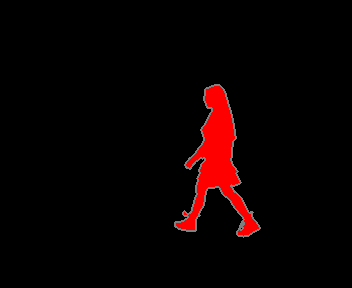} &
     \includegraphics[width=\largeur]{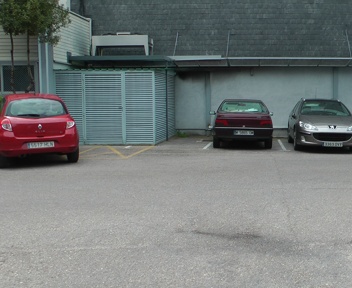} & 
       \includegraphics[width=\largeur]{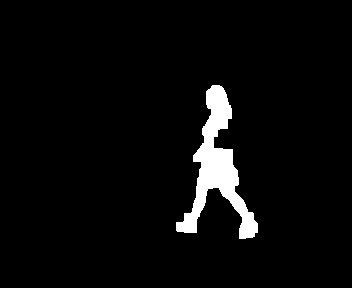} & 
          \includegraphics[width=\largeur]{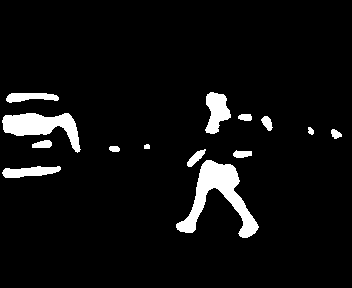} &
       \includegraphics[width=\largeur]{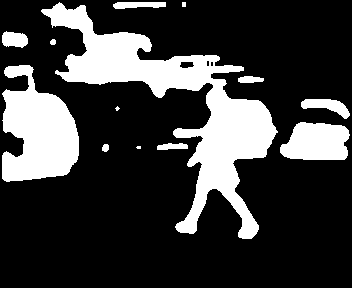} \\
       
                   \includegraphics[width=\largeur]{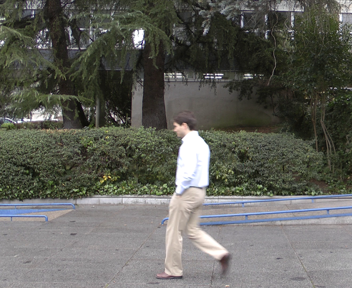} &
    \includegraphics[width=\largeur]{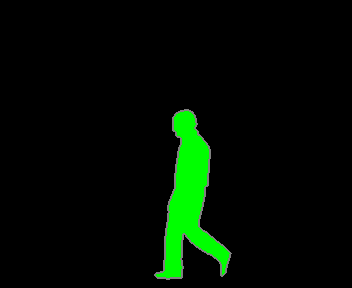} &
     \includegraphics[width=\largeur]{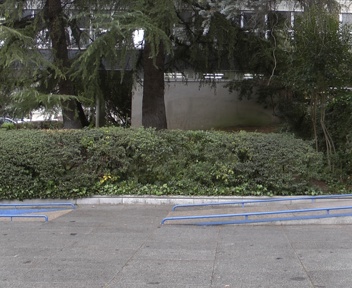} & 
       \includegraphics[width=\largeur]{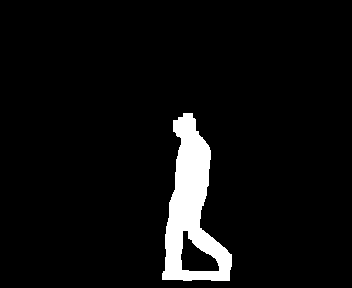} & 
          \includegraphics[width=\largeur]{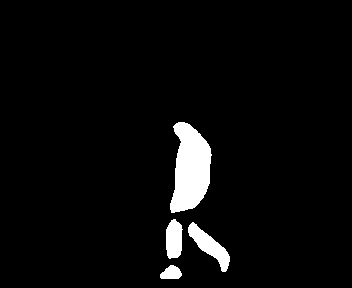} &
       \includegraphics[width=\largeur]{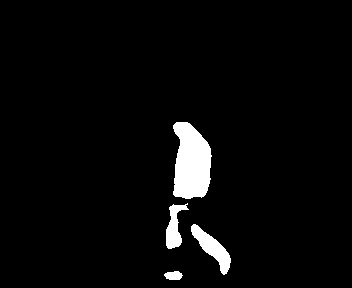} \\
       
                          \includegraphics[width=\largeur]{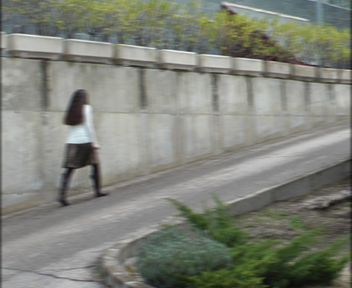} &
    \includegraphics[width=\largeur]{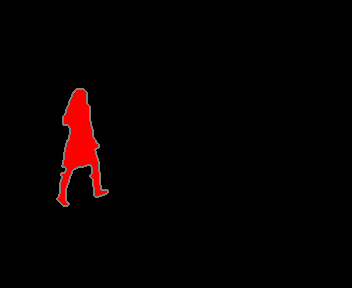} &
     \includegraphics[width=\largeur]{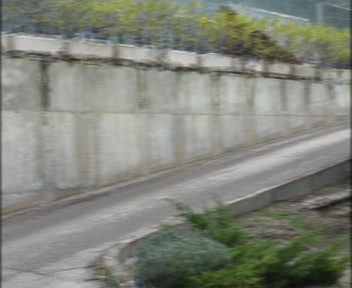} & 
       \includegraphics[width=\largeur]{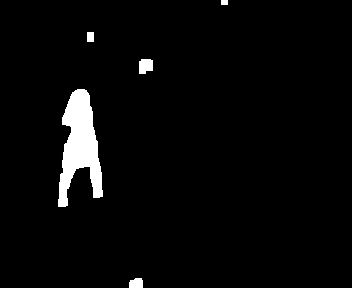} & 
          \includegraphics[width=\largeur]{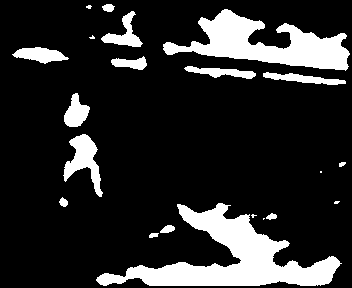} &
       \includegraphics[width=\largeur]{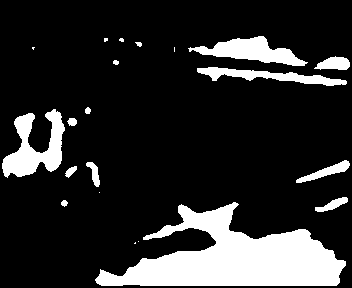} \\
       
       \includegraphics[width=\largeur]{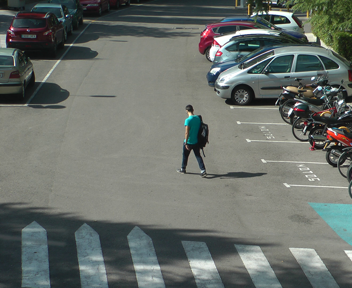} &
    \includegraphics[width=\largeur]{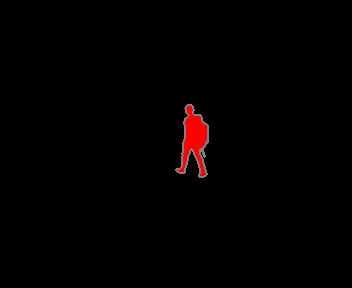} &
     \includegraphics[width=\largeur]{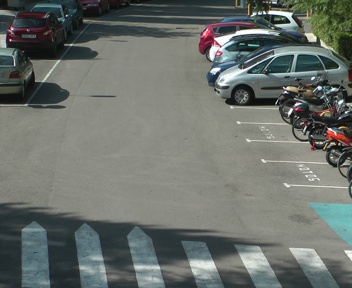} & 
       \includegraphics[width=\largeur]{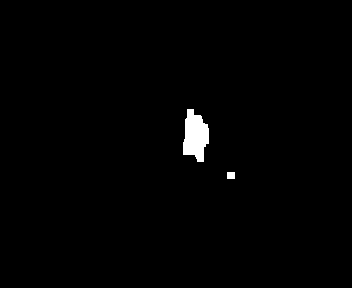} & 
          \includegraphics[width=\largeur]{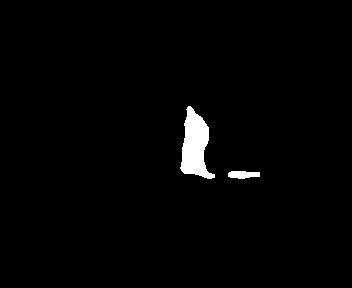} &
       \includegraphics[width=\largeur]{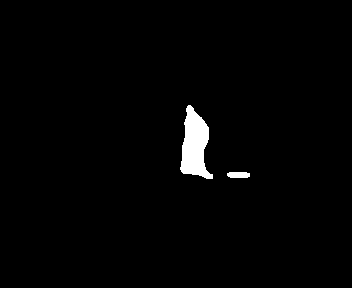} \\

    \centering  \makecell{input\\frame}  &   \makecell{foreground \\ mask \\ ground truth}  &   \makecell{predicted \\background \\AE-NE (ours)}  &  \makecell{predicted \\foreground mask \\AE-NE (ours)} &   \makecell{predicted \\ foreground mask  \\PAWCS} & \makecell{predicted \\ foreground mask  \\SuBSENSE}  
\end{tabular}}
\caption{Examples of background reconstruction and foreground segmentation on the LASIESTA dataset produced using the proposed model and comparison with PAWCS and SuBSENSE}
\label{figure:lasiesta}
\end{figure*}

\begin{figure*}
\centering
  \scalebox{0.70}{
\begin{tabular}{*{6}{m{26 mm}}}

 \includegraphics[width=\largeur]{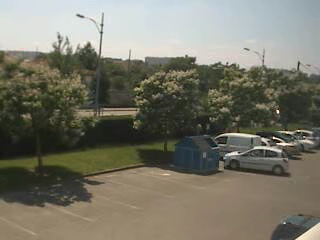} &
  \includegraphics[width=\largeur]{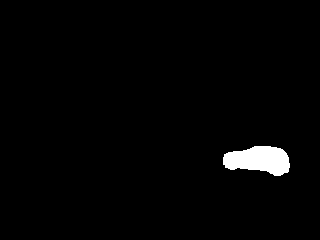} &
  \includegraphics[width=\largeur]{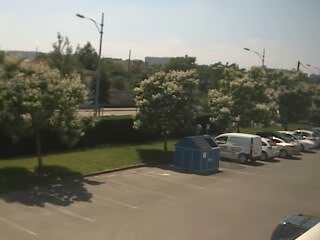} &
    \includegraphics[width=\largeur]{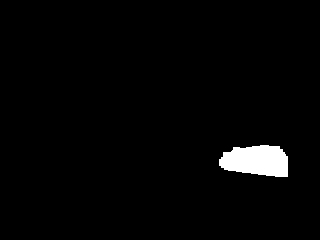} & 
      \includegraphics[width=\largeur]{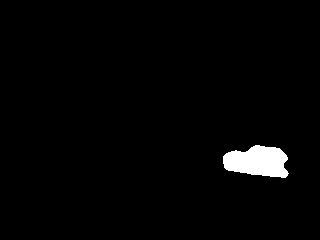} & 
        \includegraphics[width=\largeur]{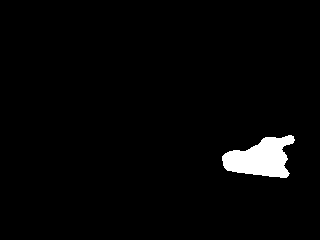}     \\
    
  \includegraphics[width=\largeur]{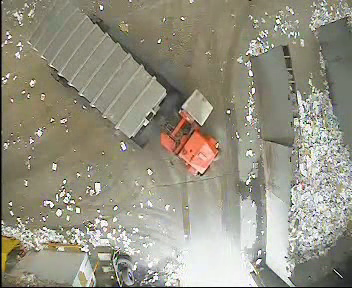} &
  \includegraphics[width=\largeur]{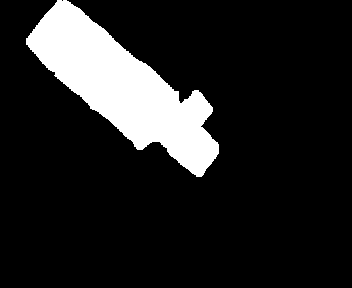} &
  \includegraphics[width=\largeur]{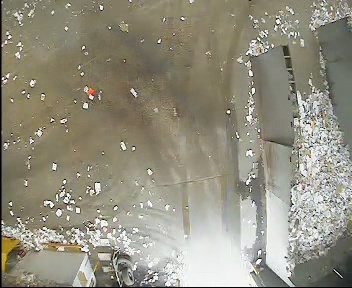} &
    \includegraphics[width=\largeur]{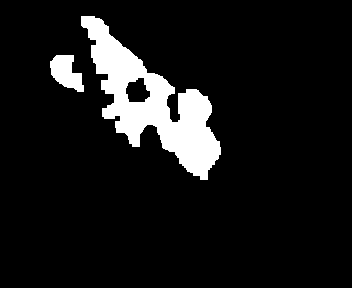} &
        \includegraphics[width=\largeur]{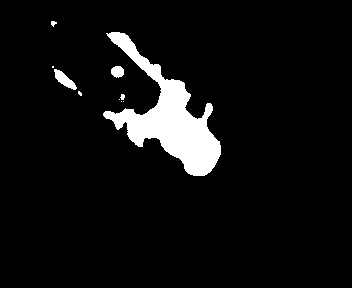} & 
        \includegraphics[width=\largeur]{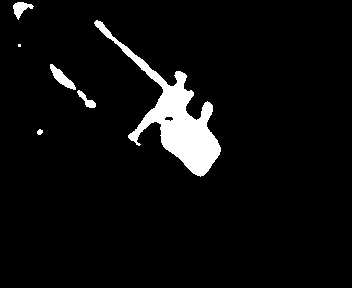}     \\
    
     \includegraphics[width=\largeur]{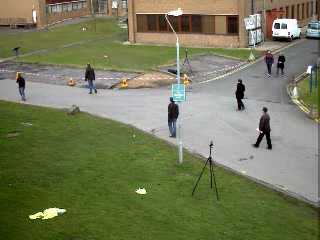} &
  \includegraphics[width=\largeur]{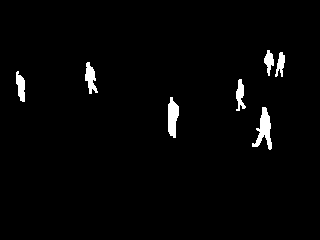} &
  \includegraphics[width=\largeur]{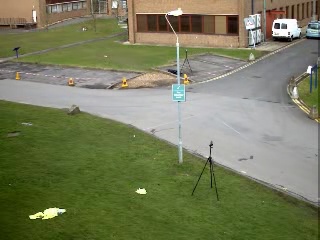} &
    \includegraphics[width=\largeur]{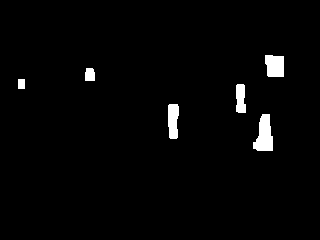} & 
         \includegraphics[width=\largeur]{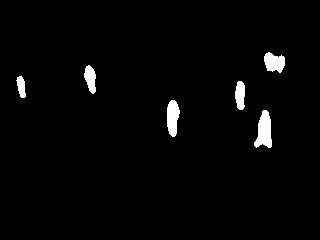} & 
        \includegraphics[width=\largeur]{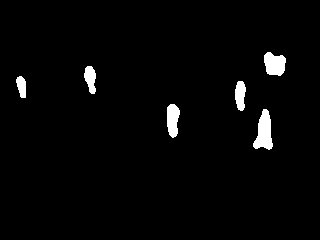}     \\
    
     \includegraphics[width=\largeur]{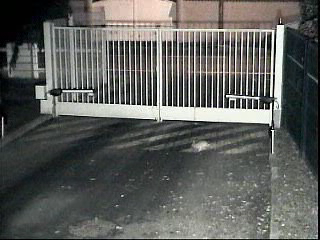} &
  \includegraphics[width=\largeur]{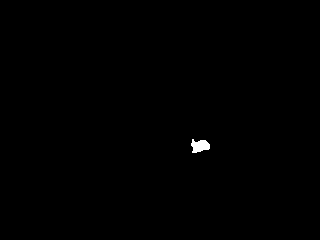} &
  \includegraphics[width=\largeur]{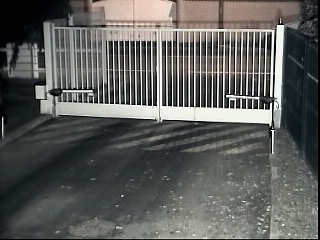} &
    \includegraphics[width=\largeur]{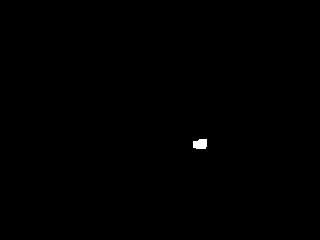} &
             \includegraphics[width=\largeur]{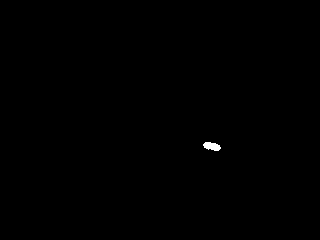} & 
        \includegraphics[width=\largeur]{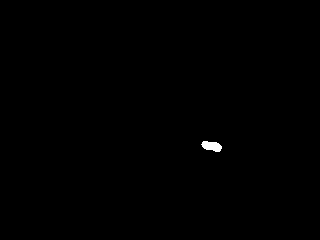}     \\
    
     \includegraphics[width=\largeur]{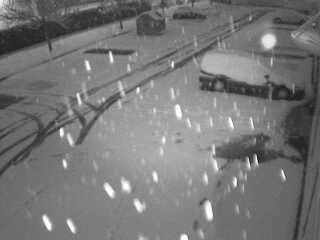} &
  \includegraphics[width=\largeur]{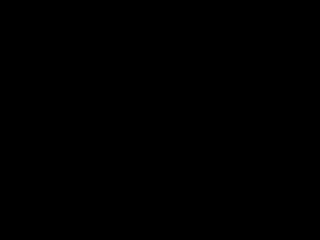} &
  \includegraphics[width=\largeur]{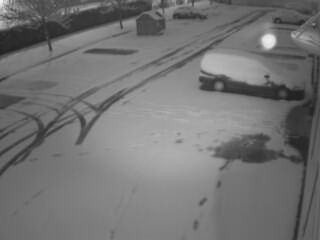} &
    \includegraphics[width=\largeur]{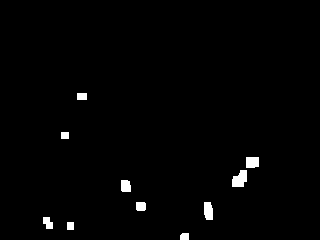} &
       \includegraphics[width=\largeur]{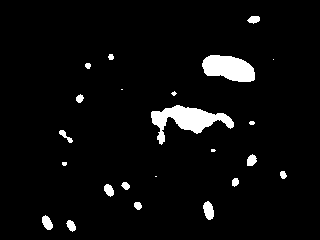} & 
        \includegraphics[width=\largeur]{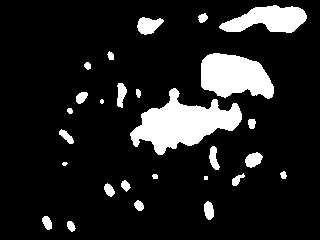}     \\

     \includegraphics[width=\largeur]{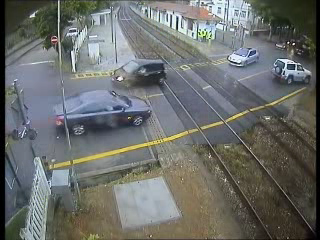} &
  \includegraphics[width=\largeur]{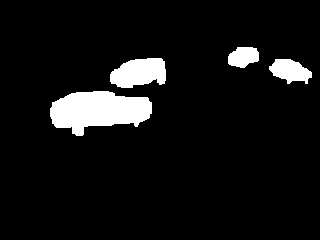} &
  \includegraphics[width=\largeur]{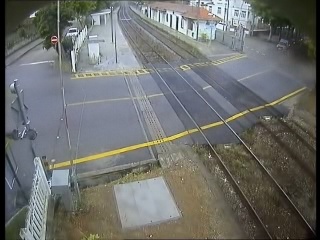} &
    \includegraphics[width=\largeur]{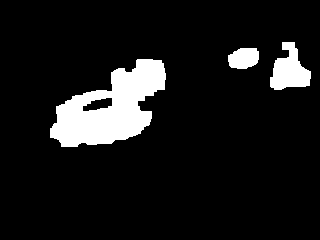} &
        \includegraphics[width=\largeur]{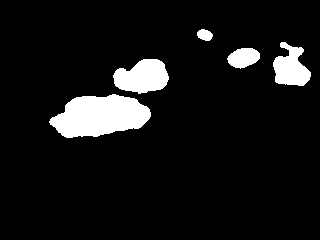} & 
        \includegraphics[width=\largeur]{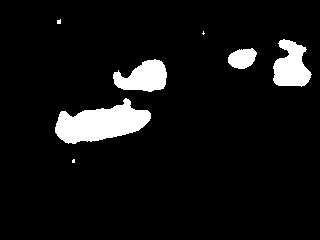}     \\

             \includegraphics[width=\largeur]{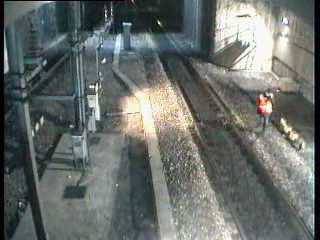} &
  \includegraphics[width=\largeur]{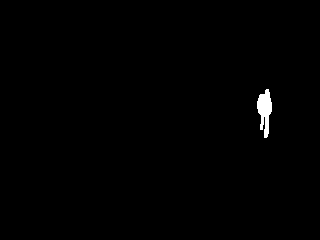} &
  \includegraphics[width=\largeur]{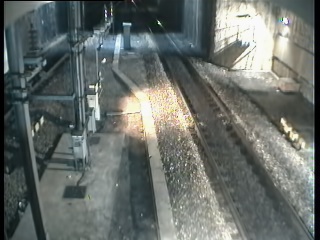} &
    \includegraphics[width=\largeur]{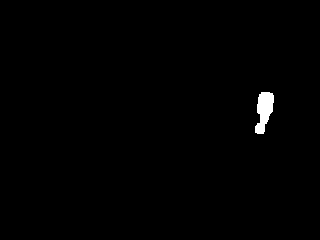} &
        \includegraphics[width=\largeur]{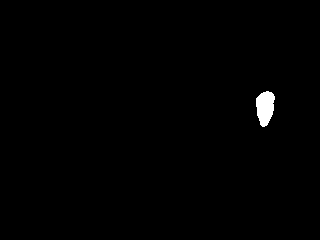} & 
        \includegraphics[width=\largeur]{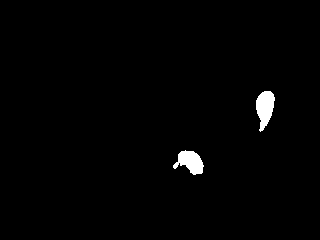}     \\

      \includegraphics[width=\largeur]{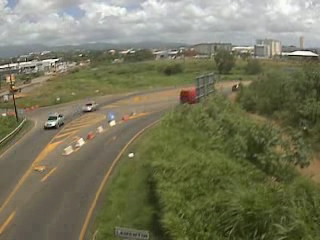} &
  \includegraphics[width=\largeur]{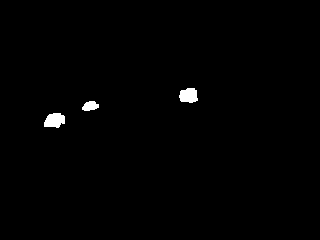} &
  \includegraphics[width=\largeur]{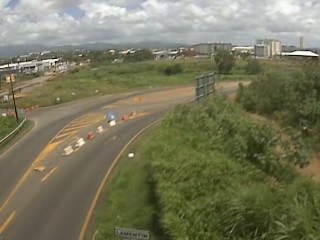} &
    \includegraphics[width=\largeur]{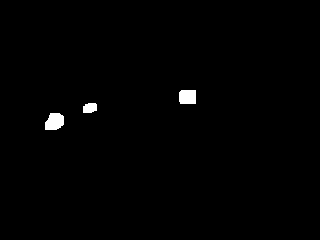} &
         \includegraphics[width=\largeur]{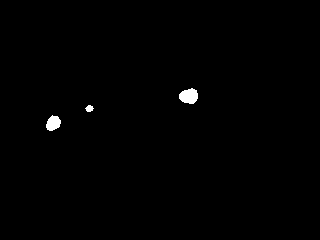} & 
        \includegraphics[width=\largeur]{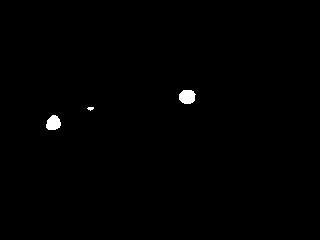}     \\
    
          \includegraphics[width=\largeur]{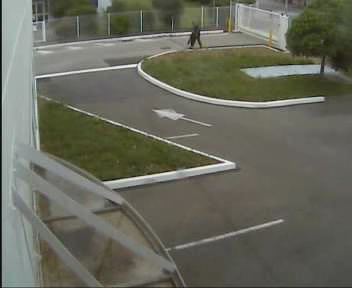} &
  \includegraphics[width=\largeur]{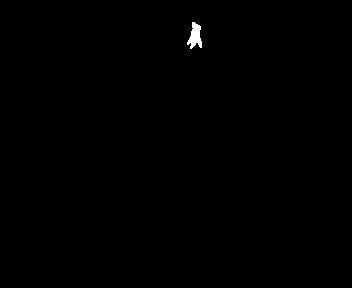} &
  \includegraphics[width=\largeur]{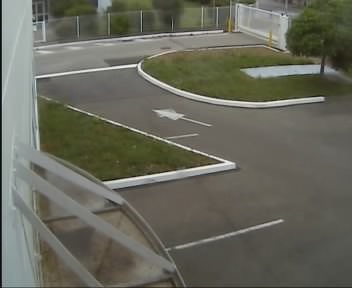} &
    \includegraphics[width=\largeur]{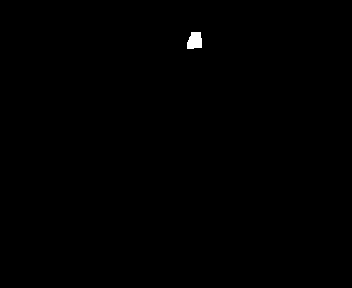} & 
            \includegraphics[width=\largeur]{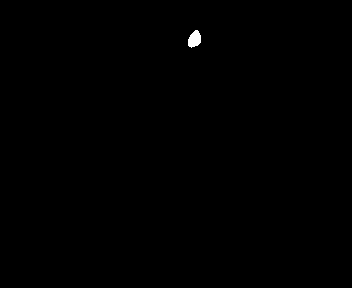} & 
        \includegraphics[width=\largeur]{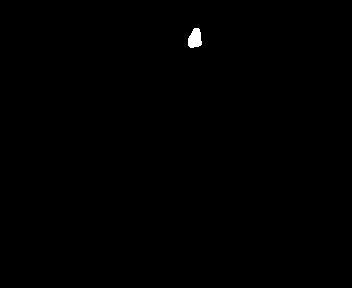}     \\
aen
    \centering  \makecell{input\\frame}  &   \makecell{foreground \\ mask \\ ground truth}  &   \makecell{predicted \\background \\AE-NE (ours)}  &  \makecell{predicted \\foreground mask \\AE-NE (ours)} &   \makecell{predicted \\ foreground mask  \\PAWCS} & \makecell{predicted \\ foreground mask  \\SuBSENSE}  
\end{tabular}}
\caption{Examples of background reconstruction and foreground segmentation on the BMC 2012 dataset produced using the proposed model and comparison with PAWCS and SuBSENSE}
\label{figure:bmc2012}
\end{figure*}

 \setlength{\largeur}{16mm}
 \begin{figure}
\centering
  \scalebox{0.7}{
\begin{tabular}{c*{8}{m{18 mm}}}

     \makecell{input \\ frame}  &             \includegraphics[width=\largeur]{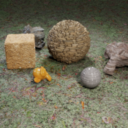} &
                     \includegraphics[width=\largeur]{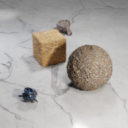} &
   \includegraphics[width=\largeur]{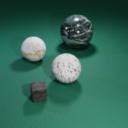} &
    \includegraphics[width=\largeur]{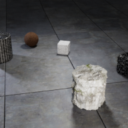} &  
       \includegraphics[width=\largeur]{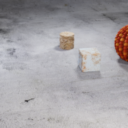} &
                     \includegraphics[width=\largeur]{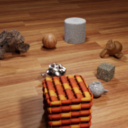} &
   \includegraphics[width=\largeur]{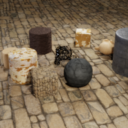} &
    \includegraphics[width=\largeur]{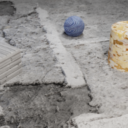}       \\       
    
          \makecell{ground \\ truth \\ object \\ segmentation} &           \includegraphics[width=\largeur]{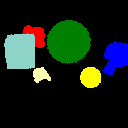} &
                         \includegraphics[width=\largeur]{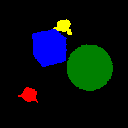}&
  \includegraphics[width=\largeur]{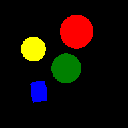} &
   \includegraphics[width=\largeur]{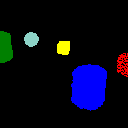} & 
      \includegraphics[width=\largeur]{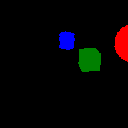} & 
        \includegraphics[width=\largeur]{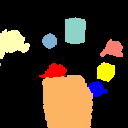}  &
         \includegraphics[width=\largeur]{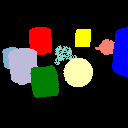} & 
        \includegraphics[width=\largeur]{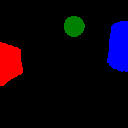}     \\

        \makecell{predicted \\ background}  &                 \includegraphics[width=\largeur]{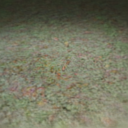}  &
                         \includegraphics[width=\largeur]{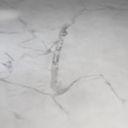} &
 \includegraphics[width=\largeur]{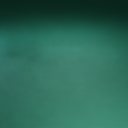}  &
 \includegraphics[width=\largeur]{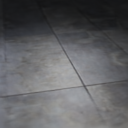} & 
  \includegraphics[width=\largeur]{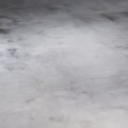}  &
 \includegraphics[width=\largeur]{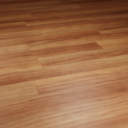} & 
  \includegraphics[width=\largeur]{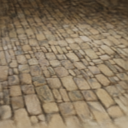}  &
 \includegraphics[width=\largeur]{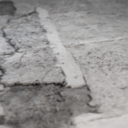}   \\

   \makecell{predicted \\ foreground \\ mask}   &         \includegraphics[width=\largeur]{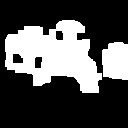} &
             \includegraphics[width=\largeur]{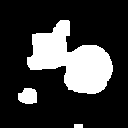}  &
  \includegraphics[width=\largeur]{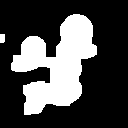}  &
 \includegraphics[width=\largeur]{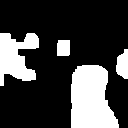}  &
   \includegraphics[width=\largeur]{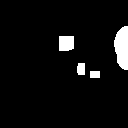}  &
 \includegraphics[width=\largeur]{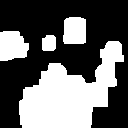}  &
   \includegraphics[width=\largeur]{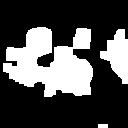}  &
 \includegraphics[width=\largeur]{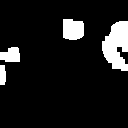}    \\
\end{tabular}}
\caption{Examples of background reconstruction and foreground segmentation on Clevrtex dataset }
\label{figure:clevrtex}
\end{figure}

 \begin{figure}
\centering
  \scalebox{0.7}{
\begin{tabular}{c*{8}{m{18 mm}}}

     \makecell{input \\ frame}  &      
    
      \includegraphics[width=\largeur]{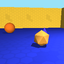} &
       \includegraphics[width=\largeur]{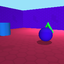} &
         \includegraphics[width=\largeur]{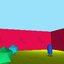} &
       \includegraphics[width=\largeur]{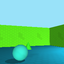} &
         \includegraphics[width=\largeur]{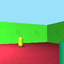} &
       \includegraphics[width=\largeur]{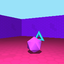} &
         \includegraphics[width=\largeur]{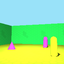} &
       \includegraphics[width=\largeur]{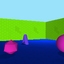} 
         \\          
    
       \makecell{ground \\ truth \\ object \\ segmentation} &                 
      \includegraphics[width=\largeur]{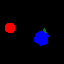} & 
        \includegraphics[width=\largeur]{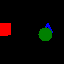}  &
      \includegraphics[width=\largeur]{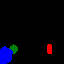} & 
        \includegraphics[width=\largeur]{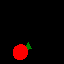}  &
              \includegraphics[width=\largeur]{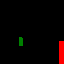} & 
        \includegraphics[width=\largeur]{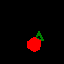}  &
              \includegraphics[width=\largeur]{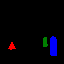} & 
        \includegraphics[width=\largeur]{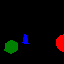}  
   \\

        \makecell{predicted \\ background}  &       

  \includegraphics[width=\largeur]{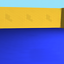}  &
 \includegraphics[width=\largeur]{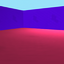} &
   \includegraphics[width=\largeur]{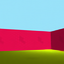}  &
 \includegraphics[width=\largeur]{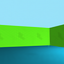} &
   \includegraphics[width=\largeur]{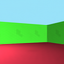}  &
 \includegraphics[width=\largeur]{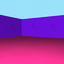} &
   \includegraphics[width=\largeur]{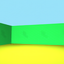}  &
 \includegraphics[width=\largeur]{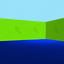} 

  \\

   \makecell{predicted \\ foreground \\ mask}   &      
  \includegraphics[width=\largeur]{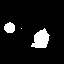}  &
   \includegraphics[width=\largeur]{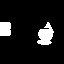} &
  \includegraphics[width=\largeur]{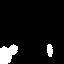}  &
   \includegraphics[width=\largeur]{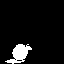} &
     \includegraphics[width=\largeur]{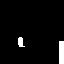}  &
   \includegraphics[width=\largeur]{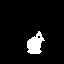} &
     \includegraphics[width=\largeur]{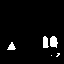}  &
   \includegraphics[width=\largeur]{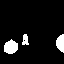} 
  \\
\end{tabular}}
\caption{Examples of background reconstruction and foreground segmentation on ObjectsRoom dataset }
\label{figure:objectsroom}
\end{figure}

 \begin{figure}
\centering
  \scalebox{0.7}{
\begin{tabular}{c*{8}{m{18 mm}}}

     \makecell{input \\ frame}  &         
     \includegraphics[width=\largeur]{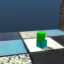} &
         \includegraphics[width=\largeur]{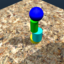} &
               \includegraphics[width=\largeur]{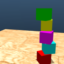} &
                    \includegraphics[width=\largeur]{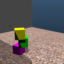} &
                         \includegraphics[width=\largeur]{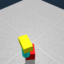} &
                              \includegraphics[width=\largeur]{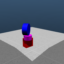} &
                                   \includegraphics[width=\largeur]{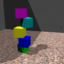} &
                                        \includegraphics[width=\largeur]{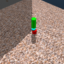}

    \\        
           \makecell{ground \\ truth \\ object \\ segmentation} &      
      \includegraphics[width=\largeur]{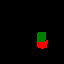} & 
           \includegraphics[width=\largeur]{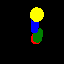} & 
                \includegraphics[width=\largeur]{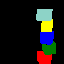} & 
                     \includegraphics[width=\largeur]{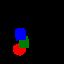} & 
                          \includegraphics[width=\largeur]{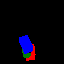} & 
                               \includegraphics[width=\largeur]{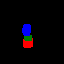} & 
                                    \includegraphics[width=\largeur]{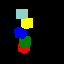} & 
                                         \includegraphics[width=\largeur]{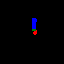} 
   \\

        \makecell{predicted \\ background}  &              
  \includegraphics[width=\largeur]{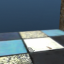}  &
    \includegraphics[width=\largeur]{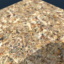}  &
  \includegraphics[width=\largeur]{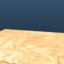}  &
  \includegraphics[width=\largeur]{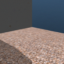}  &
  \includegraphics[width=\largeur]{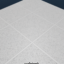}  &
  \includegraphics[width=\largeur]{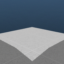}  &
  \includegraphics[width=\largeur]{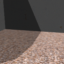}  &
  \includegraphics[width=\largeur]{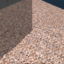}   \\
   \makecell{predicted \\ foreground \\ mask}   &         
  \includegraphics[width=\largeur]{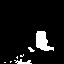}  &
    \includegraphics[width=\largeur]{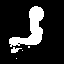}  &
   \includegraphics[width=\largeur]{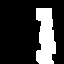}  &
   \includegraphics[width=\largeur]{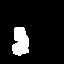}  &
   \includegraphics[width=\largeur]{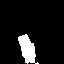}  &
   \includegraphics[width=\largeur]{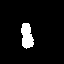}  &
   \includegraphics[width=\largeur]{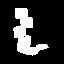}  &
   \includegraphics[width=\largeur]{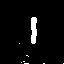}   \\
\end{tabular}}
\caption{Examples of background reconstruction and foreground segmentation on ShapeStacks dataset }
\label{figure:shapestacks}
\end{figure}

\end{document}